\documentclass[11pt,a4paper]{article}

\usepackage{CJKutf8}
\newcommand{\zh}[1]{\begin{CJK}{UTF8}{gbsn}#1\end{CJK}}
\usepackage{booktabs} 

\usepackage[margin=1in]{geometry} 
\usepackage{times} 
\usepackage{tcolorbox} 
\usepackage{graphicx} 
\usepackage{hyperref} 
\usepackage{fontawesome5} 
\usepackage{xcolor} 
\usepackage{caption}
\usepackage{amsmath}

\usepackage{graphicx}
\usepackage{subcaption}
\usepackage{booktabs}
\usepackage{multirow}
\usepackage{pifont} 
\usepackage{wrapfig} 

\usepackage{tabularx}
\usepackage{array}
\usepackage[table]{xcolor}
\usepackage[table]{xcolor}
\definecolor{DeepTeal}{HTML}{008080}
\usepackage{arydshln}
\usepackage{threeparttable}

\usepackage[numbers,sort&compress]{natbib}

\hypersetup{
    colorlinks=true,
    linkcolor=blue,
    filecolor=magenta,      
    urlcolor=teal,
}

\begin{document}

\vspace*{-2.3cm}

\noindent\rule{\textwidth}{2pt}

\begin{center}


\newcommand{\factemoji}{\raisebox{-0.20\height}{\includegraphics[height=1.6\fontcharht\font`\B]{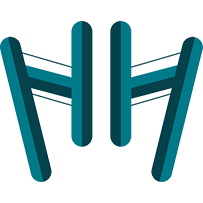}}}
\parbox{0.92\textwidth}{
    \centering
    {\fontsize{18}{22}\selectfont\bfseries
    \factemoji{} JuZhou 1.0 Technical Report
    \par}
    
    \vspace{0.28cm}
    
    {\fontsize{14}{18}\selectfont\bfseries
    The First Edge-Native Text-to-Image Foundation Model\\
    Trained Entirely on China-Developed AI Accelerators
    \par}
}


\end{center}

\noindent\rule{\textwidth}{1pt}

\begin{center}
    {\Large \textbf{JuZhou Team, HSW Group}} \\[0.6cm]
    
    \makebox[\textwidth][c]{%
    {\large
    \textcolor{teal}{\faRocket}\hspace{0.4em}
    \href{https://sc-web.huishiwei.cn/\#/scenarioDetail/48}{\textbf{Test demo}}
    \hspace{1.3em}
    \textcolor{teal}{\faDownload}\hspace{0.4em}
    \href{https://www.pgyer.com/mojiemobilellm-android}{\textbf{App Download}}
    \hspace{1.3em}
    \textcolor{teal}{\faGlobe}\hspace{0.4em}
    \href{https://hswai2026.github.io/JuZhouV1/}{\textbf{Project}}
    \hspace{1.3em}
    \textcolor{teal}{\faGithub}\hspace{0.4em}
    \href{https://github.com/HswAI2026/JuZhou-V1}{\textbf{Code}}
    \hspace{1.3em}
    \textcolor{teal}{\faPlayCircle}\hspace{0.4em}
    \href{https://www.icswb.com/default.php?mod=live_text&live_id=914&temp=live_video}{\textbf{Video}}
    }%
    }
\end{center}

\begin{figure}[h]
\centering
\includegraphics[width=\textwidth]{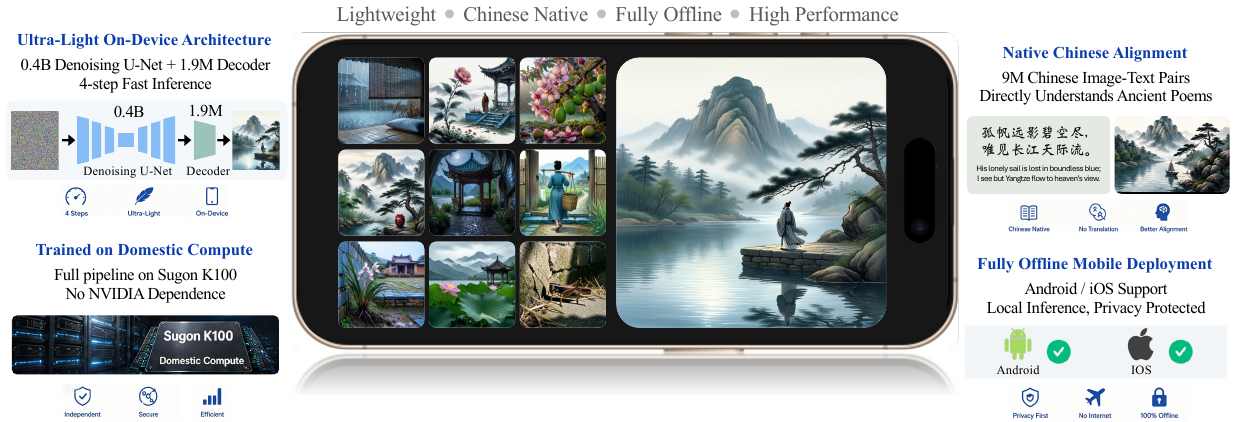}
\caption{\textbf{Overview of JuZhou 1.0.} An edge-native text-to-image foundation model featuring an ultra-light on-device architecture, native Chinese semantic alignment, and fully offline mobile deployment. The entire training pipeline is completed exclusively on domestic compute (Sugon K100).}
\label{fig:teaser}
\end{figure}

\begin{center}
\begin{tcolorbox}[colback=gray!10, colframe=white, arc=5pt, boxrule=0pt, left=6pt, right=6pt, top=8pt, bottom=8pt, width=\textwidth]
\noindent
\textbf{Abstract:}
Text-to-image (T2I) diffusion models typically require substantial computational resources and cloud infrastructure, posing significant challenges for edge deployment in terms of latency, cost, and user privacy. We present JuZhou~1.0, an ultra-lightweight T2I foundation model designed for fully offline, on-device execution. JuZhou~1.0 achieves its efficiency through four key designs:
(1) a compact image-generation backbone consisting of a 0.385B-parameter denoising U-Net and a 1.90M-parameter distilled decoder, totaling approximately 0.387B parameters;
(2) Rectified Flow training combined with DMD2 distillation, reducing inference to 4 sampling steps;
(3) Chinese semantic alignment trained on 9M curated image-text pairs, enabling direct Chinese prompting without external translation at inference time; and
(4) a training and distillation pipeline completed on domestically developed Sugon K100 AI accelerators without relying on NVIDIA GPUs for training or distillation.
Despite its compact scale, the 28-step base model of JuZhou~1.0 achieves an overall GenEval score of 0.70, outperforming published baselines including SDXL (2.6B, 0.55), SD3-Medium (2B, 0.62), and IF-XL (4.3B, 0.61).
We further validate the full poetry-to-image pipeline on Android and the core CLIP--U-Net--VAE generation branch on iOS.
On a smartphone powered by the \textbf{Snapdragon\textsuperscript{\textregistered} 8 Elite Gen 5} Mobile Platform, the 4-step U-Net denoising branch runs in approximately 1.6s, while the full Android poetry-to-image pipeline takes 4.5s with on-device prompt refinement on Xiaomi 17 Pro Max.
These results position JuZhou~1.0 as a practical approach to mobile text-to-image generation and provide a concrete reference for Chinese-native generation, domestic-compute training, and fully offline on-device deployment after one-time installation.
\end{tcolorbox}
\end{center}


\section{Introduction}
Current text-to-image (T2I) generation models~\cite{zhang2023t2i_survey,saharia2022imagen,ramesh2022dalle,podell2023sdxl,flux2024,peebles2023dit,rombach2022high,stability2022sd21,ding2021cogview,taiyi2024} have demonstrated remarkable image synthesis capabilities, but their reliance on large parameter counts, iterative denoising, and substantial computational resources leads to prohibitively high deployment costs. Recent lightweight and on-device T2I systems, such as MobileDiffusion~\cite{zhao2024mobilediffusion} and SnapGen~\cite{hu2024snapgen}, have made important progress in reducing model size and improving mobile inference efficiency. However, these efforts mainly focus on efficient model compression and deployment, while native Chinese text understanding, fully localized application scenarios, and training on domestic accelerator stacks remain underexplored. Moreover, most existing T2I applications still predominantly adopt a cloud-based client-server (C/S) architecture, requiring users to upload prompts and reference images to remote servers, thereby exposing sensitive data to privacy risks such as eavesdropping, unauthorized data collection, and potential breaches.

To address these challenges, we propose \textbf{JuZhou 1.0}, the first ultra-lightweight, native Chinese T2I foundation model supporting fully offline mobile deployment. 
As shown in Fig.~\ref{fig:teaser}, JuZhou 1.0 employs a minimalist 0.385B-parameter denoising U-Net~\cite{ho2020ddpm, rombach2022high} paired with a 1.90M-parameter distilled decoder. Together, these two image-generation modules contain approximately 0.387B parameters, which we denote as the 0.4B JuZhou image-generation backbone for readability.
By leveraging Rectified Flow training~\cite{liu2022flow} and DMD2 distillation~\cite{yin2024improved_dmd}, we compress the inference trajectory to only four steps, enabling image generation within seconds on mobile hardware while ensuring that all user data remains processed locally.

To train JuZhou 1.0, we constructed a large-scale 9M Chinese image-text corpus from filtered DiffusionDB prompts, Stable Diffusion 3.5 Large synthetic images, and Qwen3-based Chinese prompt translation. The original English prompts are used for image synthesis to preserve the open-domain visual distribution, and the translated Chinese prompts provide Chinese textual supervision for learning Chinese text-to-image alignment. For the poetry-oriented corpus, we further apply source-level cleaning, style augmentation, prompt-level quality control, and Qwen3-based recaptioning.

The model training and distillation pipeline of JuZhou 1.0 was completed on Sugon K100 heterogeneous computing clusters equipped with domestic AI accelerators, without dependence on NVIDIA accelerators for training. Quantitative evaluation was conducted separately in a unified NVIDIA V100 environment for fair baseline comparison.
The main foundational training stages used 56 compute nodes with 224 K100 DCUs.
Together with the corresponding domestic software stack, this training setup demonstrates that domestic computing infrastructure can support large-scale visual generative model training and provides a practical reference for future software-hardware co-design.

On the deployment front, we completed full-stack adaptation for iOS and Android platforms. To demonstrate JuZhou 1.0's native Chinese semantic understanding, we developed and publicly released a 4-step classical Chinese poetry-to-image application, showcasing the model's ability to capture nuanced cultural contexts without relying on external translation. On a smartphone powered by the Snapdragon\textsuperscript{\textregistered} 8 Elite Gen 5 Mobile Platform, JuZhou 1.0 completes 4-step U-Net denoising in approximately 1.6 seconds, enabling high-definition image generation entirely offline.

In summary, the primary contributions of this paper are highlighted in the following five aspects:

\begin{itemize}
    \item \textbf{Ultra-Lightweight, Native Chinese T2I Model for Offline Mobile Deployment:} JuZhou 1.0 is the first native Chinese T2I foundation model purpose-built for on-device offline execution~\cite{hu2024snapgen,alibaba2020mnn,qnn}, achieving privacy-preserving local inference through 
    a minimalist image-generation backbone consisting of a 0.385B-parameter denoising U-Net and a 1.90M-parameter distilled decoder, totaling approximately 0.387B parameters.

    \item \textbf{Scalable Chinese Data Construction Pipeline:} We construct a 9M general Chinese image-text corpus from filtered DiffusionDB prompts, SD3.5-Large synthetic images, and Qwen3-based prompt translation, and further build a 1.77M poem-grounded corpus through source cleaning, style-conditioned synthesis, prompt-level quality control, and recaptioning~\cite{schuhmann2022laion,pai2023diffusion,huang2025d2c}, establishing a scalable data foundation for native Chinese and poetry-oriented mobile T2I generation.

    \item \textbf{Domestic Computing Infrastructure Validation:} The full training pipeline, executed on Sugon K100 clusters, validates the feasibility of domestic hardware for large-scale generative AI training and provides a software-hardware co-design reference.

    \item \textbf{Heterogeneous Mobile Deployment:} Full-stack adaptation for both iOS and Android enables standardized edge AI deployment~\cite{alibaba2020mnn,qnn} across major mobile platforms.

    \item \textbf{Native Chinese Application:} A publicly released 4-step classical Chinese poetry-to-image application demonstrates JuZhou 1.0's ability to capture nuanced cultural contexts without external translation modules.
\end{itemize}

\section{Data Curation}

\begin{figure}[t]
\centering
\includegraphics[width=\textwidth]{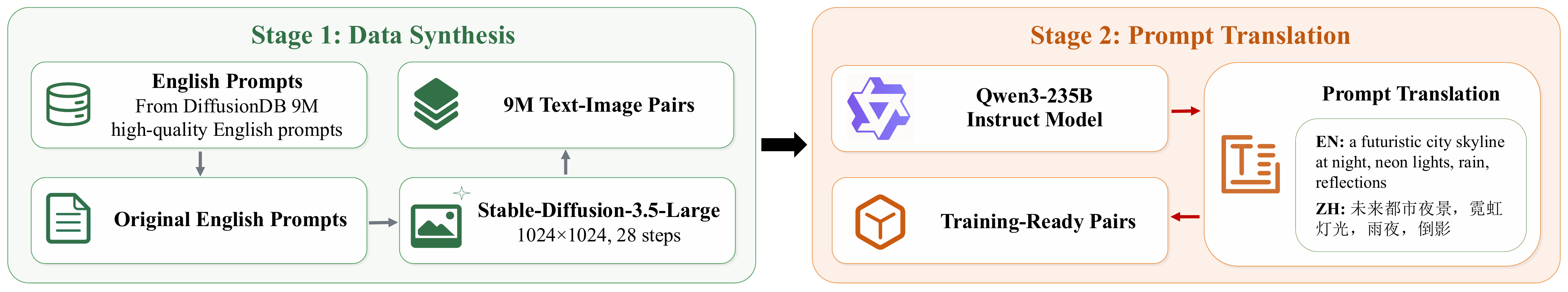}
\caption{\textbf{Overview of the General Chinese Text-to-Image Curation Pipeline.}
Two stages construct a Chinese image-text corpus for training the base model: 
(1) Data Synthesis starts from 9M filtered English text-to-image prompts from DiffusionDB and synthesize images via Stable Diffusion 3.5 Large; 
(2) Prompt Translation converts prompts into Chinese using Qwen3-235B-Instruct, producing training-ready Chinese prompt-image pairs. }
\label{fig:general_data_pipeline}
\end{figure}

\begin{figure}[t]
\centering
\includegraphics[width=\textwidth]{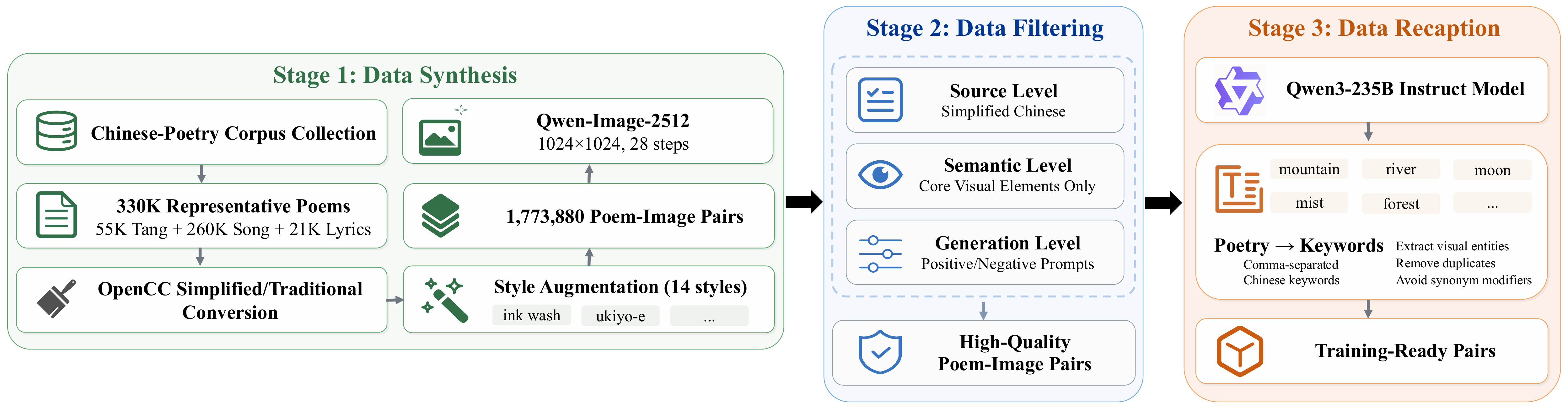}
\caption{\textbf{Overview of the Classical Poetry-to-Image Curation Pipeline.} Three stages transform classical Chinese poetry into 1,773,880 structured image-text pairs: (1)~Data Synthesis expands $\sim$330K poems with 14 stylistic descriptors and generates images via Qwen-Image-2512; (2)~Data Filtering applies source-level, semantic-level, and generation-level quality controls; (3)~Data Recaption converts each poem into compact keyword sequences via Qwen3-235B.}
\label{fig:poem_data_pipeline}
\end{figure}

We construct two complementary Chinese image-text datasets for training and evaluating Chinese text-to-image models. The first is a general-purpose 9M Chinese image-text corpus, which is used to train the base Chinese text-to-image model and provides broad coverage of open-domain visual concepts, scene compositions, and natural-language descriptions. The second is a poem-grounded synthetic corpus containing 1,773,880 poem-image pairs, which is further used to adapt the base model to classical Chinese poetry-to-image generation. This two-stage data design allows the model to first acquire general Chinese text-to-image alignment and then specialize in the visual interpretation of classical poetry. The overall curation pipelines of the general Chinese text-to-image corpus and the poem-grounded corpus are illustrated in Fig.~\ref{fig:general_data_pipeline} and Fig.~\ref{fig:poem_data_pipeline}, respectively.

Compared with general natural-language prompts, classical Chinese poems pose additional challenges for text-to-image generation. Directly using original verses as generation prompts often leads to ambiguous or unstable synthesis results, owing to the condensed syntax, allusive expressions, and semantic distance between classical poetry and modern descriptive language. To address these challenges, we standardize the poetry corpus, convert poems into generation-oriented prompts, and pair the resulting prompts with synthesized images through a dedicated poem-grounded curation pipeline.

\subsection{General Chinese Text-to-Image Corpus}

We first construct a general-purpose 9M Chinese image-text corpus for base model training. Specifically, we start from a filtered set of 9M English text-to-image prompts derived from DiffusionDB~\cite{wang2022diffusiondb}. Each English prompt is used directly as the conditioning input to Stable Diffusion 3.5 Large to synthesize the corresponding image. After image generation, we translate the original English prompts into Chinese using Qwen3-235B-Instruct. The translated Chinese prompts are then paired with the generated images, forming a large-scale Chinese image-text corpus for general Chinese text-to-image pretraining. 

This construction strategy avoids translation-induced prompt drift before image synthesis. Since the images are generated from the original English prompts, the visual semantics and open-domain distribution of the source prompt set are preserved. Meanwhile, the translated Chinese prompts provide Chinese textual supervision, enabling the model to learn general Chinese text-to-image alignment.

\subsection{Poem-Grounded Data Synthesis}

Based on the general Chinese text-to-image corpus, we further construct a poem-grounded dataset for classical Chinese poetry-to-image learning. We build the poetry corpus primarily from the public Chinese-Poetry repository~\cite{ChinesePoetry}, which contains large-scale collections of classical Chinese literature, including Tang poems, Song poems, Song lyrics, and other canonical texts. 
From this repository, we retain approximately 330,000 poems after removing duplicates, malformed entries, and non-poetic metadata, and normalize the remaining corpus into Simplified Chinese using OpenCC~\cite{BYVoidOpenCC} to ensure orthographic consistency for downstream processing and model training.

To enrich the expressive diversity of the poem-grounded dataset, we further augment the poems with a broad set of stylistic descriptors. These descriptors span traditional East Asian visual styles, Western art movements, modern illustration styles, and speculative visual genres. This augmentation strategy expands the visual distribution of the generated data while keeping the original poem as the primary semantic anchor.

The 14 stylistic descriptors form a candidate augmentation pool rather than an exhaustive Cartesian product. A full expansion of approximately 330,000 poems across all 14 styles would produce about 4.6 million candidate combinations. We instead retain only poem--style pairs that pass the filtering process, yielding 1,773,880 poem--image pairs, corresponding to an average of approximately 5.4 valid style-conditioned images per poem.

After prompt construction, we synthesize poem-conditioned images using the open-source Qwen-Image-2512~\cite{wu2025qwenimagetechnicalreport}. All images are generated at a resolution of $1024 \times 1024$ with 28 denoising steps. Each poem is paired with a subset of style-conditioned prompts after filtering, rather than being exhaustively combined with all descriptors. Following this pipeline, we construct 1,773,880 poem-conditioned image-text pairs, which provide domain-specific multimodal data for poem-grounded image generation.

\subsection{Prompt Filtering and Quality Control}

The two datasets follow different prompt-processing strategies according to their roles in training. For the general 9M Chinese image-text corpus, we preserve the original English prompt semantics during image synthesis. Specifically, the filtered DiffusionDB prompts are used directly to condition Stable Diffusion 3.5 Large, without additional semantic pruning, style augmentation, or positive/negative prompt modification. The Chinese prompts are produced only after image generation through translation. This design avoids introducing translation-induced prompt drift before synthesis and maintains the broad open-domain distribution of the original prompt set for base model training.

For the poem-grounded dataset, we adopt the three-stage filtering pipeline illustrated in Fig.~\ref{fig:poem_data_pipeline}. At the source level, we clean and standardize the poetry corpus by removing duplicate poems, malformed entries, and non-poetic metadata, and then normalize Traditional Chinese text into Simplified Chinese using OpenCC. At the semantic level, we retain only visually grounded entities, objects, and scene descriptions from the recaptioned content, while eliminating redundant elements and near-synonymous modifiers. At the generation level, we incorporate carefully designed positive and negative prompts to promote high-resolution rendering and coherent composition while mitigating common visual artifacts.

At the image generation stage for the poem-grounded dataset, we further regulate output quality through positive and negative prompt design, following common practices in large-scale text-to-image generation and data construction~\cite{schuhmann2022laion,wang2022diffusiondb,openclip2022,zhang2024longclip}. The positive prompt encourages desirable visual attributes such as high-resolution rendering, clear composition, and cinematic visual quality. In parallel, the negative prompt suppresses common generation artifacts, including low-resolution images, texture defects, anatomical distortions, oversaturation, unnatural facial appearance, blurred or distorted text, and visually disordered layouts. Rather than relying on an additional post-hoc classifier or filtering module, these controls are integrated directly into the prompt construction and generation process. They serve as a lightweight mechanism for encouraging higher visual quality and reducing visually implausible outputs during large-scale poem-grounded dataset construction.

\subsection{Data Recaption}

The recaptioning stage reformulates each source poem into a compact, visually grounded keyword sequence suitable for Chinese text-to-image generation. This step is necessary because classical Chinese poetry often encodes scenery, objects, emotions, and historical allusions in a highly condensed literary form, whereas modern text-to-image models typically require explicit and visually grounded prompts. To bridge this gap, we employ Qwen3-235B-A22B-Instruct-2507~\cite{qwen3technicalreport} to convert each poem into a comma-separated list of Chinese keywords compatible with diffusion-based text-to-image prompting.

The recaptioning process operationalizes the semantic pruning constraints described above through a constrained instruction format. Rather than performing literal translation, the model maps the condensed literary expression of each poem into explicit visual elements and standardized prompt tokens. The instruction asks the model to extract core realistic visual elements, retain a compact keyword or phrase for each salient object or scene component, remove repeated descriptions, avoid synonymous stacking, and output concise Chinese keywords separated by commas. As a result, the generated prompts are compact, non-repetitive, visually grounded, and can be directly used as inputs to downstream image generation models.

When a stylistic descriptor is appended to the source poem, the recaptioning model incorporates the corresponding style token into the generated prompt while keeping the extracted scene elements unchanged. This enables style-conditioned generation using a unified prompt format. For example, a poem describing a ruined palace site and flowing water can be reformulated into visual keywords such as “palace ruins, millet seedlings, ruined terrace, deer, fresh grass, river, white waves”. When the poem is associated with a specific artistic style, the corresponding style token is appended to the extracted scene elements.

Overall, the resulting corpus is organized as poem-prompt-image triplets. In each triplet, the original poem provides the literary source, the recaptioned prompt provides a standardized generation-oriented textual representation, and the synthesized image provides the corresponding visual realization. These triplets provide structured text-image supervision for poem-grounded image generation and related downstream multimodal learning~\cite{liu2023bdm,chinese-clip}.

\subsection{Edge-Oriented Prompt Refiner Supervision}

While the teacher-generated prompts described above are effective for large-scale dataset construction, our deployed Android application must handle raw classical poems directly provided by users. This creates a deployment-time interface gap: the image generation model expects explicit and visually grounded prompts, whereas users naturally interact with the system through original poetic text. Therefore, beyond constructing poem-prompt-image triplets for image model training, we further reuse the raw poems and their teacher-generated prompts to build text-to-text supervision for an edge-oriented prompt refiner.

Specifically, we use Qwen3-235B-A22B-Instruct-2507 as an offline teacher model to generate structured visual prompts from raw classical poems. The teacher-generated prompts preserve the core poetic semantics while making implicit imagery more explicit and visually grounded. Each training instance is organized as a raw-poem/teacher-generated-prompt pair, where the raw poem serves as the input sequence and the teacher-generated prompt serves as the target output. This formulation distills the teacher model's prompt-construction behavior into a supervised learning problem for a substantially smaller language model.

Since Qwen3-235B-A22B-Instruct-2507 is impractical for mobile edge devices due to its prohibitive memory and computational requirements, we fine-tune Qwen3-1.7B using LoRA on the constructed raw-poem/teacher-generated-prompt pairs. The LoRA adaptation specializes the smaller model for poetry-oriented prompt refinement without requiring full-parameter fine-tuning. As a result, the adapted Qwen3-1.7B model learns to transform raw classical poems into generation-ready prompts that are better aligned with the downstream diffusion model.

In the final system, the fine-tuned Qwen3-1.7B model is deployed and executed on Android devices as an on-device prompt refiner. Given a user-provided poem, it first generates a refined visual prompt, which is then passed to the poetry-oriented text-to-image model for image synthesis. For Android deployment, the adapted model is optimized for mobile inference to reduce memory usage and latency. This design decouples offline large-model supervision from online edge inference: the 235B teacher model provides prompt supervision during data construction, while the LoRA-tuned~\cite{hu2024lora} 1.7B model enables practical on-device prompt refinement in the deployed Android application.
\begin{figure}[t]
\centering
\includegraphics[width=\textwidth]{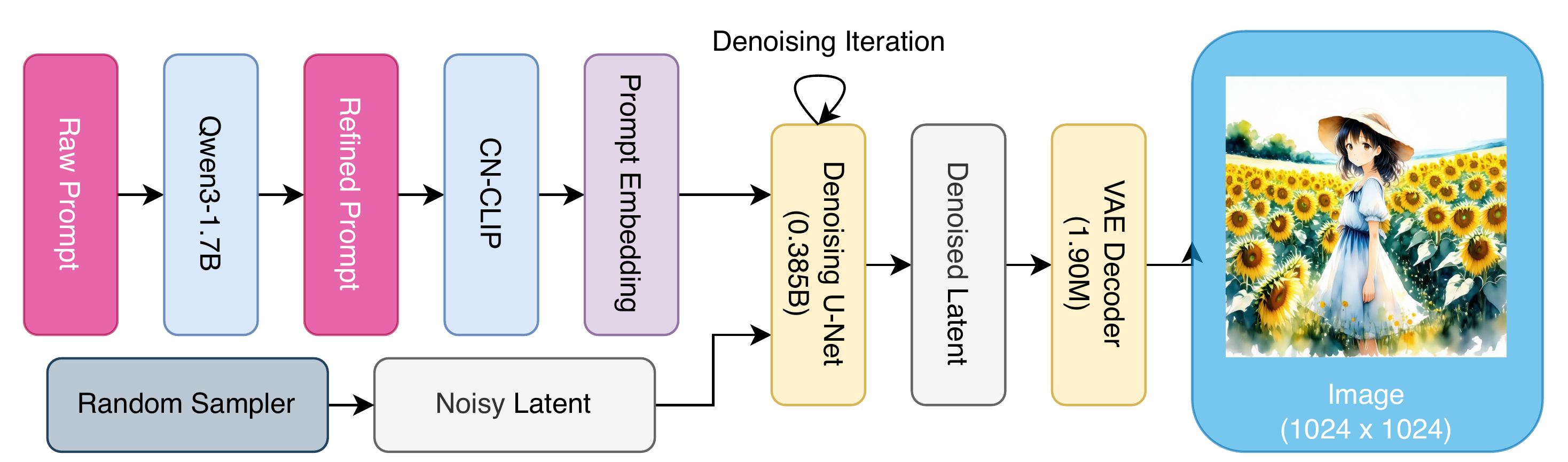}
\caption{\textbf{Overview of the JuZhou 1.0 framework.} A raw poem or user prompt is first refined by Qwen3-1.7B and then encoded by CN-CLIP\protect\footnotemark{} to obtain Chinese semantic conditioning. The conditioning signal is injected into a 0.385B-parameter denoising U-Net for efficient high-resolution generation. The generated latent representation is decoded by an ultra-compact 1.9M-parameter VAE decoder without attention layers. DMD2 distillation further shortens the sampling trajectory from 28 steps to 4 steps, enabling efficient $1024 \times 1024$ image synthesis.}
\label{fig:architecture_overview}
\end{figure}
\footnotetext{The English version adopts CLIP-L and CLIP-G as text encoders, whereas the Chinese version replaces them with Chinese-CLIP ViT-H/14~\cite{chinese-clip}.}

\begin{table}[t]
    \centering
    \fontsize{7.2pt}{8.2pt}\selectfont
    \setlength{\tabcolsep}{2.0pt}

    \begin{minipage}[t]{0.64\linewidth}
        \centering
        \caption{Architectural comparison between JuZhou 1.0 and mainstream text-to-image diffusion models.}
        \label{tab:architecture_comparison}

        \renewcommand{\arraystretch}{1.08}
        \begin{tabular*}{\linewidth}{@{\extracolsep{\fill}}lcccc@{}}
            \toprule
            \textbf{Model} & \textbf{Denoiser+VAE} & \textbf{Denoiser} & \textbf{VAE} & \textbf{Mobile} \\
            \midrule
            SD v1.5              & $\sim$0.91B & $\sim$0.86B & $\sim$49.49M & {\color{red}\ding{55}} \\
            SD v2.1              & $\sim$0.92B & $\sim$0.87B & $\sim$49.49M & {\color{red}\ding{55}} \\
            SDXL 1.0 Base        & $\sim$2.62B & $\sim$2.57B & $\sim$49.49M & {\color{red}\ding{55}} \\
            SD 3.5 Large         & $\sim$8.11B & $\sim$8.06B & $\sim$49.55M & {\color{red}\ding{55}} \\
            LCM-DreamShaper v7   & $\sim$0.91B & $\sim$0.86B & $\sim$49.49M & {\color{red}\ding{55}} \\
            SDXL-Lightning 4-Step& $\sim$2.57B & $\sim$2.57B & --           & {\color{red}\ding{55}} \\
            Sana-600M-1024px     & $\sim$0.59B & $\sim$0.59B & --           & {\color{red}\ding{55}} \\
            MobileDiffusion      & \textit{$\sim$0.396B} & \textit{$\sim$0.386B} & \textit{$\sim$9.8M}  & {\color{green}\ding{51}} \\
            SnapGen              & \textbf{$\sim$0.373B} & \textbf{$\sim$0.372B} & \textbf{$\sim$1.38M} & {\color{green}\ding{51}} \\
            \midrule
            \textbf{JuZhou 1.0}
            & \underline{$\sim$0.387B}
            & \underline{$\sim$0.385B}
            & \underline{$\sim$1.90M}
            & {\color{green}\ding{51}} \\
            \bottomrule
        \end{tabular*}
    \end{minipage}
    \hspace{0.025\linewidth}
    \begin{minipage}[t]{0.30\linewidth}
        \vspace{0pt}
        \centering
        \caption{K100 training cluster specifications.}
        \label{tab:k100_specs}
    
        \renewcommand{\arraystretch}{1.09}
        \begin{tabular*}{\linewidth}{@{\extracolsep{\fill}}ll@{}}
            \toprule
            \textbf{Spec.} & \textbf{Value} \\
            \midrule
            Accelerator & Hygon DCU K100 \\
            FP16/BF16 Perf. & 196 TFLOPS \\
            HBM3 & 64 GB \\
            Mem. Bandwidth & 896 GB/s \\
            Host Interface & PCIe 5.0 x16 \\
            Ecosystem & ROCm \\
            \midrule
            Nodes & 56 \\
            Accel./Node & 4 \\
            Total Accel. & 224 \\
            Interconnect & InfiniBand RDMA \\
            \bottomrule
        \end{tabular*}
    \end{minipage}

    \renewcommand{\arraystretch}{1.0}
\end{table}

\section{Ultra-Lightweight Architecture Design}

To enable high-resolution text-to-image generation on resource-constrained edge devices, JuZhou 1.0 adopts an ultra-lightweight architecture centered on a compact denoising U-Net and a highly compressed VAE decoder. As illustrated in Fig.~\ref{fig:architecture_overview}, a raw poem or user prompt is first refined by the lightweight prompt refiner and then encoded by the Chinese text encoder to obtain semantic conditioning. 
The conditioning signal is injected into a 0.385B-parameter denoising U-Net for latent-space generation, and the resulting latent representation is decoded by a 1.90M-parameter attention-free VAE decoder. Together, the denoiser and decoder contain approximately 0.387B parameters; the rounded 0.4B figure refers only to this image-generation backbone.
DMD2 distillation further reduces sampling from 28 to 4 steps, enabling efficient $1024 \times 1024$ synthesis.
Tab.~\ref{tab:architecture_comparison} compares JuZhou 1.0 with mainstream text-to-image diffusion models in parameter scale. We then introduce the lightweight denoising network and compact VAE decoder.

\subsection{Denoising Network}
\begin{figure}[t]
\centering
\captionsetup{skip=6pt}
\includegraphics[ width=\textwidth, trim=0 20pt 0 0, clip ]{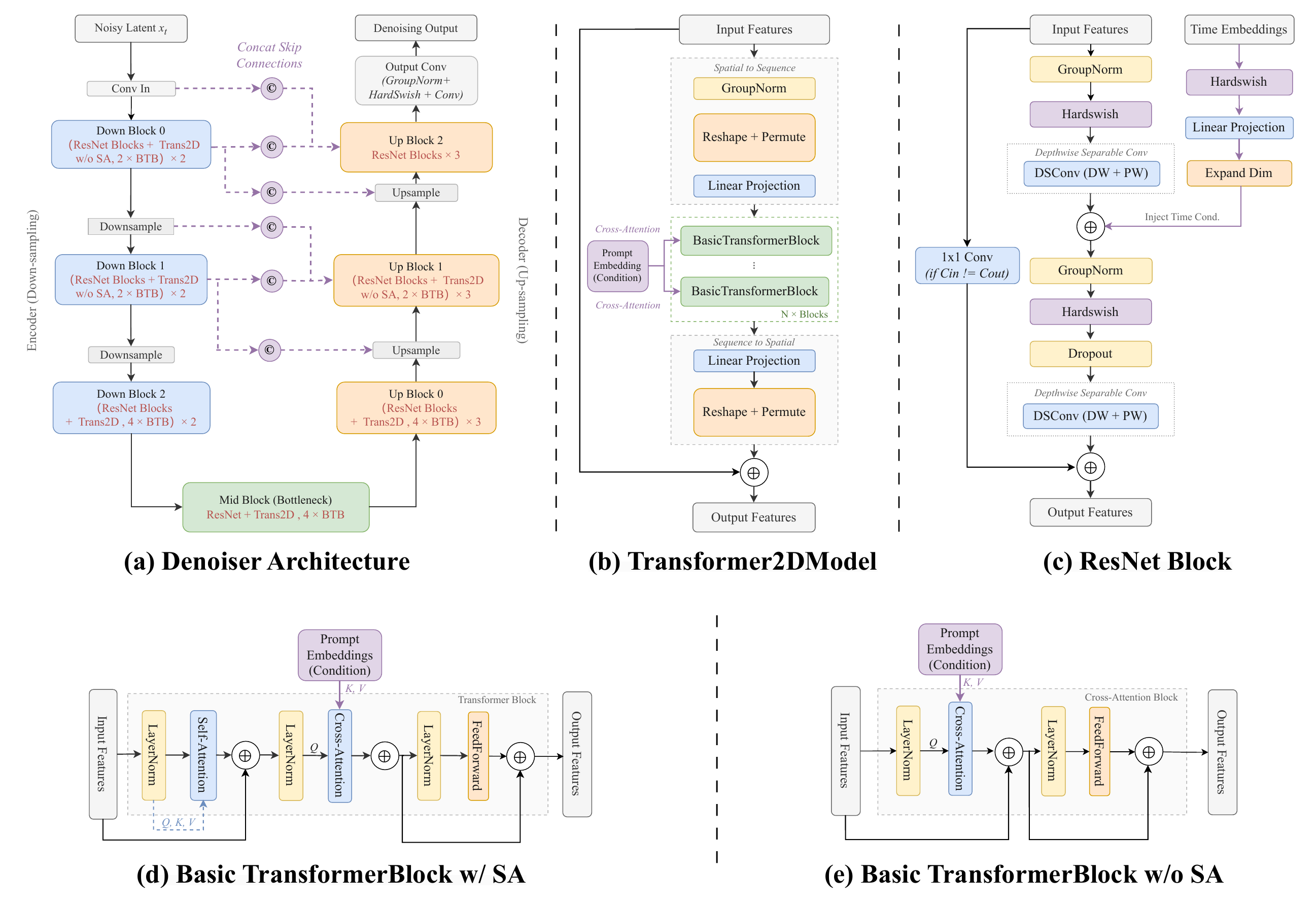}
\caption{Overview of the lightweight denoising network. (a) Denoiser architecture. (b) Transformer2DModel. (c) ResNet Block. (d) BasicTransformerBlock w/ SA. (e) BasicTransformerBlock w/o SA. BTB denotes BasicTransformerBlock, and SA denotes self-attention. }
\label{fig:denoising_overview}
\end{figure}

\begin{figure}[t]
\centering
\captionsetup{skip=6pt}
\includegraphics[ width=\textwidth, trim=0 20pt 0 0, clip ]{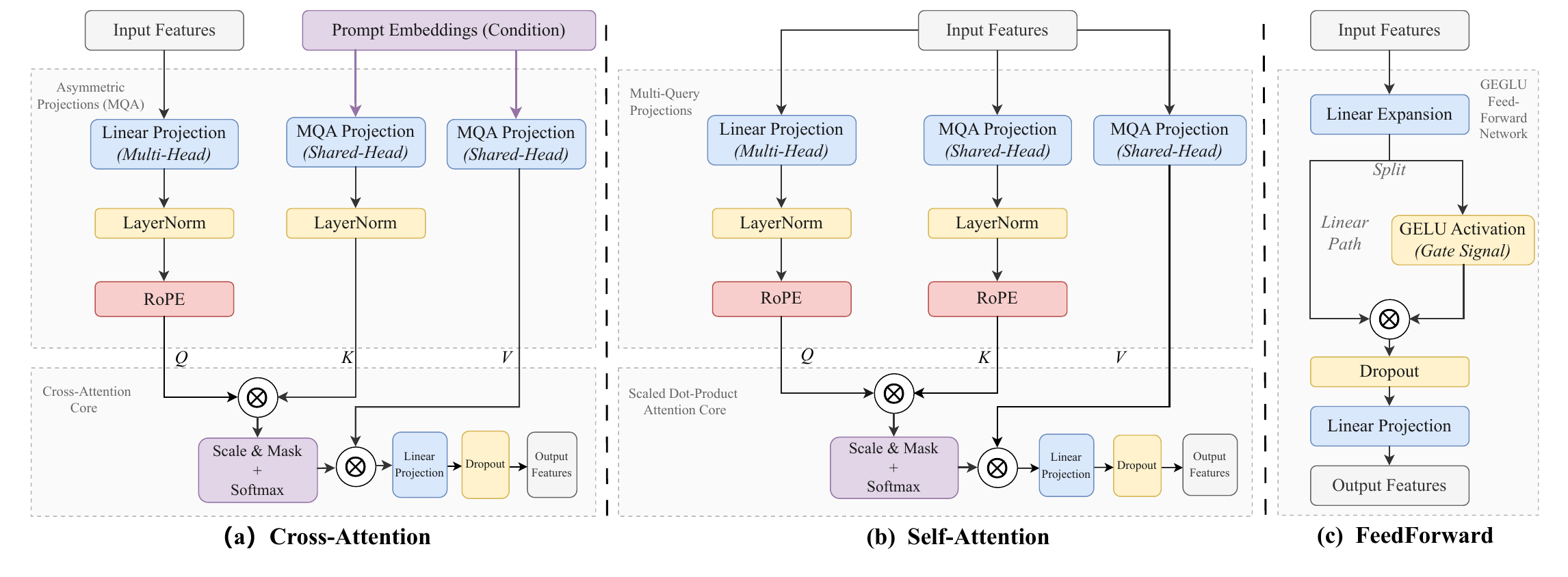}
\caption{Detailed modules in the denoising network. (a) Cross-attention. (b) Self-attention. (c) GEGLU-based feed-forward network. }
\label{fig:denoising_Detailed_modules}
\end{figure}

As illustrated in Fig.~\ref{fig:denoising_overview}, the denoising network follows an encoder--bottleneck--decoder U-Net structure while substantially reducing the parameter count and computational cost. Given an input latent representation ($x_t$), the model first projects it into the backbone feature space using a ($3 \times 3$) convolutional input layer. The encoder consists of three down blocks, each combining ResNet blocks with Transformer2D modules to capture both local spatial patterns and text-conditioned semantics. After the encoder, the middle block further integrates ResNet blocks with Transformer2D modules to refine the bottleneck representation through global semantic interactions. The decoder contains three resolution-aware up blocks, where the first two up blocks also integrate ResNet blocks with Transformer2D modules, while the final up block is composed only of ResNet blocks. The decoder progressively recovers spatial resolution and produces the final denoising prediction through an output projection head composed of GroupNorm, Hardswish, and a ($3 \times 3$) convolution.

The detailed attention and feed-forward modules used in the denoising network are shown in Fig.~\ref{fig:denoising_Detailed_modules}. 
To improve efficiency, the network reduces high-resolution redundancy and selectively applies attention based on stage-specific resolution and semantic needs.
Specifically, the Transformer2D modules in the first two down blocks and the first two up blocks remove self-attention and retain only cross-attention, allowing text-conditioned feature modulation with lower computational cost. In contrast, the remaining Transformer-equipped stages, including the third down block and the middle block, employ both self-attention and cross-attention to enhance global spatial reasoning and text-image interaction. Within each ResNet block, GroupNorm and Hardswish are used to stabilize training and provide hardware-friendly nonlinear transformation. 
In Transformer2D modules, LayerNorm is applied before attention and feed-forward operations to stabilize feature interactions under the compact model capacity.

The skip connections are redesigned to preserve multi-scale spatial details while reducing redundant feature fusion near the bottleneck. Specifically, the shallow features from the input convolution are directly connected to the highest-resolution decoder stage, while selected features from Down Block 0 and Down Block 1 are fused into the intermediate upsampling and decoder stages. In contrast, the skip connection from Down Block 2 to Up Block 0 is removed to avoid unnecessary bottleneck-level feature transfer. These lightweight multi-scale skip pathways help the decoder recover local structures and semantic details without substantially increasing model capacity.

\subsection{VAE Decoder}
\begin{figure}[t]
\centering
\captionsetup{skip=4pt}
\includegraphics[
    width=\textwidth,
    trim=0 20pt 0 0,
    clip
]{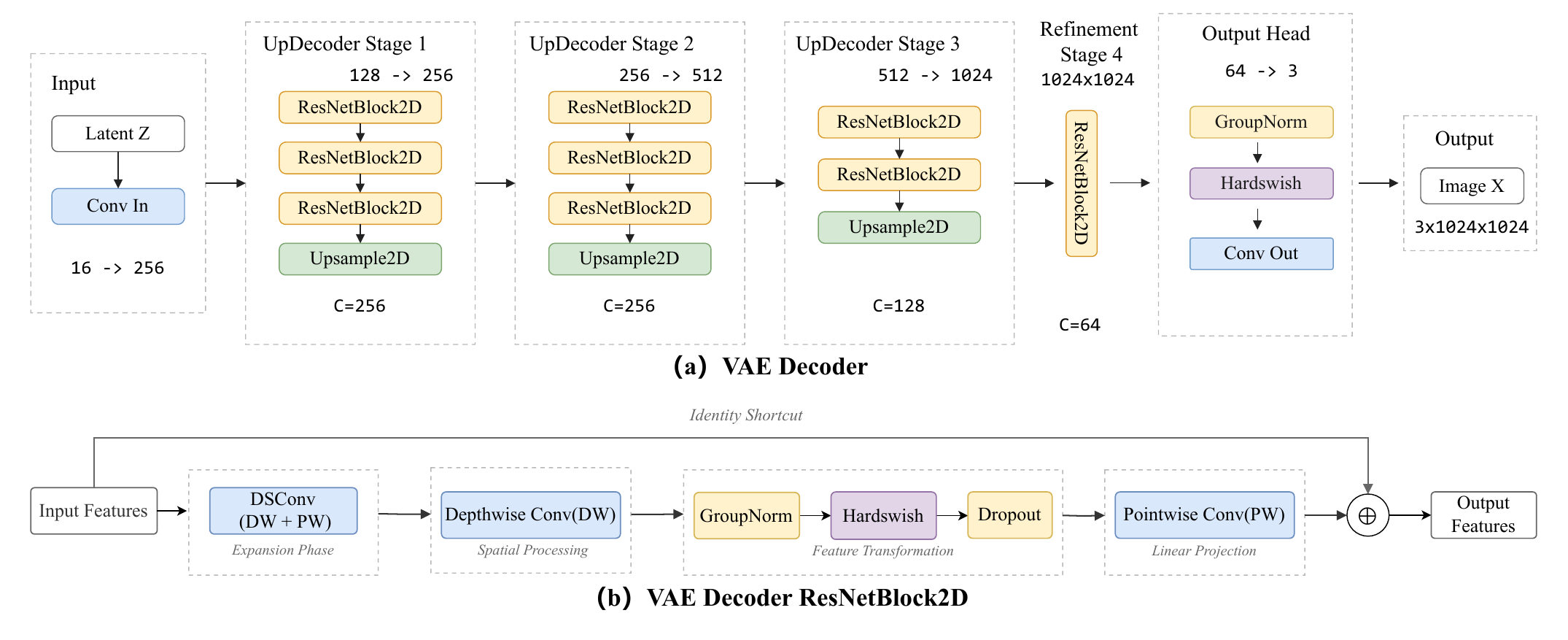}
\caption{Overview of the compact VAE decoder. (a) VAE decoder architecture. (b) VAE decoder ResNetBlock2D.}
\label{fig:VAE_decoder_overview}
\end{figure}

As shown in Fig.~\ref{fig:VAE_decoder_overview}, image decoding is a major memory bottleneck for on-device high-resolution generation. We therefore redesign the VAE decoder as a compact attention-free architecture. It projects the 16-channel latent into 256 channels and progressively decodes features with channel widths of 256, 256, 128, and 64, reducing parameter storage and peak activation memory.

Each decoder block contains multiple lightweight ResNet blocks. Instead of using standard dense convolutions throughout the block, we adopt a depthwise--pointwise convolutional design to reduce convolutional redundancy. Specifically, each block sequentially applies depthwise convolution, pointwise convolution, another depthwise convolution, GroupNorm, Hardswish, Dropout, and a final pointwise convolution. This factorized structure separates spatial filtering from channel mixing, making the decoder more efficient under mobile memory and bandwidth constraints. The attention-free decoding path further helps reduce peak activation memory during high-resolution image reconstruction, while preserving local feature transformation capacity.

The output head further refines the final 64-channel feature map with an additional lightweight ResNet block, followed by GroupNorm, Hardswish, and a final convolution that maps the feature representation to a 3-channel RGB image. By removing attention layers, narrowing channel widths, and simplifying residual block configurations, the proposed decoder reduces the VAE decoder to approximately 1.9M parameters, making it suitable for memory-constrained mobile deployment.

\section{Large-scale Training Adaptation on Hygon DCU K100}

The hardware specifications of the Sugon K100 AI accelerator and the main 56-node training configuration are summarized in Tab.~\ref{tab:k100_specs}. To reduce dependence on foreign hardware supply chains, we deployed the JuZhou~1.0 model training and distillation pipeline on domestic Sugon K100 AI accelerators (Hygon DCU K100 AI). Each K100 AI unit provides 196\,TFLOPS of FP16/BF16 compute, 64\,GB of HBM3 memory, 896\,GB/s memory bandwidth, and PCIe 5.0 x16 connectivity. The main training configuration uses 56 nodes with 4 accelerators per node, totaling 224 K100 AI devices. Built on a ROCm-compatible software stack, the platform supports PyTorch-based diffusion training after targeted runtime adaptation and operator-level optimization.

Adapting the training pipeline to the K100 AI platform required substantial software engineering in collaboration with Sugon's engineering team. Starting from the vendor-provided PyTorch Docker image, we upgraded the runtime to support modern diffusion training dependencies, including \texttt{transformers} and \texttt{accelerate}. A key challenge was ensuring correct gradient backpropagation under the Rectified Flow formulation, as several operators showed numerical discrepancies or unsupported backward passes in the initial environment. Through targeted debugging, operator-level patching, and end-to-end gradient validation, we established a stable training runtime. We further optimized performance-critical modules, including attention variants and separable convolutions, to better match the K100 AI hardware characteristics.

Scaling to 56 nodes introduced significant cross-node communication overhead, particularly during gradient synchronization in distributed data-parallel training. To mitigate this bottleneck, we leveraged the InfiniBand (IB) networking modules integrated into the Sugon servers. With high-bandwidth, low-latency RDMA support, IB reduces CPU-side data movement and improves the efficiency of cross-node collective communication such as All-Reduce. Together with feature pre-computation for reusable text-side embeddings and image latents, this communication stack helped the 224-device cluster maintain high throughput during both low-resolution pre-training and progressive resolution scaling.

To our knowledge, JuZhou~1.0 is among the first text-to-image foundation models trained and distilled on domestic Chinese AI accelerators, providing practical evidence for large-scale visual generative model training on a domestic software--hardware stack.

\begin{figure}[t]
\centering
    \centering
    \begin{minipage}{0.63\textwidth}
        \centering
        \includegraphics[width=\textwidth]{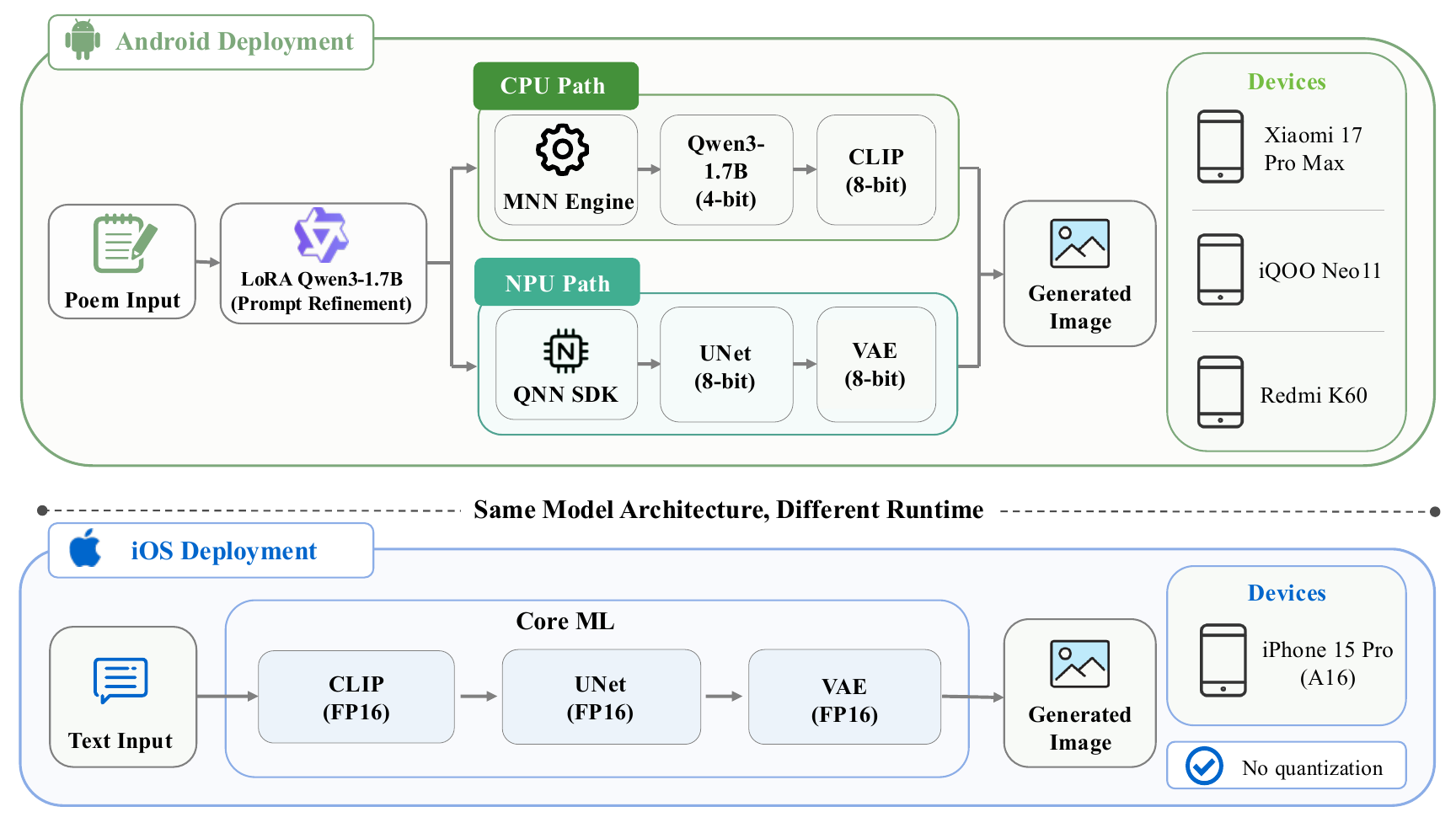}
        \caption{\textbf{Heterogeneous deployment architecture across Android and iOS platforms.} On Android, the pipeline is partitioned between CPU (MNN engine for language model and text encoding) and NPU (QNN for U-Net and VAE decoding). On iOS, all diffusion modules run on Core ML in FP16 precision without additional quantization.}
        \label{fig:deployment_architecture}
    \end{minipage}
    \hfill  
    \begin{minipage}{0.33\textwidth}
        \centering
        \includegraphics[width=\textwidth]{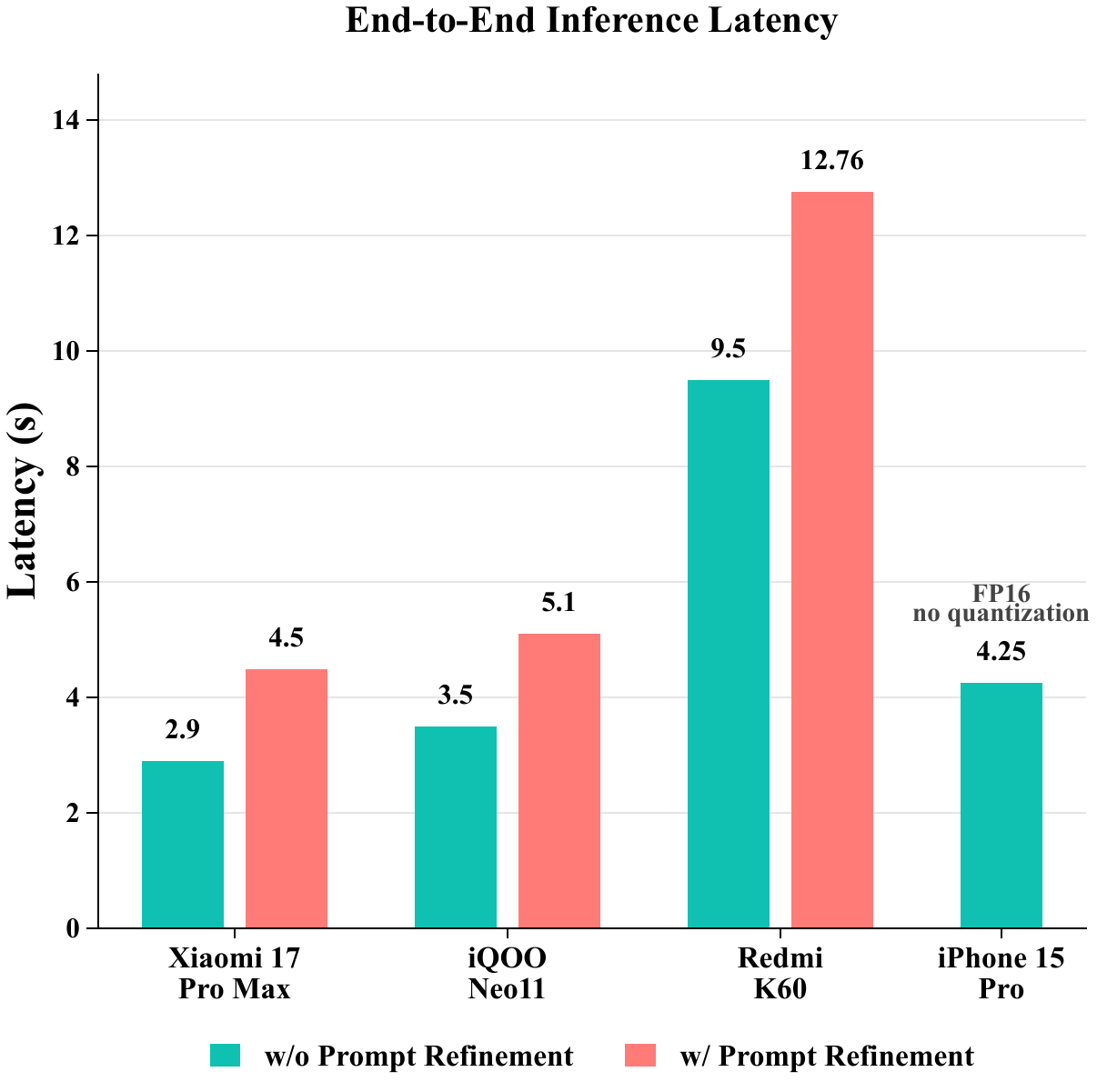}
        \caption{\textbf{Mobile inference latency.} Xiaomi 17 Pro Max completes the full pipeline in 4.5\,s; iPhone 15 Pro runs the CLIP--U-Net--VAE branch in 4.25\,s (FP16).}
        \label{fig:latency_comparison}
    \end{minipage}
\end{figure}

\section{Edge Device Deployment}

To enable fully on-device T2I generation, we adopt a heterogeneous execution pipeline shown in Fig.~\ref{fig:deployment_architecture} that separates text processing from image synthesis. In the poetry application, a LoRA-tuned Qwen3-1.7B language model~\cite{qwen3technicalreport} first rewrites the input poem into a clearer, generation-oriented description. The refined prompt then feeds into the core diffusion pipeline, which comprises three modules: a Chinese CLIP text encoder~\cite{chinese-clip}, a denoising network, and a VAE decoder.

These components exhibit substantially different computational characteristics and are therefore not mapped to a single runtime backend. Instead, control-dominant and text-heavy modules execute on the CPU via lightweight inference engines, while the computationally intensive denoising and decoding stages are offloaded to dedicated neural accelerators whenever available. This design minimizes cloud dependence, preserves data locality, and enables responsive interactive generation on mobile devices. Recent advances in low-bit quantization for both diffusion~\cite{li2023qdiffusion} and transformer~\cite{wu2023int4,liu2025quantization_survey} models informed our deployment strategy.

\subsection{Adaptation Solutions for Snapdragon-based Android devices}

The Android deployment adopts a heterogeneous runtime strategy that partitions the generation pipeline across CPU and NPU backends according to the characteristics of each module. Prompt refinement and text encoding run on the CPU via the MNN engine, while arithmetic-intensive U-Net denoising and VAE decoding are offloaded to the Qualcomm NPU through the QNN SDK. We validate this partitioning across three Snapdragon-based devices spanning multiple hardware generations.

We use the Snapdragon\textsuperscript{\textregistered} 8 Elite Gen 5
Mobile Platform (SM8850) as the primary Android acceleration target and
treat the other Snapdragon devices as cross-generation validation platforms.
In this setup, model execution is not merely a direct port of the desktop
diffusion pipeline. Instead, the diffusion branch is compiled into
QNN-compatible artifacts and executed through the HTP backend, whereas
the text branch remains on the CPU-side MNN runtime. The Android application
identifies the target SoC at runtime and loads the corresponding U-Net and
VAE binaries, allowing the same high-level model architecture to be preserved
while the low-level execution path is specialized for each Snapdragon platform.

\subsubsection{Device platforms and runtime partitioning}
We validate the Android implementation on three Snapdragon-based smartphones across different hardware generations: Xiaomi 17 Pro Max (SM8850, Android 16), iQOO Neo11 (SM8750, Android 16), and Redmi K60 (SM8475, Android 15). The LoRA-tuned Qwen3-1.7B prompt-refinement model and CLIP text encoder run on the CPU via MNN~\cite{alibaba2020mnn} with 4-bit and 8-bit quantization, respectively. In contrast, the computation-heavy U-Net and VAE are deployed on the NPU via Qualcomm AI Engine Direct SDK~\cite{qnn} with 8-bit quantization.

We select this runtime partitioning for two reasons. First, U-Net denoising and VAE decoding dominate arithmetic intensity and benefit most from hardware acceleration. Second, the language model and the text encoder benefit from the portability and lower integration overhead of a CPU-side runtime. In deployment, all model files are hosted remotely, downloaded into an application-specific directory, and subsequently loaded through native C++ interfaces for local inference. On flagship devices, the end-to-end application generates a high-quality image from a poem in approximately 5 seconds.

\subsubsection{Model conversion, packaging, and runtime orchestration}

For the diffusion branch, U-Net and VAE checkpoints are exported from PyTorch to ONNX and converted into QNN-compatible artifacts for NPU execution via the HTP backend. The Android application selects chip-specific U-Net and VAE binaries according to device identification while maintaining a unified application interface.

For the prompt-refinement model and the CLIP text encoder, we adopt the MNN deployment path. We convert the CLIP encoder from ONNX to MNN with 8-bit quantization and export the LoRA-tuned Qwen3-1.7B model through the MNN language-model pipeline with 4-bit compression. At runtime, the application downloads the required model files from cloud object storage into the local app sandbox and initializes them through native C++ interfaces. The execution order remains fixed: poem input, prompt refinement, CLIP encoding, iterative U-Net denoising, VAE decoding, and final image output.



\subsubsection{Stability, resource usage, and latency}

We evaluate Android deployment stability from three aspects: hardware adaptation, runtime performance, and long-run resource usage. For hardware adaptation, the NPU-side U-Net and VAE binaries are converted separately for each Snapdragon platform and dynamically loaded according to the detected chip, while the CPU-side MNN models are shared across devices.

For runtime stability, we repeatedly measure language-model first-token latency and throughput, CLIP encoding time, U-Net single-step denoising latency, and VAE decoding time, and conduct 100 consecutive image-generation runs to monitor CPU utilization and memory usage. Across all evaluated Android devices, the application remains operational without abnormal interruption, with latency and memory footprint staying within stable ranges. Tab.~\ref{tab:android_deployment_results} reports the end-to-end latency and memory usage with and without prompt refinement: Panel~(a) summarizes the results across three Snapdragon-based Android devices, while Panel~(b) compares hybrid CPU+NPU execution and pure-NPU execution on Xiaomi 17.

\begin{table}[t]
\centering
\caption{Android deployment performance on Snapdragon-based mobile platforms. ``Ref.'' denotes prompt refinement.}
\label{tab:android_deployment_results}
\footnotesize
\setlength{\tabcolsep}{4pt}
\renewcommand{\arraystretch}{1.08}

\textbf{(a) End-to-end deployment performance across devices}

\vspace{2pt}

\begin{tabular*}{0.98\linewidth}{@{\extracolsep{\fill}}lllcccc@{}}
\toprule
\textbf{Device} &
\textbf{Mobile Platform} &
\textbf{OS} &
\textbf{Lat. w/o Ref.} &
\textbf{Lat. w/ Ref.} &
\textbf{Mem. w/o Ref.} &
\textbf{Mem. w/ Ref.} \\
\midrule

Xiaomi 17 Pro Max &
Snap. 8 Elite Gen 5 &
Android 16 &
2.9 s &
4.5 s &
200 MB &
1.3 GB \\

iQOO Neo11 &
Snap. 8 Elite &
Android 16 &
3.5 s &
5.1 s &
200 MB &
1.3 GB \\

Redmi K60 &
Snap. 8+ Gen 1 &
Android 15 &
9.5 s &
12.7 s &
180 MB &
1.3 GB \\

\bottomrule
\end{tabular*}

\vspace{5pt}

\textbf{(b) Backend comparison on Xiaomi 17}

\vspace{2pt}

\begin{tabular*}{0.88\linewidth}{@{\extracolsep{\fill}}llcccc@{}}
\toprule
\textbf{Backend} &
\textbf{OS} &
\textbf{Lat. w/o Ref.} &
\textbf{Lat. w/ Ref.} &
\textbf{App Mem. w/o Ref.} &
\textbf{App Mem. w/ Ref.} \\
\midrule

CPU+NPU &
Android 16 &
-- &
4.906 s &
-- &
1.34 GB \\

NPU &
Android 16 &
1.672 s &
3.568 s &
155 MB &
187 MB \\

\bottomrule
\end{tabular*}

\vspace{2pt}

\begin{minipage}{0.88\linewidth}
\scriptsize
\emph{Note:} For Xiaomi 17 with prompt refinement, the CPU+NPU latency is composed of 3.191 s for prompt refinement and 1.715 s for image generation, while the pure-NPU latency is composed of 1.907 s and 1.661 s, respectively. For the pure-NPU backend, memory consumption is reported as host-visible application-process memory measured by Android Studio Profiler. The internal memory allocated by the vendor NPU/HTP runtime, including device-side buffers and on-chip memory, is not exposed by standard Android profiling interfaces and is therefore not separately reported.
\end{minipage}

\renewcommand{\arraystretch}{1.0}
\end{table}

\subsection{Cross-Platform Deployment Mapping}

This deployment preserves the same high-level model decomposition across mobile ecosystems despite different runtime stacks. On Android, U-Net and VAE run on the NPU via QNN, while the language model and CLIP encoder run on the CPU via MNN. On iOS, the validated deployment focuses on the CLIP--U-Net--VAE branch, with all three models converted to Core ML packages and invoked through a Swift application. Tab.~\ref{tab:correspondence_android_ios} reports the component-level correspondence between the two deployment pipelines.

\begin{table}[t]
    \centering
    \caption{Correspondence between Android and iOS deployment components.}
    \label{tab:correspondence_android_ios}
    \footnotesize
    \setlength{\tabcolsep}{4pt}
    \renewcommand{\arraystretch}{1.12}

    \begin{tabularx}{0.96\linewidth}{
        @{}
        >{\raggedright\arraybackslash}p{0.28\linewidth}
        >{\raggedright\arraybackslash}X
        >{\raggedright\arraybackslash}X
        @{}
    }
        \toprule
        \textbf{Component} & \textbf{Android} & \textbf{iOS} \\
        \midrule
        Prompt refinement & MNN CPU, 4-bit & Not used in validated setup \\
        CLIP text encoder & MNN CPU, INT8 & Core ML, FP16 \\
        U-Net denoiser & QNN NPU, INT8 & Core ML, FP16 \\
        VAE decoder & QNN NPU, INT8 & Core ML, FP16 \\
        Model delivery & Remote download to app sandbox & Core ML model packages \\
        Orchestration layer & Native C++ Android app & Swift app with Core ML \\
        \bottomrule
    \end{tabularx}

    \renewcommand{\arraystretch}{1.0}
    \setlength{\tabcolsep}{6pt}
\end{table}

\subsection{iOS Deployment Adaptation}

On iOS, Apple's Core ML framework provides a unified runtime for the CLIP text encoder, U-Net denoiser, and VAE decoder. Unlike Android's MNN--QNN partitioned execution, the iOS pipeline consolidates all three diffusion modules into a single \texttt{coremltools}-based inference stack, running in FP16 precision without additional quantization.

\subsubsection{Model export and Core ML conversion}

In the validated iOS deployment, the iOS branch consists of three core image-generation modules: CLIP, U-Net, and VAE. We first export each module to ONNX, after which the ONNX graphs are converted into \texttt{.mlpackage} format using \texttt{coremltools}. This conversion path preserves the modular boundaries of the original diffusion system and aligns the deployment with Apple’s native model format.

At the current stage, the validated iOS implementation focuses on the CLIP--U-Net--VAE image-generation stack rather than the auxiliary Qwen3-1.7B prompt-refinement model used in the Android pipeline. We note this distinction explicitly because the reported iPhone measurements correspond to the deployed three-model image-generation branch. This branch represents the computationally dominant part of the system and therefore provides a meaningful validation of Apple-side edge deployment.

\begin{table}[t]
    \centering
    \caption[UNet-only on-device profiling at $1024\times1024$ resolution.]%
    {UNet-only on-device profiling at $1024\times1024$ resolution across flagship mobile platforms. The reported latency measures only the 4-step denoising branch of the distilled JuZhou 1.0 model. CLIP text encoding, VAE decoding, prompt refinement, and application-level orchestration are excluded. Mainstream baselines such as SD 1.5 and SDv2.1 fail to execute under the same mobile profiling setup because of excessive memory requirements or impractical latency.\protect\footnotemark}
    \label{tab:hardware_profiling}
    \footnotesize
    \setlength{\tabcolsep}{3.5pt}
    \renewcommand{\arraystretch}{1.10}

    \begin{tabular}{lcccccccc}
        \toprule
        \textbf{Model} 
        & \textbf{\begin{tabular}[c]{@{}c@{}}Denoiser\\+ VAE Dec.\end{tabular}}
        & \textbf{\begin{tabular}[c]{@{}c@{}}Denoiser\\Params\end{tabular}}
        & \textbf{\begin{tabular}[c]{@{}c@{}}VAE Dec.\\Params\end{tabular}}
        & \textbf{\begin{tabular}[c]{@{}c@{}}Platform\\(SoC)\end{tabular}}
        & \textbf{Prec.}
        & \textbf{\begin{tabular}[c]{@{}c@{}}Peak Mem\\(MB)\end{tabular}}
        & \textbf{\begin{tabular}[c]{@{}c@{}}Step\\(s)\end{tabular}}
        & \textbf{\begin{tabular}[c]{@{}c@{}}4-step\\Total (s)\end{tabular}} \\
        \midrule
        SD 1.5 
        & $\sim$0.91B & $\sim$0.86B & $\sim$49.49M 
        & iPhone 15 (A17 Pro) 
        & FP16 & -- & -- & -- \\
        SD 1.5 
        & $\sim$0.91B & $\sim$0.86B & $\sim$49.49M 
        & OnePlus 13 (SD 8 Elite) 
        & INT8 & -- & -- & -- \\
        SDv2.1 
        & $\sim$0.92B & $\sim$0.87B & $\sim$49.49M 
        & iPhone 15 (A17 Pro) 
        & FP16 & -- & -- & -- \\
        SDv2.1 
        & $\sim$0.92B & $\sim$0.87B & $\sim$49.49M 
        & OnePlus 13 (SD 8 Elite) 
        & INT8 & -- & -- & -- \\
        \midrule
        \rowcolor{DeepTeal!10} 
        JuZhou 1.0
        & $\sim$0.387B & $\sim$0.385B & $\sim$1.90M
        & iPhone 15 (A17 Pro) 
        & FP16 & $\sim$373 & $\sim$0.71 & $\sim$2.84 \\
        \rowcolor{DeepTeal!10} 
        JuZhou 1.0
        & $\sim$0.387B & $\sim$0.385B & $\sim$1.90M
        & OnePlus 13 (SD 8 Elite) 
        & INT8 & $\sim$200 & $\sim$0.40 & $\sim$1.60 \\
        \bottomrule
    \end{tabular}

    \renewcommand{\arraystretch}{1.0}
    \setlength{\tabcolsep}{6pt}
\end{table}

\begin{figure}[htbp]
    \centering
    \begin{subfigure}[b]{0.23\linewidth}
        \centering
        \includegraphics[height=3in]{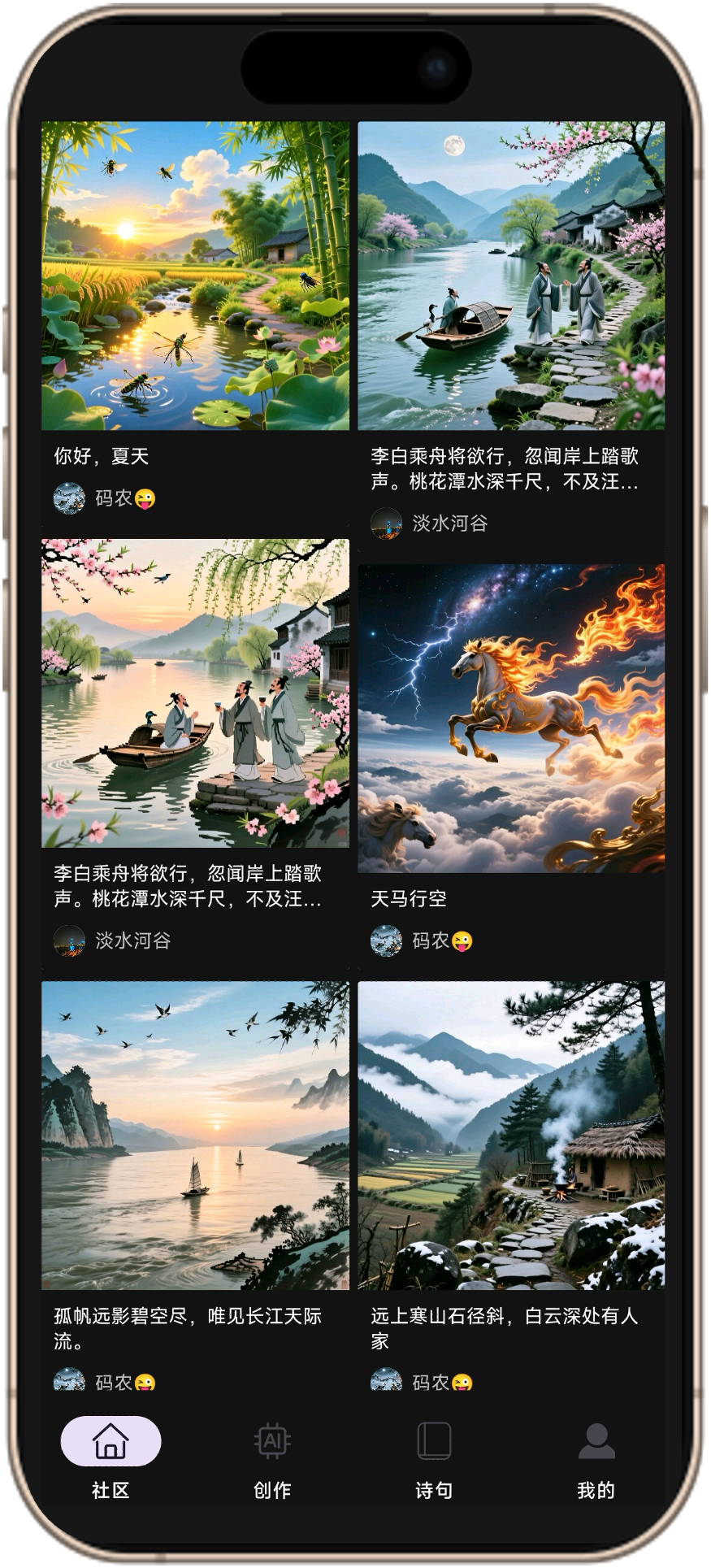}
        \caption{Overview}
        \label{fig:mojie_overview}
    \end{subfigure}
    \hfill
    \begin{subfigure}[b]{0.23\linewidth}
        \centering
        \includegraphics[height=3in]{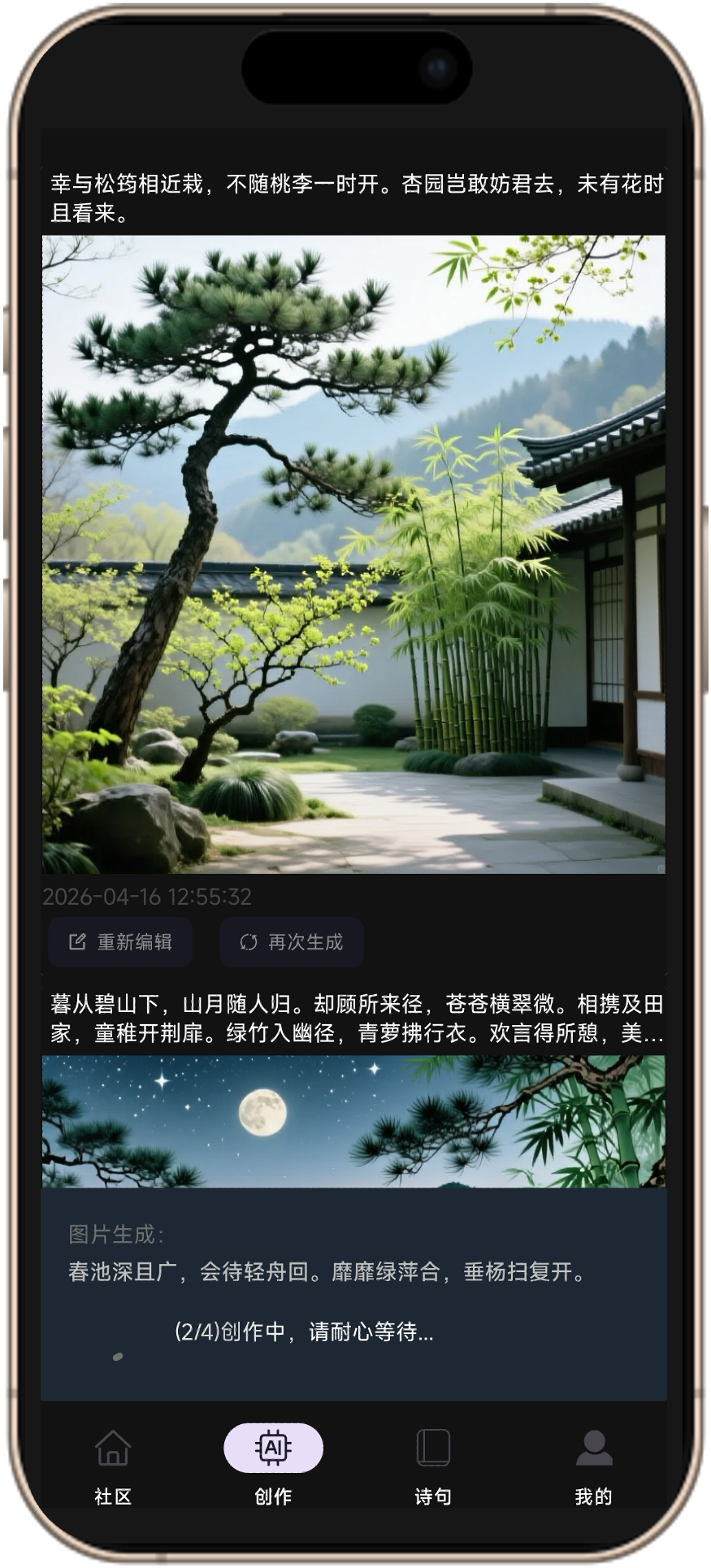}
        \caption{Offline Generation}
        \label{fig:generation_in_mojie}
    \end{subfigure}
    \hfill
    \begin{subfigure}[b]{0.23\linewidth}
        \centering
        \includegraphics[height=3in]{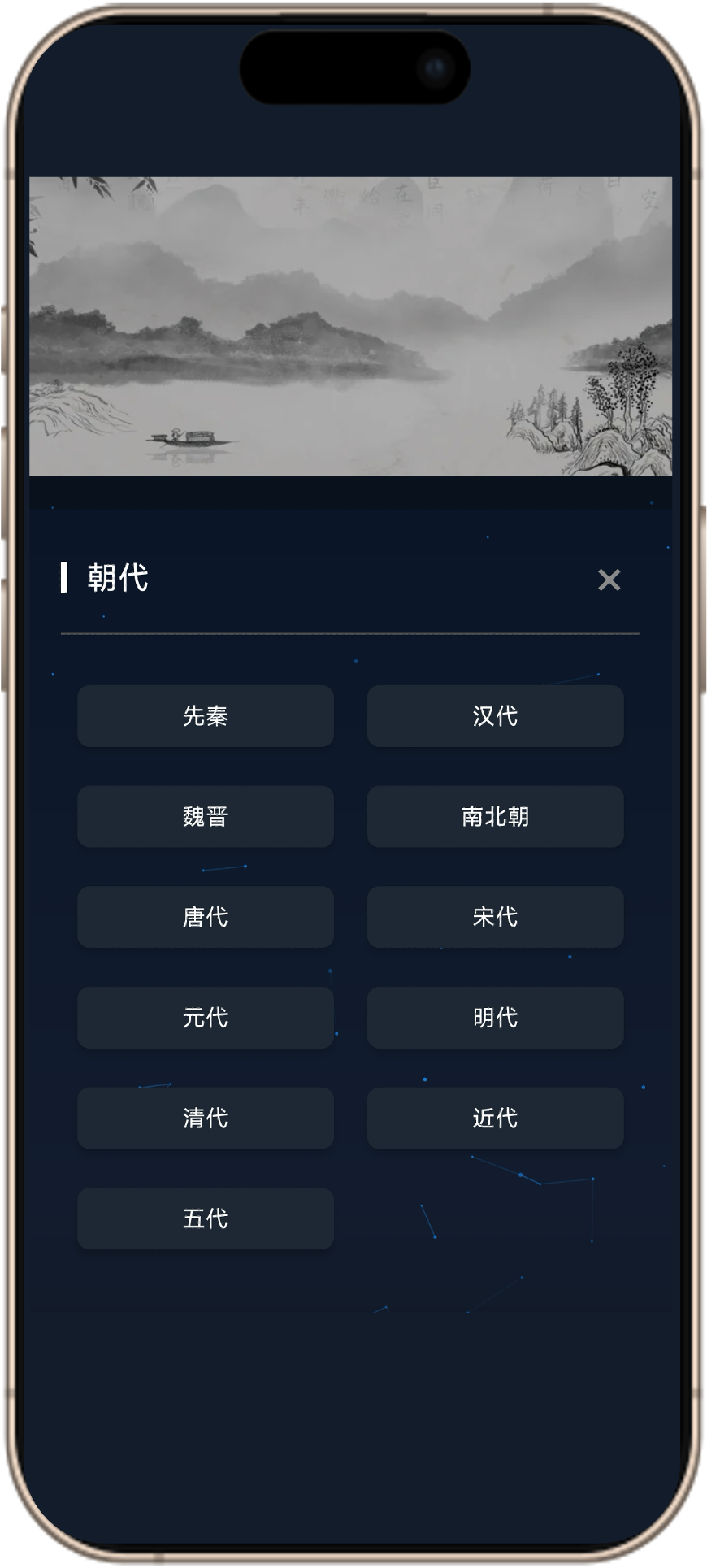}
        \caption{Poetry Library}
        \label{fig:built_in_poetry_library}
    \end{subfigure}
    \hfill
    \begin{subfigure}[b]{0.23\linewidth}
        \centering
        \includegraphics[height=3in]{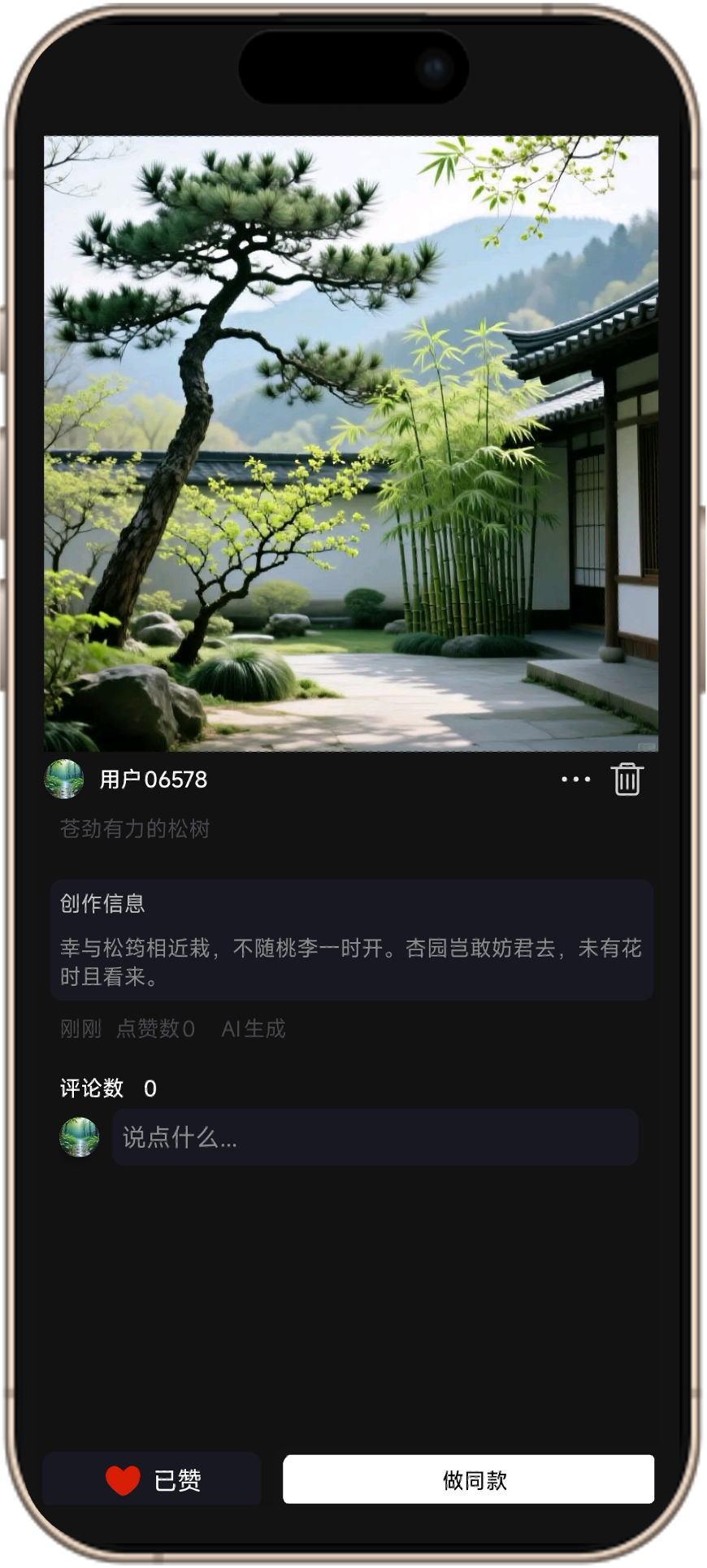}
        \caption{Community}
        \label{fig:community_mojie}
    \end{subfigure}
    \caption{Screenshots of the Mojie application on Android devices, illustrating (a)~the main interface, (b)~offline poetry-to-image generation, (c)~the built-in classical poetry library, and (d)~community sharing and social interaction.}
    \label{fig:mojie_screenshots}
\end{figure}

\subsubsection{iOS application integration and runtime execution}

On the application side, we implement an iOS app in Swift that uses the Core ML framework to load and invoke the converted models. During inference, the input text is first encoded by CLIP, the latent representation is iteratively denoised by U-Net, and the final image is reconstructed by the VAE decoder. This execution order preserves the original structure of the diffusion pipeline while enabling the full three-model generation branch to run natively on iPhone.

Compared with the Android deployment, which partitions execution across MNN and QNN, the iOS implementation uses a unified Core ML runtime for both model invocation and inference. This unified deployment path simplifies software integration and demonstrates that the JuZhou 1.0 diffusion backbone can be migrated to the Apple mobile ecosystem without modifying its core model architecture.

\subsubsection{Experimental results on iPhone}

We evaluate the Apple-side deployment on an iPhone 15 Pro equipped with the A17 Pro chip. Tab.~\ref{tab:hardware_profiling} further profiles the denoising network at 1024 $\times$ 1024 resolution on flagship mobile platforms, showing that JuZhou 1.0 remains executable under practical mobile memory and latency constraints, whereas mainstream SD 1.5 and SDv2.1 baselines fail under the same on-device setting. 
In this implementation, CLIP, U-Net, and VAE all run in FP16 without additional quantization. The total runtime memory footprint is 373\,MB, and the latency of a 4-step diffusion run is 4.25\,s.
These results confirm that the three-model diffusion stack can be executed natively on current iPhone hardware without post-training quantization. Although the present Apple-side validation does not yet include the auxiliary prompt-refinement language model, it demonstrates that the core image-generation path of JuZhou 1.0 is already deployable and operational on iOS devices.

\subsection{Cross-platform implications}
Taken together, the Android and iOS results demonstrate that JuZhou 1.0 can be deployed across heterogeneous mobile ecosystems through platform-specific toolchains while preserving the same high-level model decomposition. Android emphasizes engine-level partitioning between MNN and QNN, whereas iOS relies on a unified Core ML stack.
This cross-platform evidence shown in Fig.~\ref{fig:latency_comparison} is significant for edge-AI deployment because it indicates that the same poetry-to-image architecture can adapt to different mobile hardware and software environments without redesigning the model itself. The Apple-side result further shows that an unquantized FP16 diffusion branch can still achieve practical latency on flagship phones.
\footnotetext{``--'' denotes model failure due to excessive VRAM requirements or extreme latency exceeding practical utility. JuZhou 1.0 metrics cover the denoising network only: MACs 579.65\,G and FLOPs 1161.82\,G. CLIP and VAE overheads are excluded.}

\section{Experiments}
\label{sec:experiments}

We comprehensively evaluate JuZhou 1.0 from both quantitative and qualitative perspectives. The quantitative evaluation measures semantic alignment on the GenEval benchmark against mainstream English-centric baselines (SDv2.1 and SDXL), while the qualitative evaluation examines foundational generation fidelity, distillation efficacy, and native Chinese cultural alignment through systematic visual comparisons. Unless otherwise noted, all GenEval evaluations use the 28-step base model for fair comparison, and all on-device latency measurements use the 4-step distilled model.

\subsection{Experimental Setup}
\label{sec:experimentalSetup}
We adopt a progressive resolution scaling strategy to enable a smooth transition from low- to high-resolution synthesis. 
The training pipeline consists of two phases and three reported stages. The first phase is 28-step foundational training, which includes Stage 1 low-resolution pre-training at $256\times256$ and Stage 2 progressive-resolution training from $512\times512$ to $1024\times1024$. The second phase is Stage 3 DMD2-based step compression, which distills the 28-step model into a 4-step model at $1024\times1024$.

All model training and distillation experiments were conducted on the Sugon K100 AI cluster, with stage-specific resource allocation. The low-resolution pre-training and progressive-resolution foundational training stages used 56 compute nodes with 224 K100 DCUs, while the DMD2 step-compression stage used 40 compute nodes with 160 K100 DCUs.
Quantitative evaluation and baseline comparison are performed in a unified NVIDIA V100 environment, while on-device latency is measured on the target mobile devices.

For quantitative evaluation on the GenEval benchmark, we use our 28-step English base model to ensure a fair comparison with established English-centric baselines, including Stable Diffusion v2.1 (SDv2.1) and SDXL. The detailed configuration of each training stage is reported in Tab.~\ref{tab:training_config}, including resolution, dataset, cluster scale, global batch size, training steps, learning rate, and optimizer settings.

\begin{table}[t]
    \centering
    \caption{Training configuration for each stage of the JuZhou 1.0 multi-stage training framework. All stages are conducted on the Sugon K100 AI cluster. (BS: Global Batch Size, LR: Learning Rate, Opt.: Optimizer)}
    \label{tab:training_config}
    \small
    \setlength{\tabcolsep}{5pt}
    \renewcommand{\arraystretch}{1.15}

    \resizebox{\linewidth}{!}{
    \begin{tabular}{@{}lccccccc@{}}
        \toprule
        \textbf{Stage} 
        & \textbf{Resolution} 
        & \textbf{Data} 
        & \textbf{Nodes $\times$ DCUs}
        & \textbf{BS}
        & \textbf{Steps} 
        & \textbf{LR}
        & \textbf{Opt.} \\
        \midrule
        1: Low-Res Pre-training 
        & $256\times256$ 
        & ImageNet-1K 
        & $56 \times 4$ 
        & 114{,}688 
        & 30K 
        & $3 \times 10^{-4}$ 
        & AdamW \\
        2: Progressive Resolution 
        & $512{\to}1024$ 
        & 9M Chinese pairs 
        & $56 \times 4$ 
        & 2{,}688 
        & 150K 
        & $1 \times 10^{-4}$ 
        & AdamW \\
        3: Step Compression (DMD2) 
        & $1024\times1024$ 
        & 9M Chinese pairs 
        & $40 \times 4$ 
        & 960 
        & 10K 
        & $1 \times 10^{-5}$ 
        & AdamW \\
        \bottomrule
    \end{tabular}
    }

    \renewcommand{\arraystretch}{1.0}
    \setlength{\tabcolsep}{6pt}
\end{table}

\subsection{Quantitative Results}

To evaluate the semantic understanding and compositional reasoning capabilities of the JuZhou~1.0 base model, we conduct a systematic evaluation on the GenEval benchmark~\cite{ghosh2023geneval}. GenEval measures fine-grained text-to-image alignment across several complementary dimensions, including object generation (\textit{Single} and \textit{Two}), numerical reasoning (\textit{Count}), attribute recognition (\textit{Color}), spatial reasoning (\textit{Position}), and attribute--object binding (\textit{Color Attr.}).

As shown in Tab.~\ref{tab:geneval_performance}, JuZhou~1.0 achieves an overall GenEval score of \textbf{0.70} with only \textbf{0.387B} core image-generation backbone parameters. Among all evaluated models, this represents the second-highest overall score, surpassed only by FLUX.1-schnell at 0.71, while JuZhou~1.0 uses approximately $31\times$ fewer parameters. It also substantially outperforms several larger server-oriented models, including SD3-Medium, SDXL, IF-XL, PlayGround~v2.5, Hunyuan-DiT, and both Sana variants. Compared with compact baselines, JuZhou~1.0 improves the overall score from 0.66 for SnapGen and Sana-1.6B to 0.70, while maintaining a highly compact model size. With 0.387B parameters, JuZhou~1.0 is the second-smallest model in the comparison, only slightly larger than the 0.373B SnapGen.

These results are obtained under a substantially constrained deployment setting. JuZhou~1.0 is trained end-to-end on a China-developed accelerator stack and is explicitly designed for offline execution on mobile devices. In contrast, most evaluated baselines contain substantially more parameters and are primarily intended for server-side inference. Among the models listed in Tab.~\ref{tab:geneval_performance}, only JuZhou~1.0 and SnapGen support mobile deployment, while JuZhou~1.0 achieves a notably higher overall GenEval score.

A closer examination of the category-level results further demonstrates that the compact architecture preserves strong compositional and attribute-binding capabilities. JuZhou~1.0 achieves a \textit{Single} score of \textbf{1.00}, tied for the best performance, and a \textit{Two} score of \underline{0.89}, ranking second among all evaluated models and trailing only FLUX.1-schnell. It also obtains a \textit{Color} score of \textbf{0.88}, tied for the best performance, and a \textit{Color Attr.} score of \textbf{0.67}, which is the highest score in the comparison. These results indicate that JuZhou~1.0 can reliably generate single or multiple objects with specified visual attributes and associate colors with the corresponding objects despite its highly compact backbone. In addition, its \textit{Position} score of \textit{0.23} ranks third overall, outperforming all compact and efficient baselines included in the comparison.

JuZhou~1.0 nevertheless exhibits limitations in numerical reasoning. Its \textit{Count} score of 0.51 remains below those of several larger models, such as FLUX.1-dev, FLUX.1-schnell, Hunyuan-DiT, IF-XL, and SD3-Medium, and is also lower than some compact baselines such as SnapGen and Sana variants. This suggests that accurately representing object multiplicity remains challenging under strict model-capacity constraints and constitutes an important direction for future improvement.

Overall, JuZhou~1.0 achieves a strong balance between GenEval performance, compact model scale, and practical offline deployment. With a 0.387B core image-generation backbone, it attains the second-highest overall GenEval score in the comparison while supporting on-device inference. These results suggest that compact T2I models trained on domestic accelerator stacks can retain competitive semantic alignment and compositional ability without relying on server-scale architectures.

\begin{table}[t]
    \centering
    \caption{Quantitative evaluation on the GenEval benchmark. All models are evaluated at $1024 \times 1024$ resolution. Params and overall score are placed side by side to highlight the performance--parameter trade-off. The ``Mobile'' column indicates whether a model has reported or validated smartphone-side image-generation deployment, rather than merely compact parameter count. JuZhou~1.0 achieves an overall score of 0.70 with a 0.387B-parameter image-generation backbone, showing a favorable accuracy--efficiency trade-off compared with larger baselines such as Hunyuan-DiT, SD3-Medium, and IF-XL.}
    \label{tab:geneval_performance}
    \resizebox{\textwidth}{!}{%

    \begin{tabular}{lccccccccc}
        \toprule
        \textbf{Method} 
        & \textbf{Params$\downarrow$} 
        & \textbf{Overall$\uparrow$} 
        & \textbf{Mobile} 
        & \textbf{Single$\uparrow$} 
        & \textbf{Two$\uparrow$} 
        & \textbf{Count$\uparrow$} 
        & \textbf{Color$\uparrow$} 
        & \textbf{Pos.$\uparrow$} 
        & \textbf{Color Attr.$\uparrow$} \\
        \midrule
        \midrule
    
        \multicolumn{10}{l}{\textit{Large-scale baselines}} \\
        \midrule
    
        LUMINA-Next~\cite{zhuo2024luminanext}
        & 2.0B
        & 0.46
        & {\color{red}\ding{55}}
        & 0.92
        & 0.46
        & 0.48
        & 0.70
        & 0.09
        & 0.13 \\
    
        SD3-Medium~\cite{esser2024scaling}
        & 2.0B
        & 0.62
        & {\color{red}\ding{55}}
        & \textit{0.98}
        & 0.74
        & 0.63
        & 0.67
        & \textbf{0.34}
        & 0.36 \\
    
        SDXL~\cite{podell2023sdxl}
        & 2.6B
        & 0.55
        & {\color{red}\ding{55}}
        & \textit{0.98}
        & 0.74
        & 0.39
        & \underline{0.85}
        & 0.15
        & 0.23 \\
    
        PlayGround v2.5~\cite{li2024playgroundv25}
        & 2.6B
        & 0.56
        & {\color{red}\ding{55}}
        & \textit{0.98}
        & 0.77
        & 0.52
        & \textit{0.84}
        & 0.11
        & 0.17 \\
    
        IF-XL~\cite{deepfloyd2023if}
        & 4.3B
        & 0.61
        & {\color{red}\ding{55}}
        & 0.97
        & 0.74
        & 0.66
        & 0.81
        & 0.13
        & 0.35 \\
    
        FLUX.1-dev~\cite{flux2024}
        & 12.0B
        & \textit{0.67}
        & {\color{red}\ding{55}}
        & \underline{0.99}
        & 0.81
        & \textbf{0.79}
        & 0.74
        & 0.20
        & \textit{0.47} \\
    
        FLUX.1-schnell~\cite{flux2024}
        & 12.0B
        & \textbf{0.71}
        & {\color{red}\ding{55}}
        & \underline{0.99}
        & \textbf{0.92}
        & \underline{0.73}
        & 0.78
        & \underline{0.28}
        & \underline{0.54} \\
    
        \midrule
        \multicolumn{10}{l}{\textit{Compact and efficient baselines}} \\
        \midrule
    
        SnapGen~\cite{hu2024snapgen}
        & \textbf{0.373B}
        & 0.66
        & {\color{green}\ding{51}}
        & \textbf{1.00}
        & \textit{0.84}
        & 0.60
        & \textbf{0.88}
        & 0.18
        & 0.45 \\
    
        Sana-0.6B~\cite{xie2024sana}
        & \textit{0.6B}
        & 0.64
        & {\color{red}\ding{55}}
        & \underline{0.99}
        & 0.76
        & 0.64
        & \textbf{0.88}
        & 0.18
        & 0.39 \\
    
        SDv2.1~\cite{stability2022sd21}
        & 1.3B
        & 0.50
        & {\color{red}\ding{55}}
        & \textit{0.98}
        & 0.51
        & 0.44
        & \underline{0.85}
        & 0.07
        & 0.17 \\
    
        Hunyuan-DiT~\cite{li2024hunyuandit}
        & 1.5B
        & 0.63
        & {\color{red}\ding{55}}
        & 0.97
        & 0.77
        & \textit{0.71}
        & \textbf{0.88}
        & 0.13
        & 0.30 \\
    
        Sana-1.6B~\cite{xie2024sana}
        & 1.6B
        & 0.66
        & {\color{red}\ding{55}}
        & \underline{0.99}
        & 0.77
        & 0.62
        & \textbf{0.88}
        & 0.21
        & \textit{0.47} \\
    
        \midrule
        \rowcolor{DeepTeal!15}
        \textbf{Ours}
        & \underline{0.387B}
        & \underline{0.70}
        & {\color{green}\ding{51}}
        & \textbf{1.00}
        & \underline{0.89}
        & 0.51
        & \textbf{0.88}
        & \textit{0.23}
        & \textbf{0.67} \\
    
        \bottomrule
    \end{tabular}%

    }
\end{table}

\subsection{Qualitative Results}

We conduct qualitative evaluations along three complementary axes: foundational generation quality compared against the Stable Diffusion family, the efficacy of our 4-step distillation in preserving visual fidelity, and the model's capability to capture native Chinese cultural semantics through classical poetry generation.

\begin{figure}[htbp]
\centering
\small
\setlength{\tabcolsep}{2pt}
\begin{tabular}{ccccc}
\textbf{Ours (28 steps)} & \textbf{SD1.5 (40 steps)} & \textbf{SD2.1 (40 steps)} & \textbf{SDXL (40 steps)} & \textbf{SD3.5-L (28 steps)} \\

\includegraphics[width=0.19\linewidth]{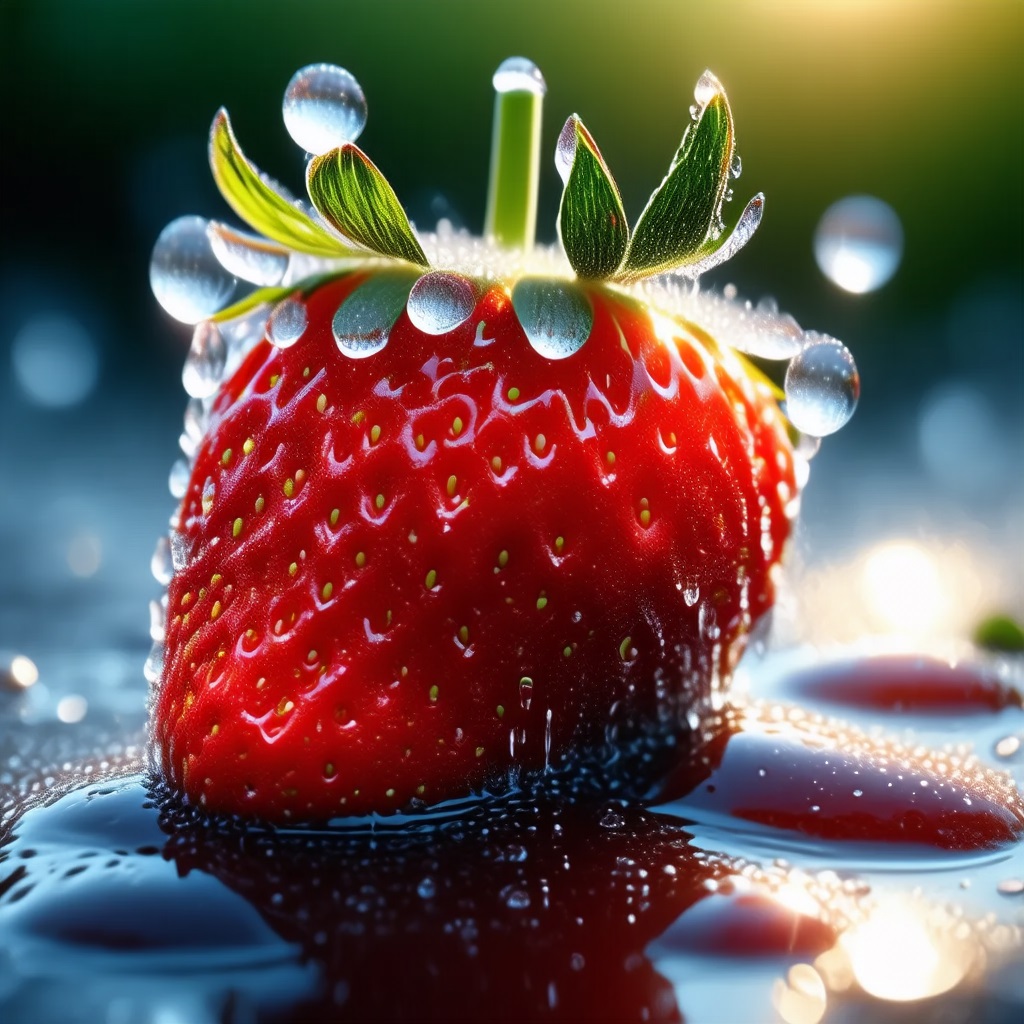} &
\includegraphics[width=0.19\linewidth]{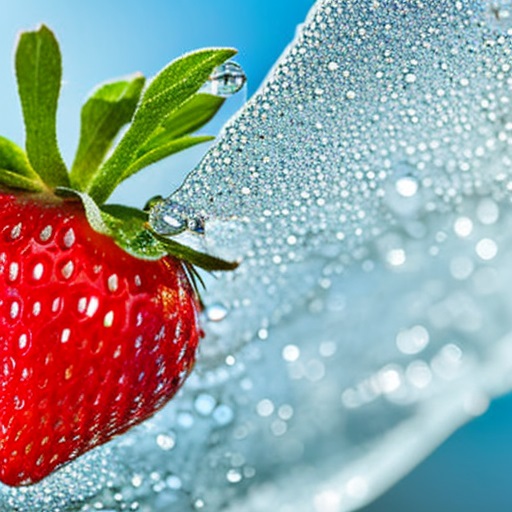} &
\includegraphics[width=0.19\linewidth]{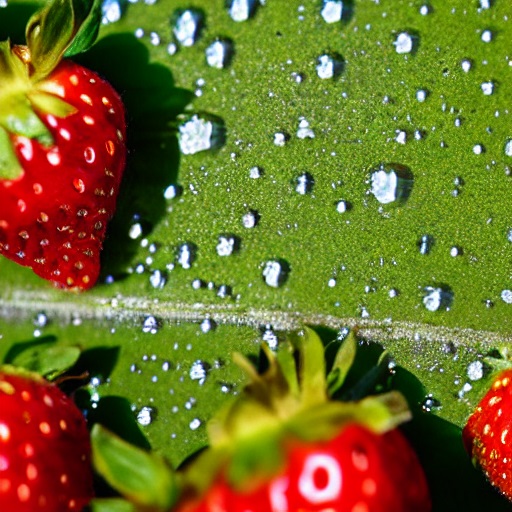} &
\includegraphics[width=0.19\linewidth]{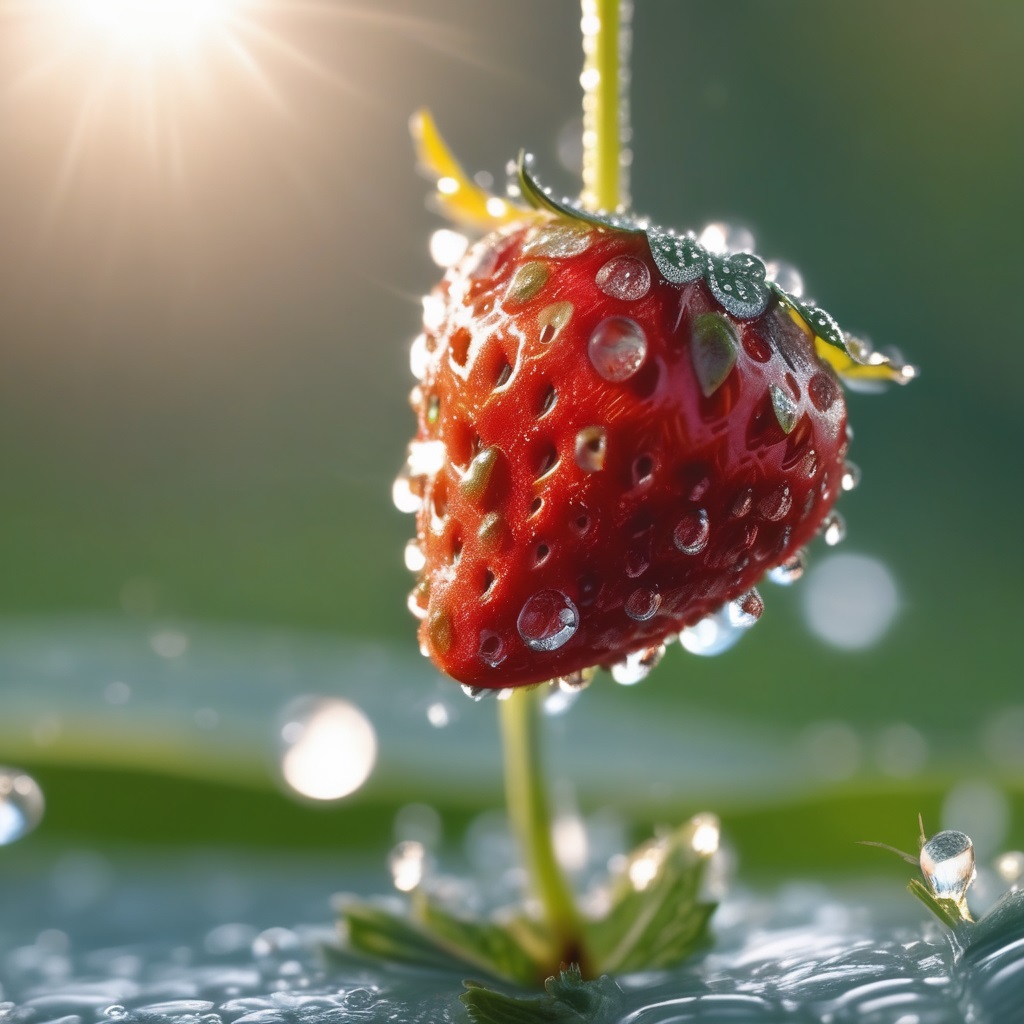} &
\includegraphics[width=0.19\linewidth]{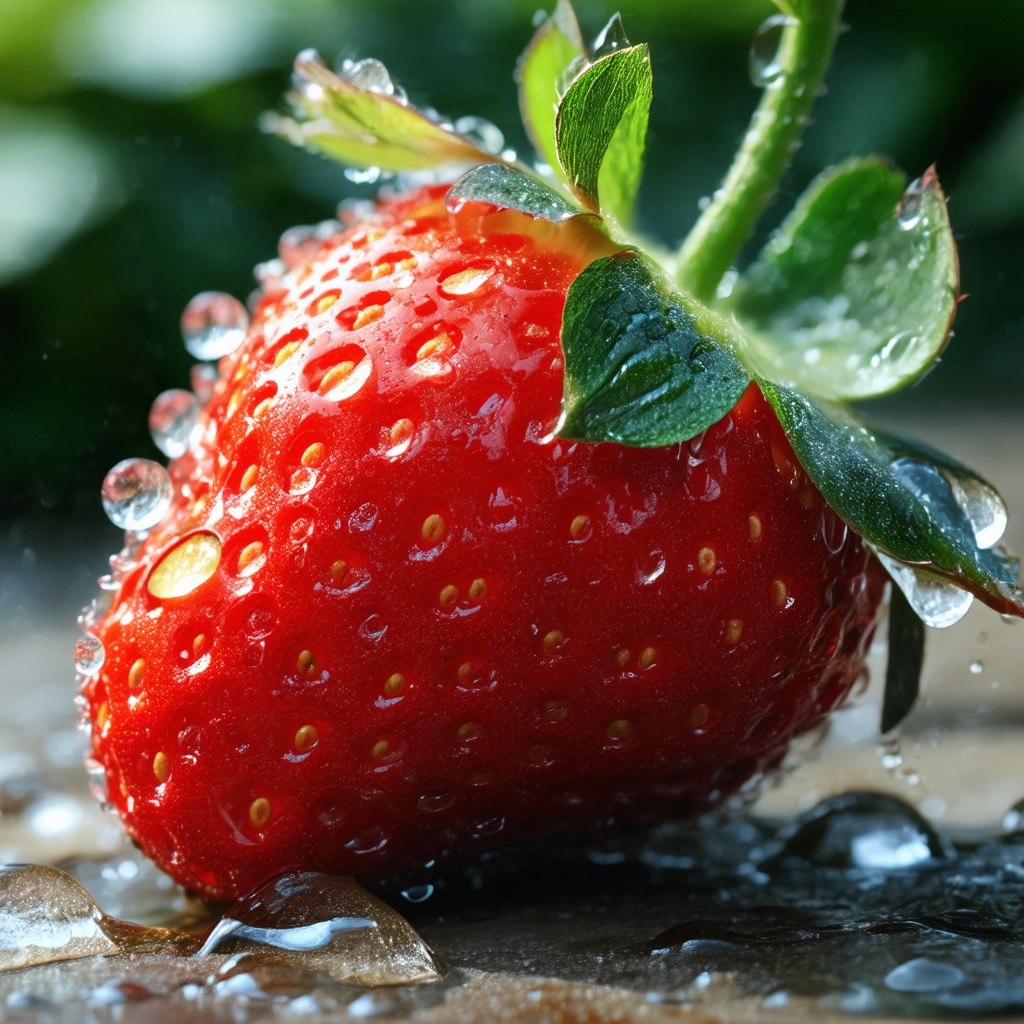} \\
\multicolumn{5}{p{0.98\linewidth}}{\centering\footnotesize (a) Macro shot of a dew-covered strawberry, crystal water droplets, soft morning sun, 8k resolution, photorealistic.} \\[1pt]

\includegraphics[width=0.19\linewidth]{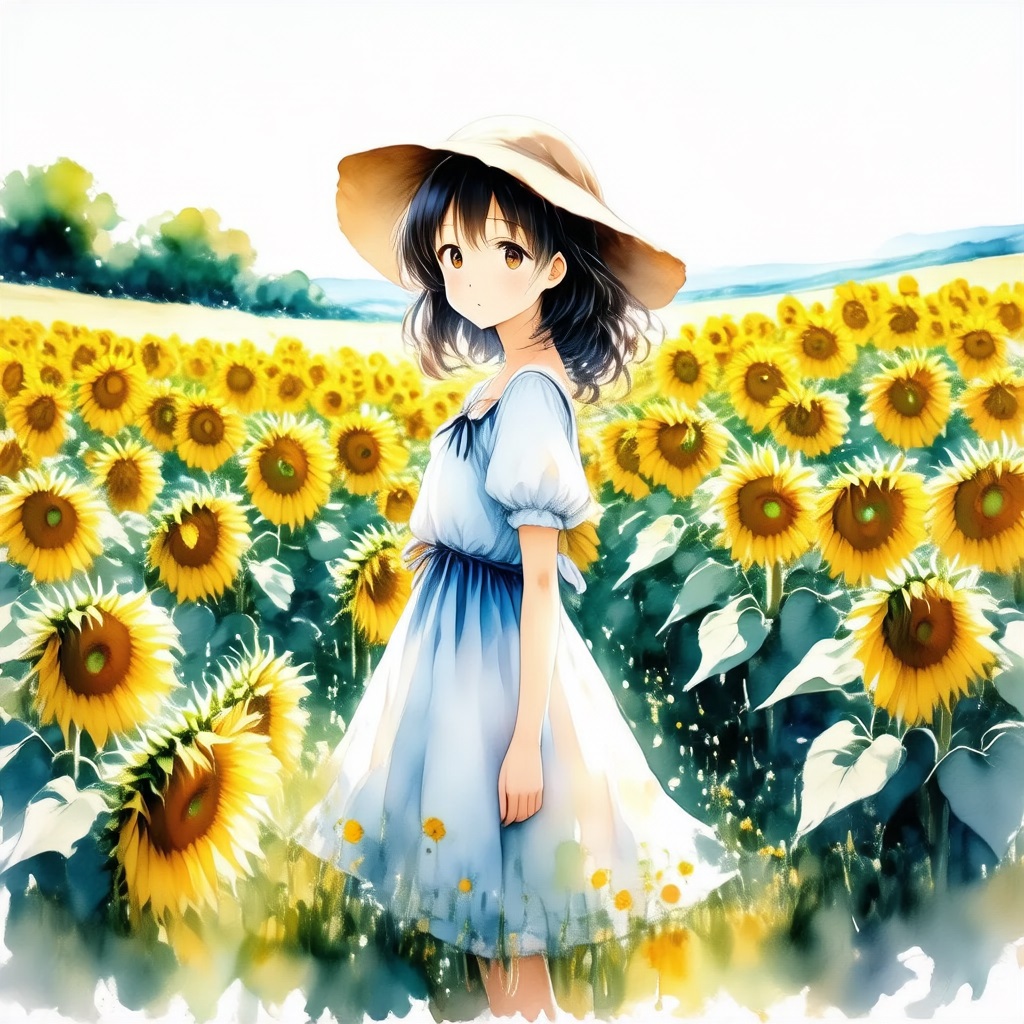} &
\includegraphics[width=0.19\linewidth]{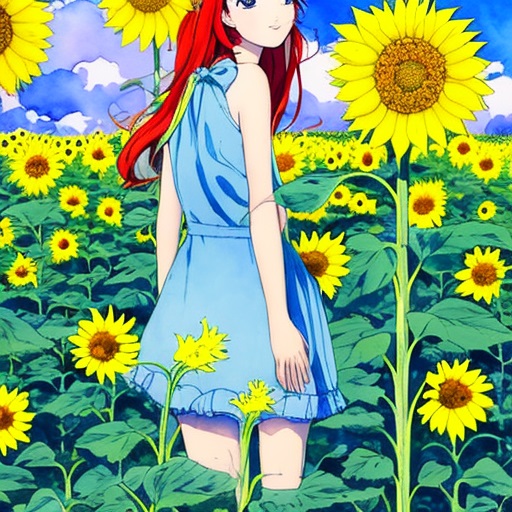} &
\includegraphics[width=0.19\linewidth]{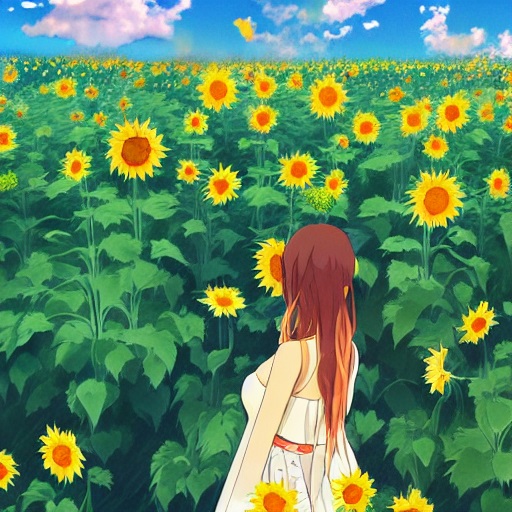} &
\includegraphics[width=0.19\linewidth]{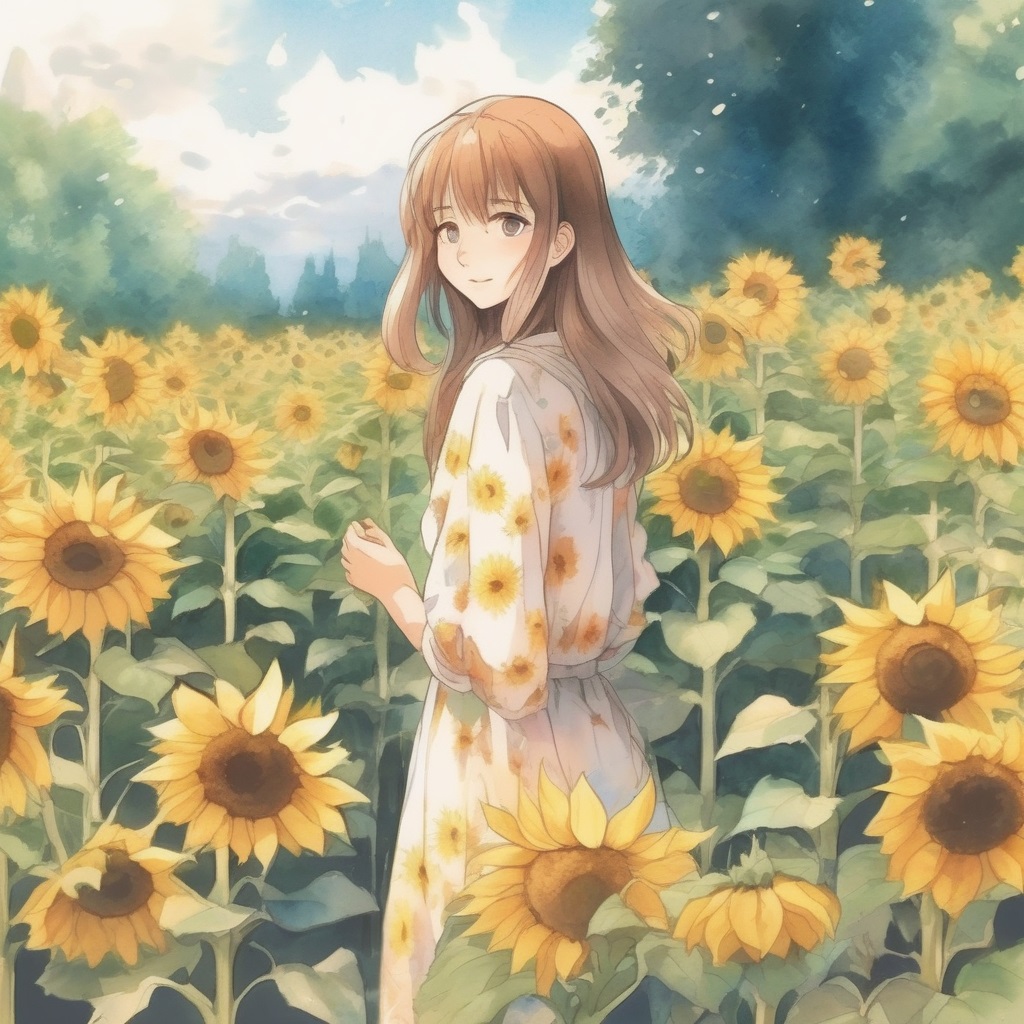} &
\includegraphics[width=0.19\linewidth]{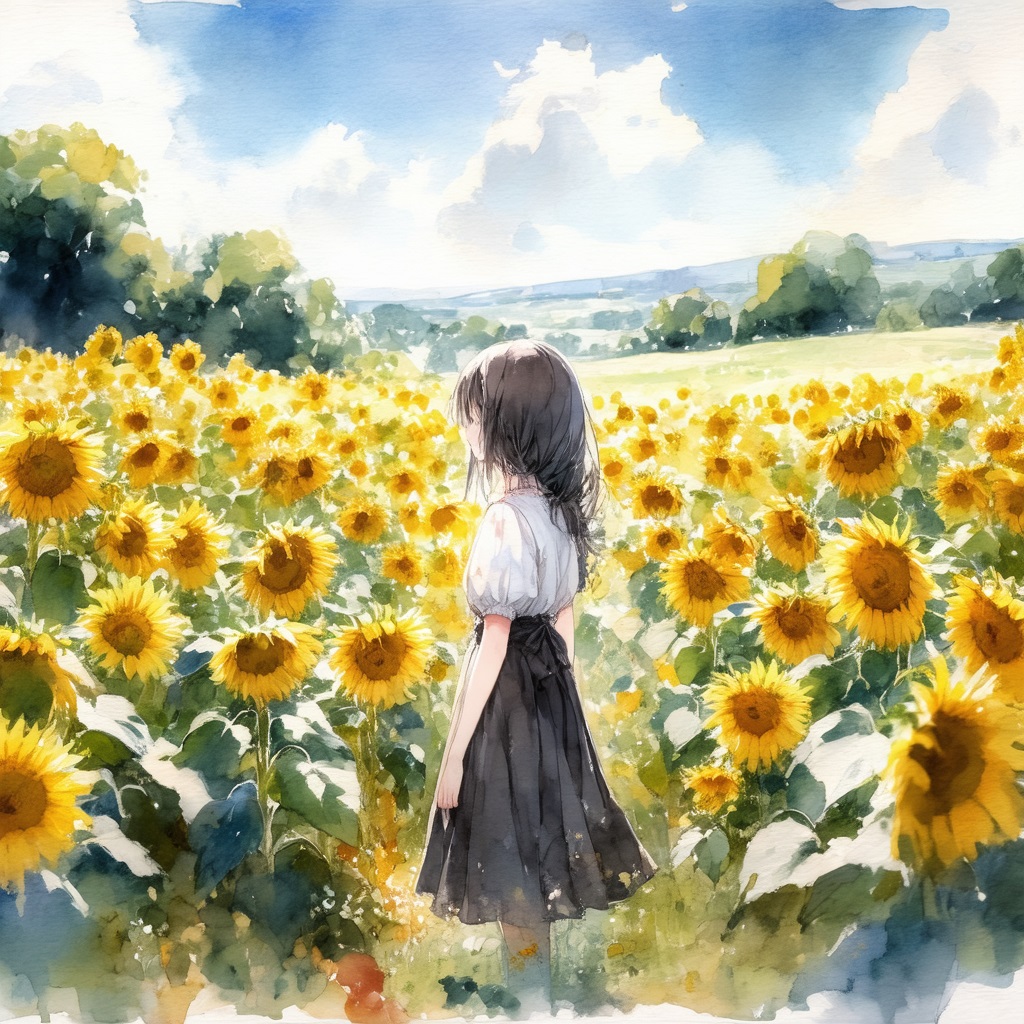} \\
\multicolumn{5}{p{0.98\linewidth}}{\centering\footnotesize (b) A beautiful anime girl standing in a field of sunflowers, soft watercolor anime style, hand-drawn animation aesthetics, dreamy mood.} \\[1pt]

\includegraphics[width=0.19\linewidth]{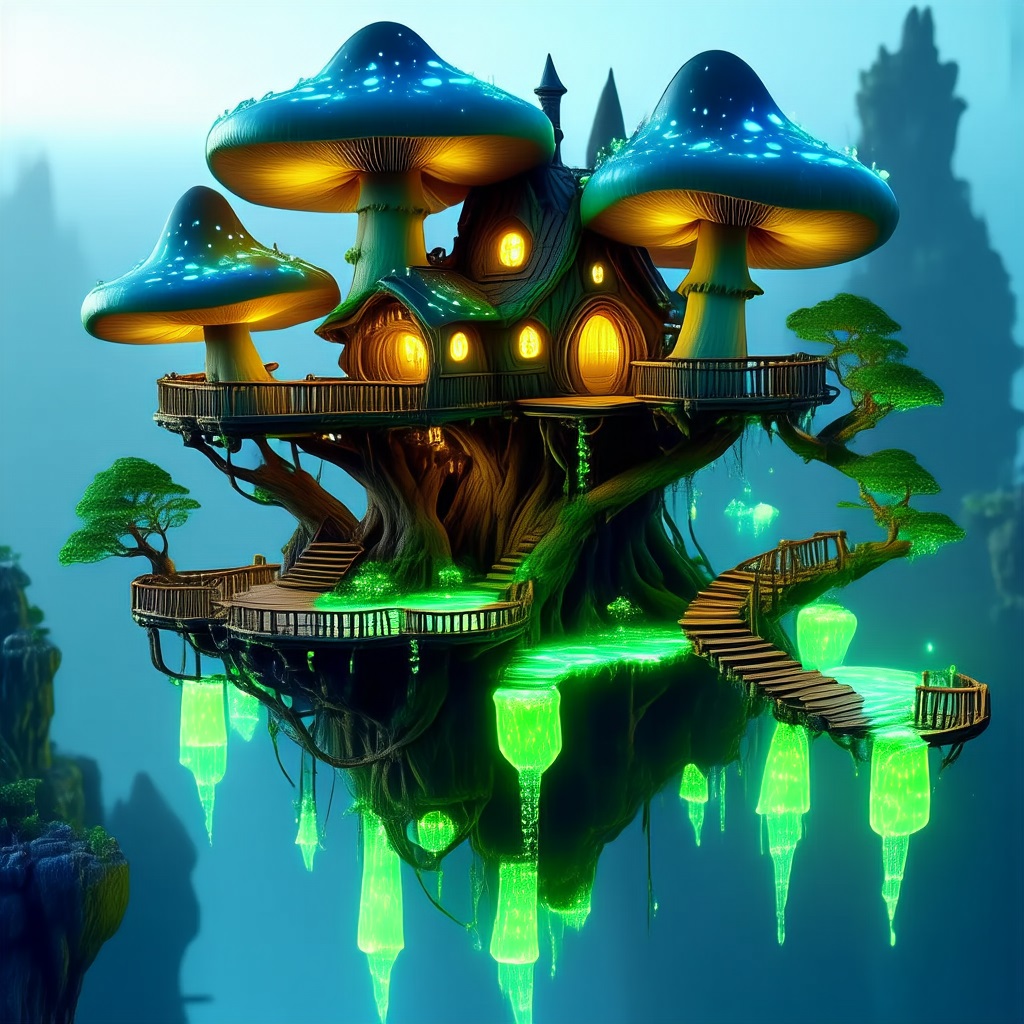} &
\includegraphics[width=0.19\linewidth]{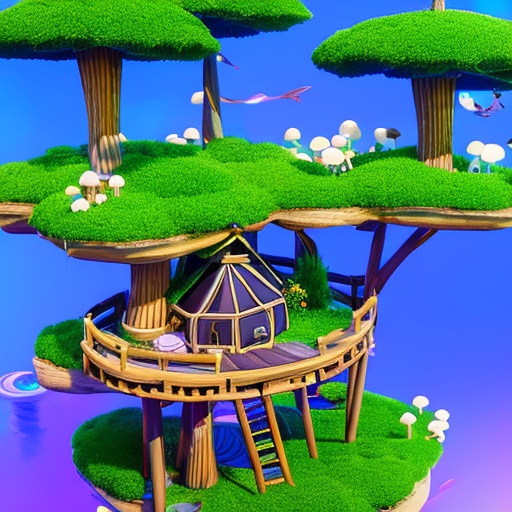} &
\includegraphics[width=0.19\linewidth]{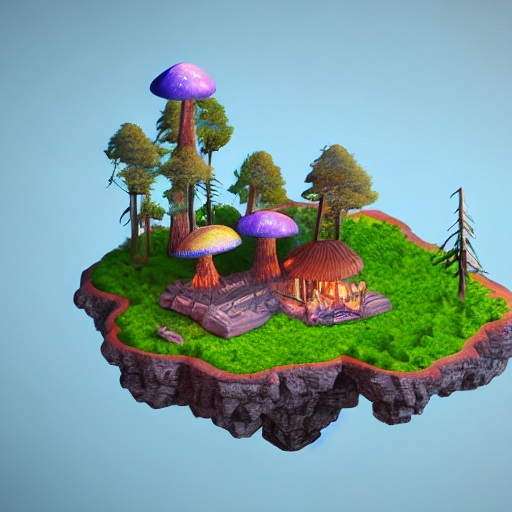} &
\includegraphics[width=0.19\linewidth]{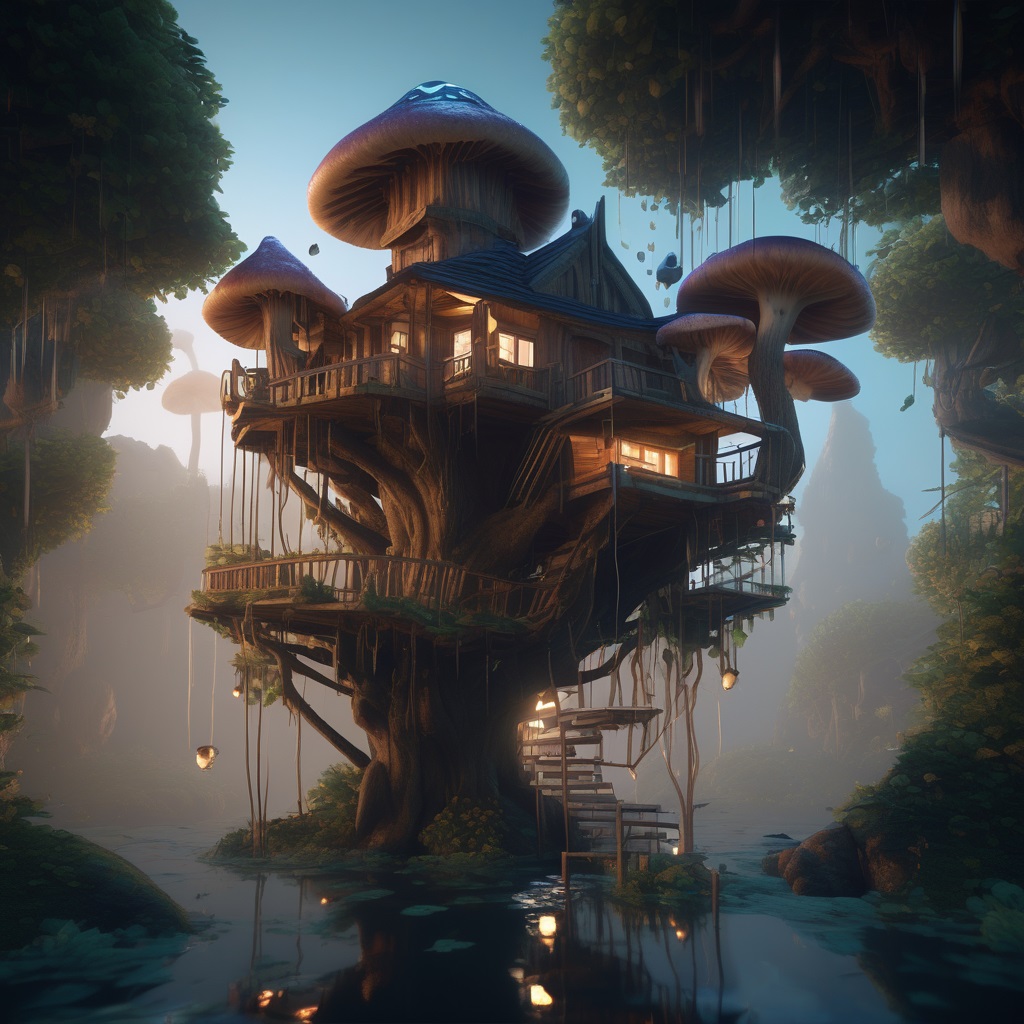} &
\includegraphics[width=0.19\linewidth]{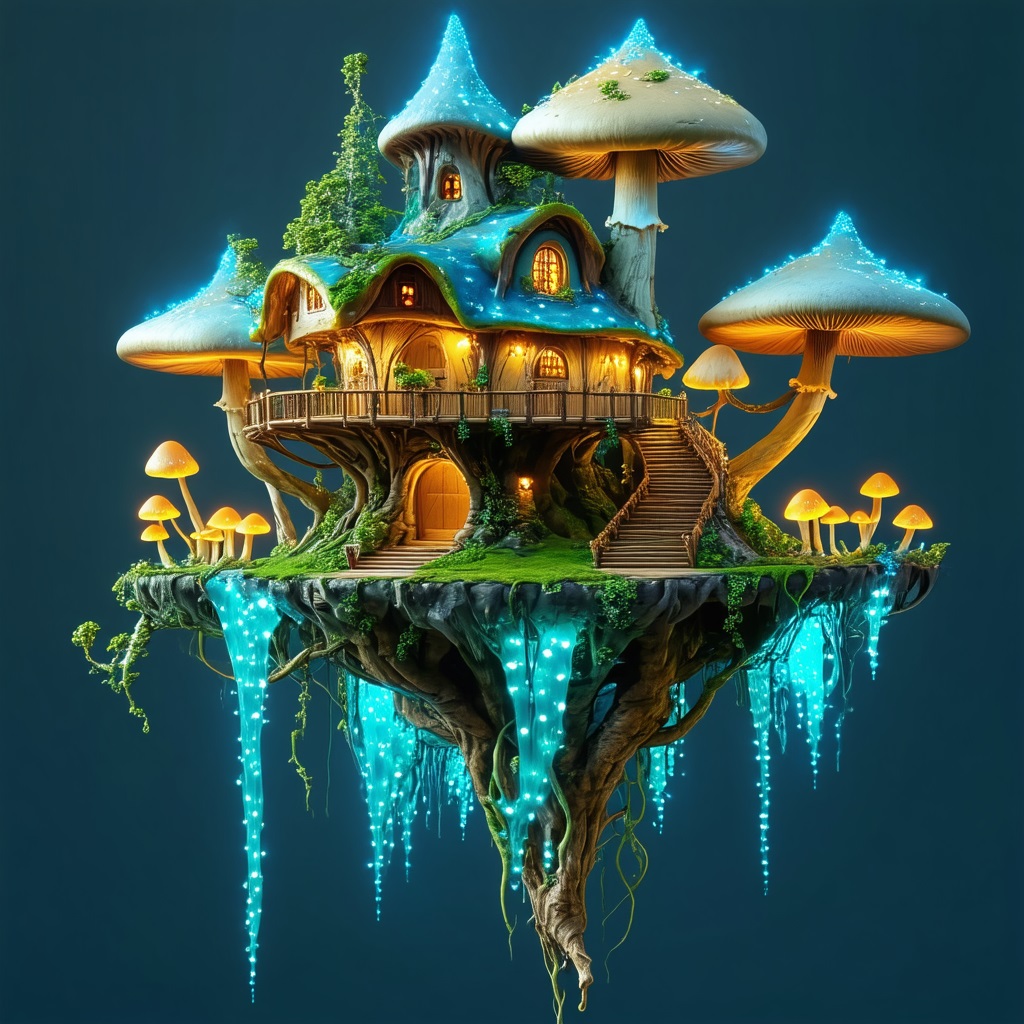} \\
\multicolumn{5}{p{0.98\linewidth}}{\centering\footnotesize (c) An isometric view of a magical floating treehouse, bioluminescent mushrooms, fantasy art, Unreal Engine 5 render.} \\[1pt]

\includegraphics[width=0.19\linewidth]{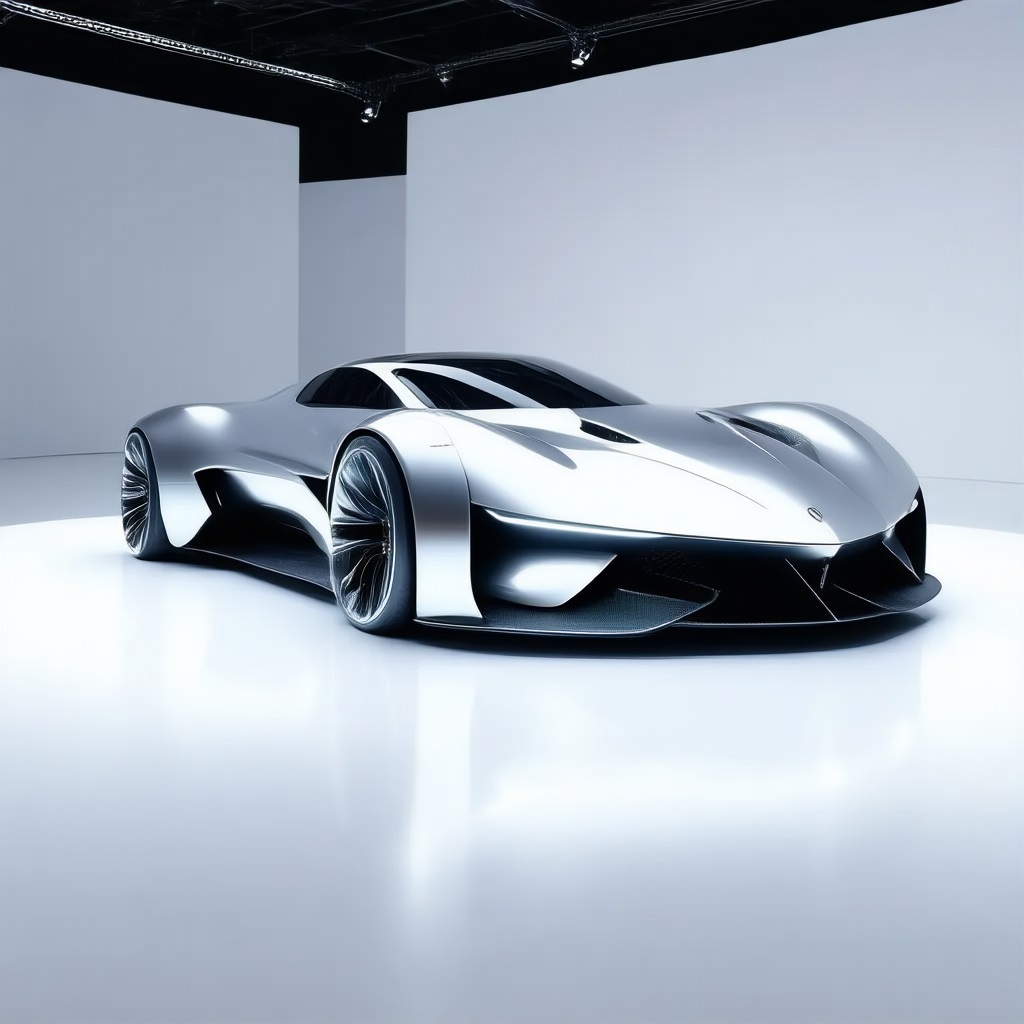} &
\includegraphics[width=0.19\linewidth]{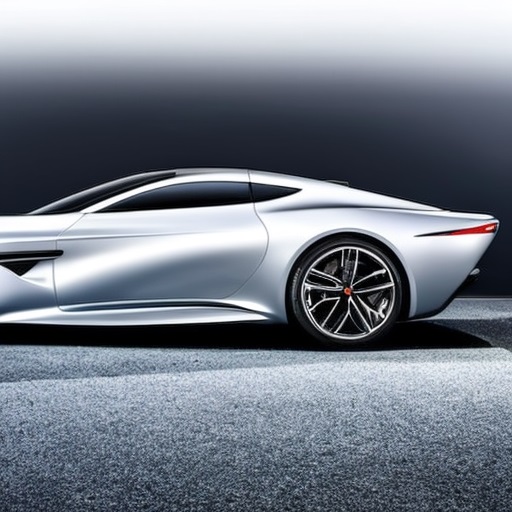} &
\includegraphics[width=0.19\linewidth]{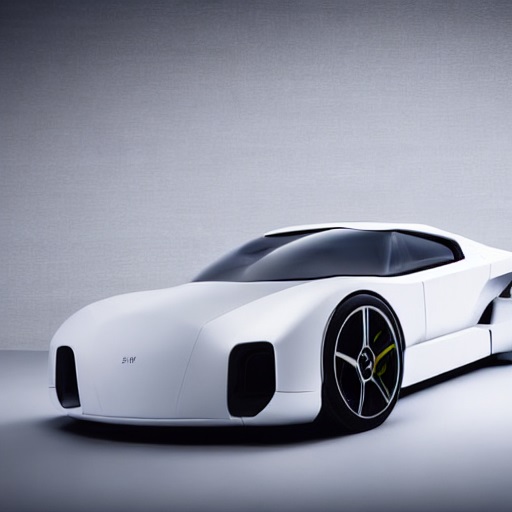} &
\includegraphics[width=0.19\linewidth]{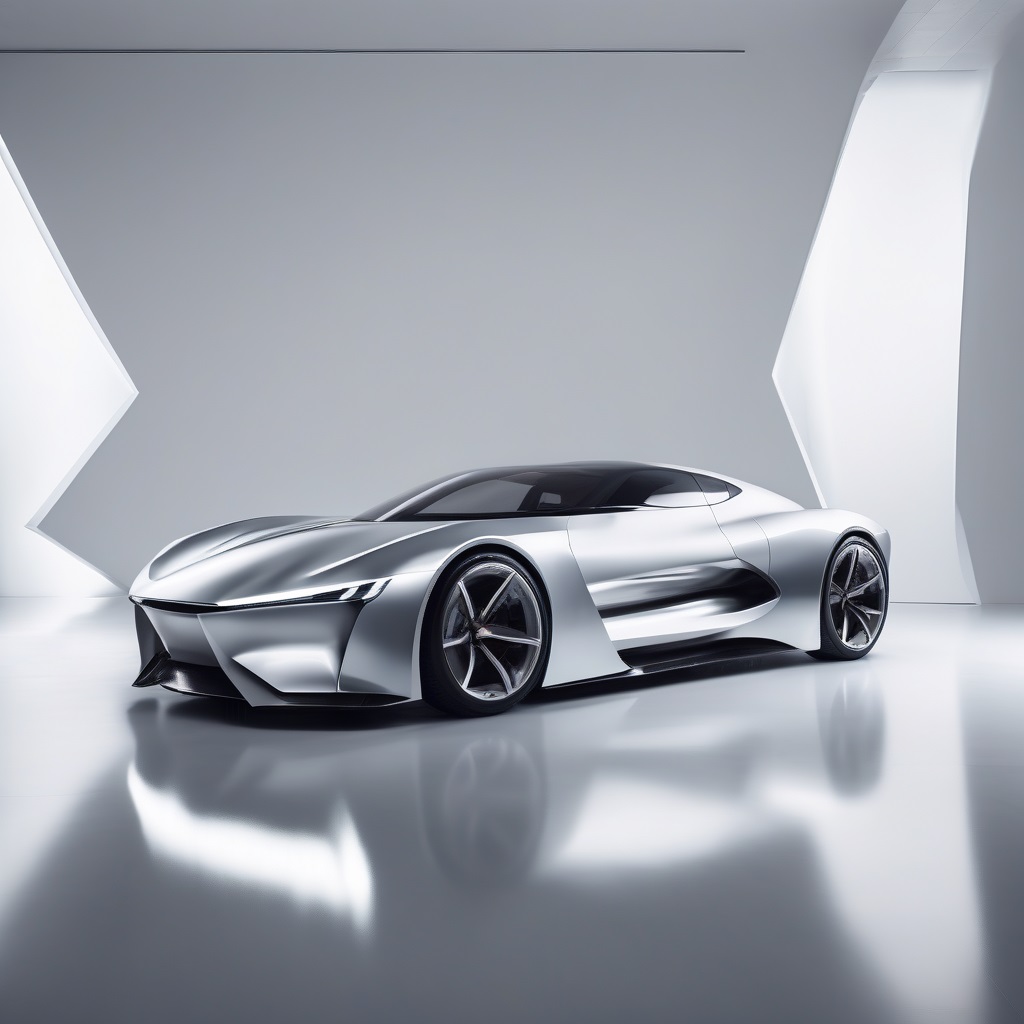} &
\includegraphics[width=0.19\linewidth]{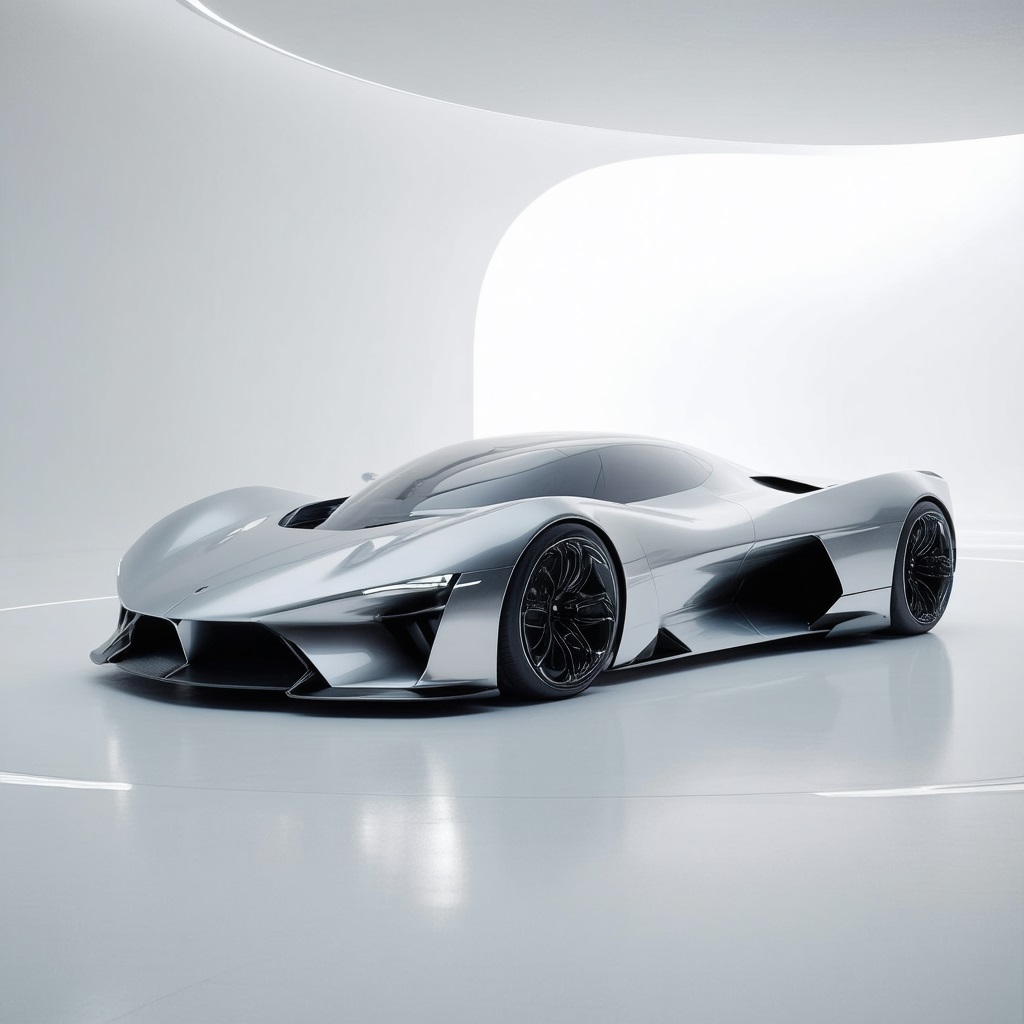} \\
\multicolumn{5}{p{0.98\linewidth}}{\centering\footnotesize (d) A futuristic sports car on a minimalist white stage, sleek silver body, bright studio spotlights, automotive photography.} \\[1pt]

\includegraphics[width=0.19\linewidth]{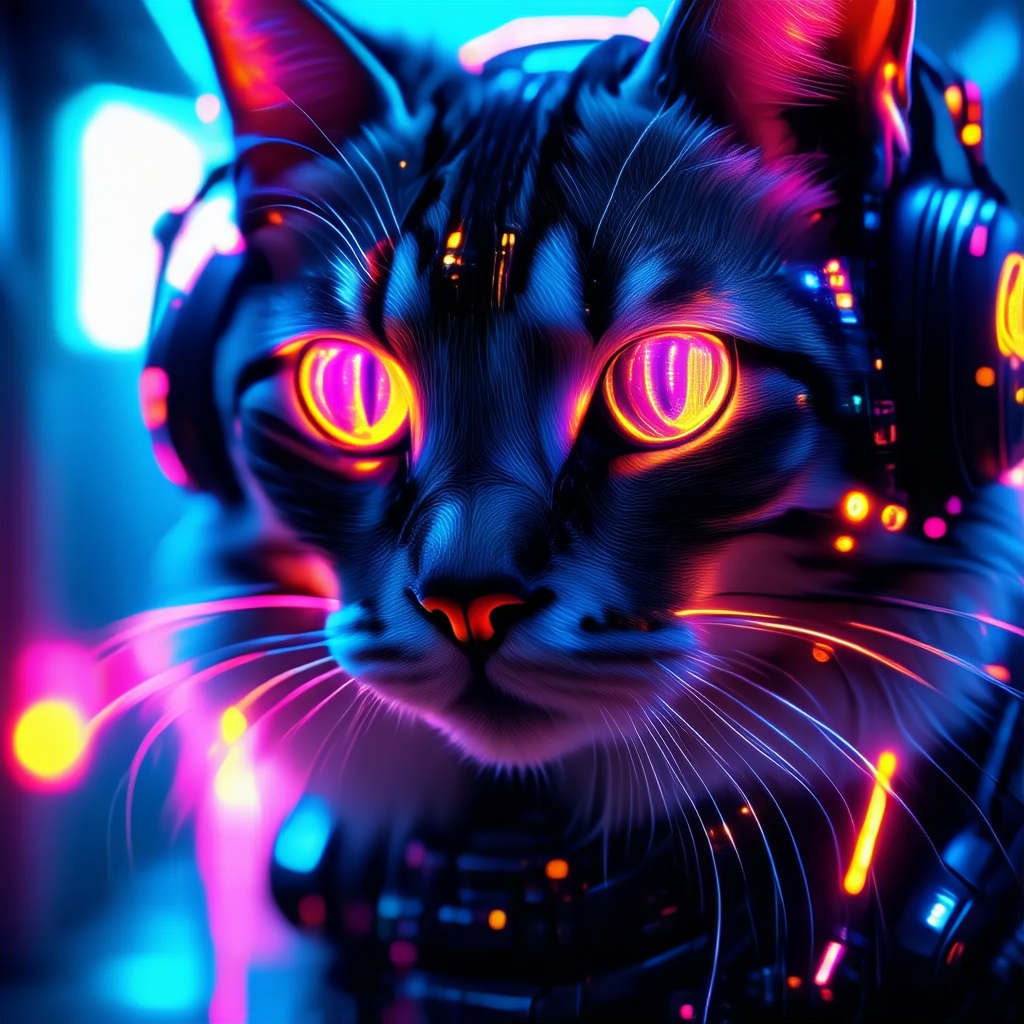} &
\includegraphics[width=0.19\linewidth]{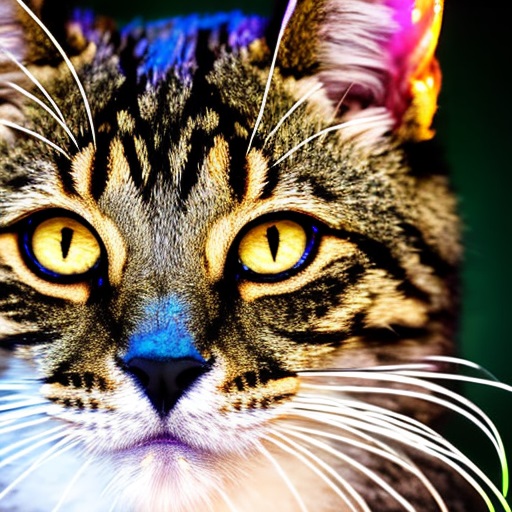} &
\includegraphics[width=0.19\linewidth]{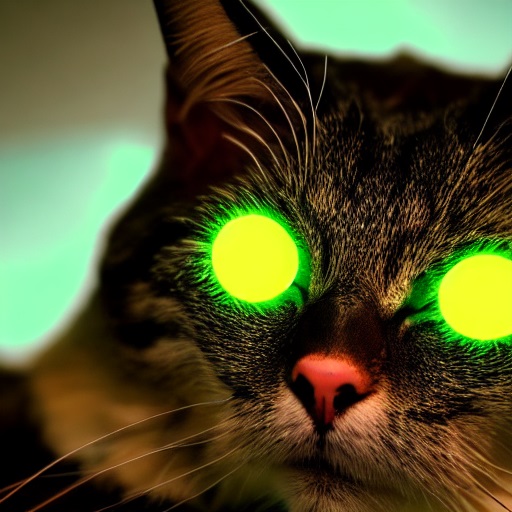} &
\includegraphics[width=0.19\linewidth]{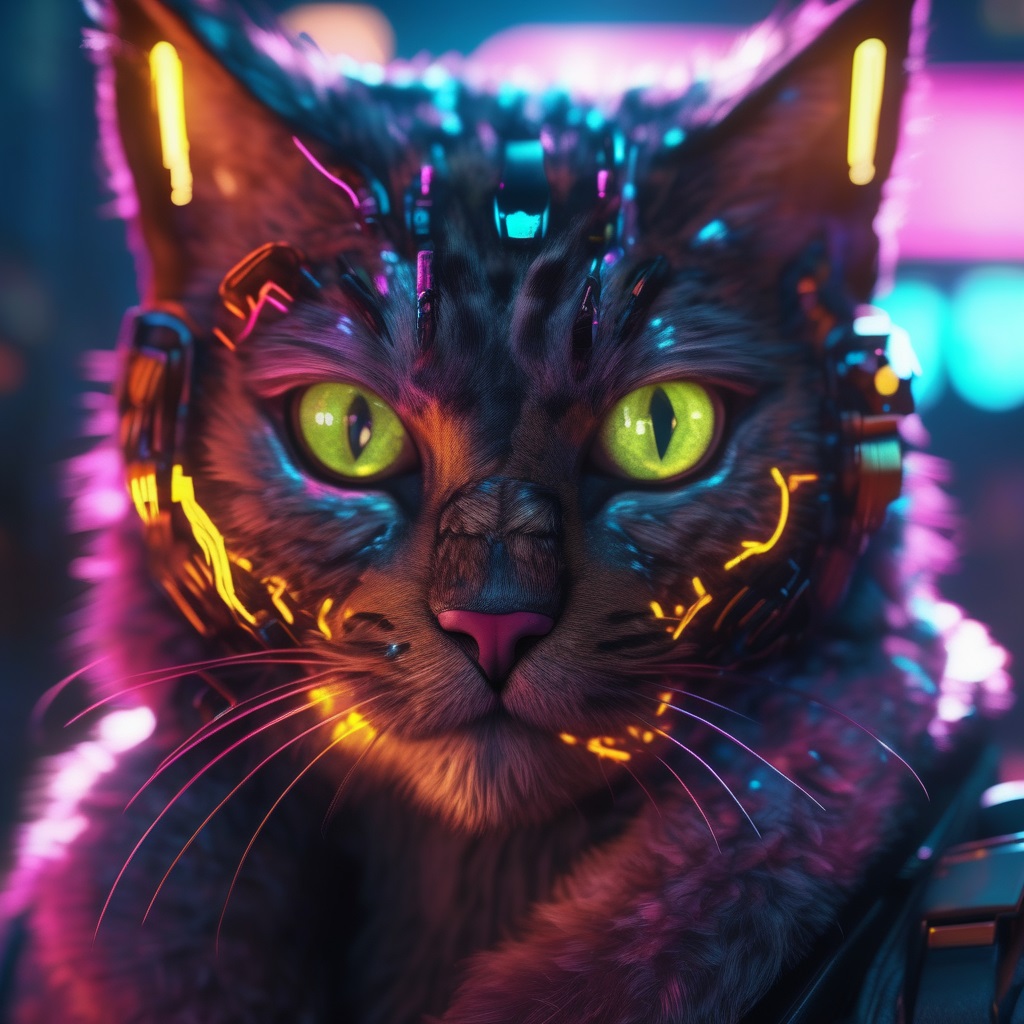} &
\includegraphics[width=0.19\linewidth]{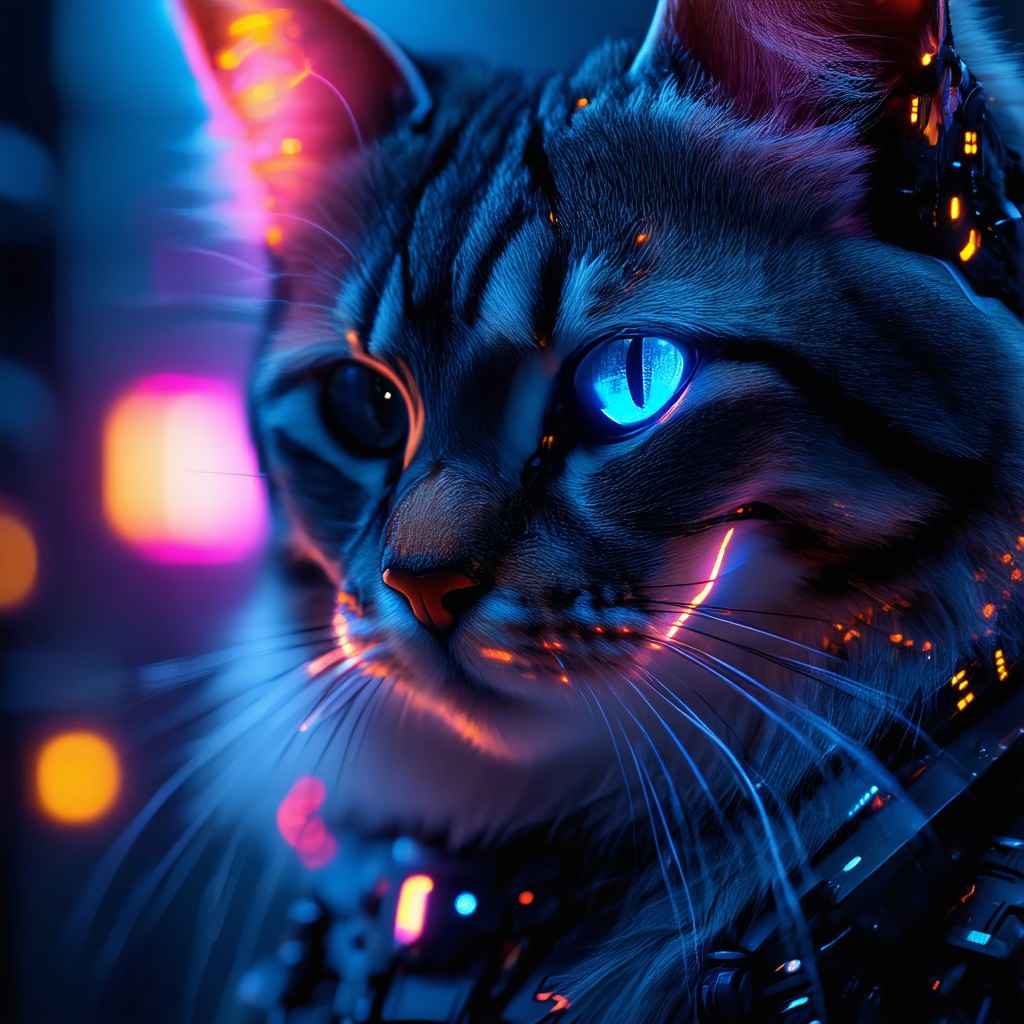} \\
\multicolumn{5}{p{0.98\linewidth}}{\centering\footnotesize (e) A striking close-up portrait of a cyberpunk cat, highly detailed fur, glowing neon reflections in its eyes, cinematic lighting, bokeh background, 8k resolution, photorealistic.} \\[1pt]

\end{tabular}
\caption{Qualitative comparison of $1024\times1024$ images generated by our 28-step base model against the Stable Diffusion family (SD\,1.5, SD\,2.1, SDXL, SD\,3.5-L) across five diverse prompts. Despite its 0.385B parameter budget, JuZhou 1.0 consistently achieves competitive visual fidelity in lighting, texture, and compositional structure.}
\label{fig:foundational_fig}
\end{figure}

\subsubsection{Foundational Generation Quality}
We assess overall visual fidelity by comparing images generated by our 28-step base model against Stable Diffusion variants (v1.5, v2.1, XL, and 3.5-L) using identical English prompts.
As illustrated in Fig.~\ref{fig:foundational_fig}, despite its significantly smaller parameter count, JuZhou 1.0 produces images with complex lighting, fine-grained textures, and strong 3D structural consistency, achieving perceptual quality comparable to---and occasionally surpassing---that of considerably larger models.

\begin{figure}[htbp]
\centering
\setlength{\tabcolsep}{1.2pt}
\renewcommand{\arraystretch}{1.05}

\newcommand{\cmpimg}[1]{%
\includegraphics[width=0.194\linewidth,height=0.118\textheight,keepaspectratio]{#1}
}

\begin{tabular}{ccccc}
\textbf{Orig. 4 steps} & \textbf{Orig. 8 steps} & \textbf{Orig. 16 steps} & \textbf{Orig. 28 steps} & \textbf{Ours (4 steps)} \\[3pt]

\cmpimg{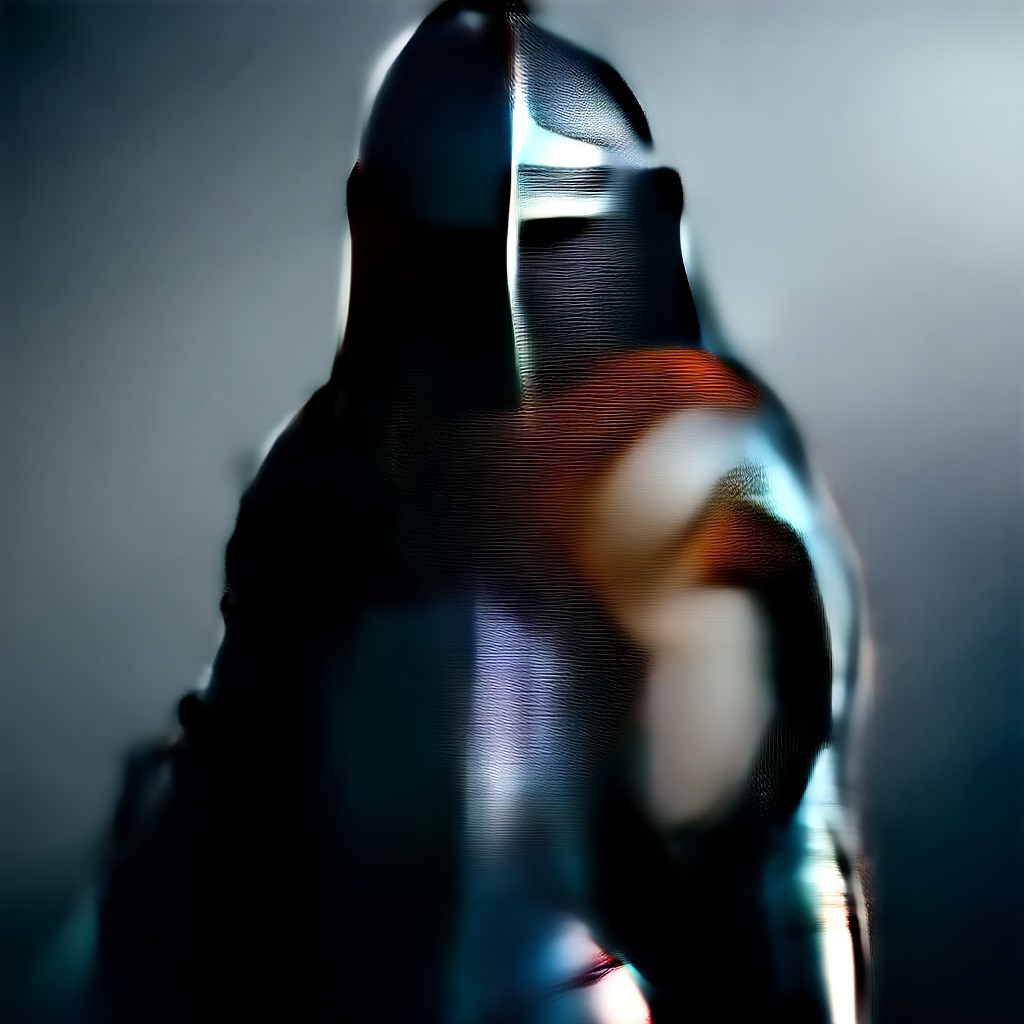} &
\cmpimg{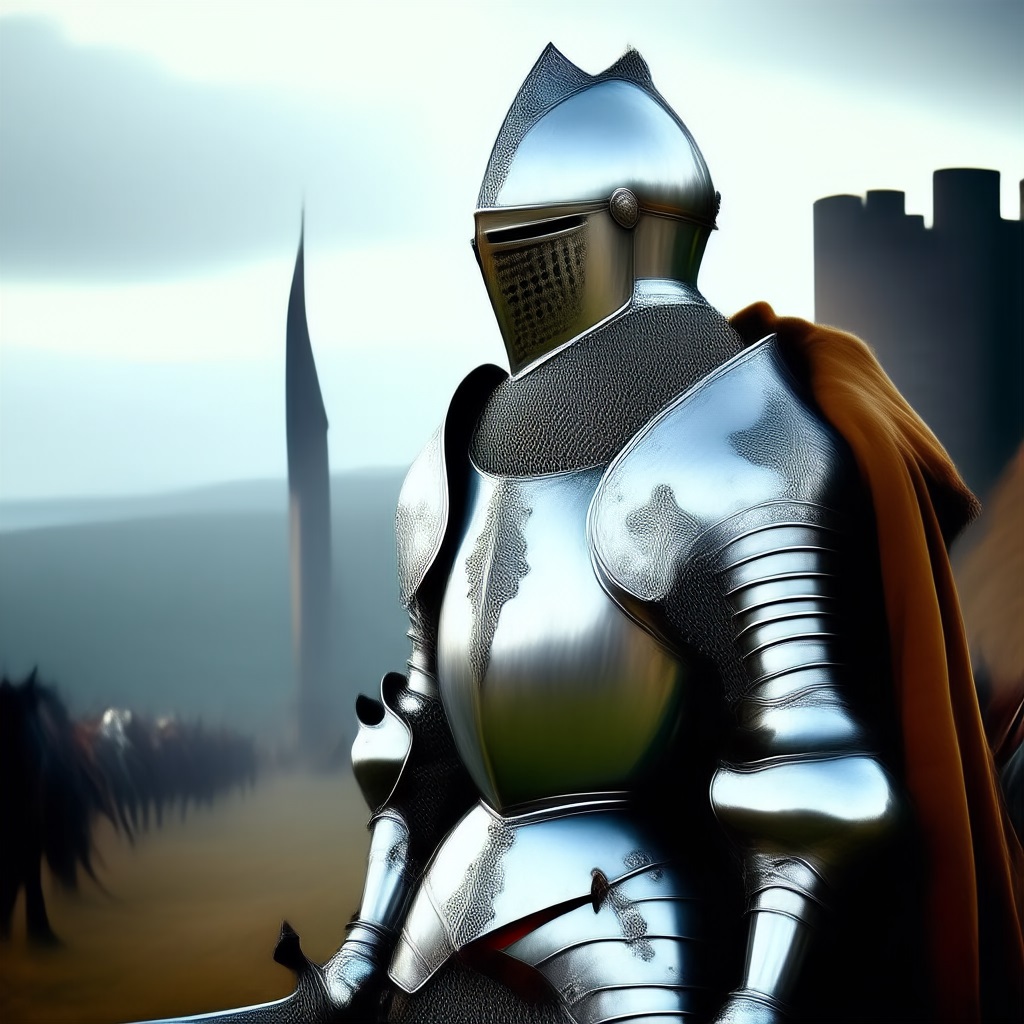} &
\cmpimg{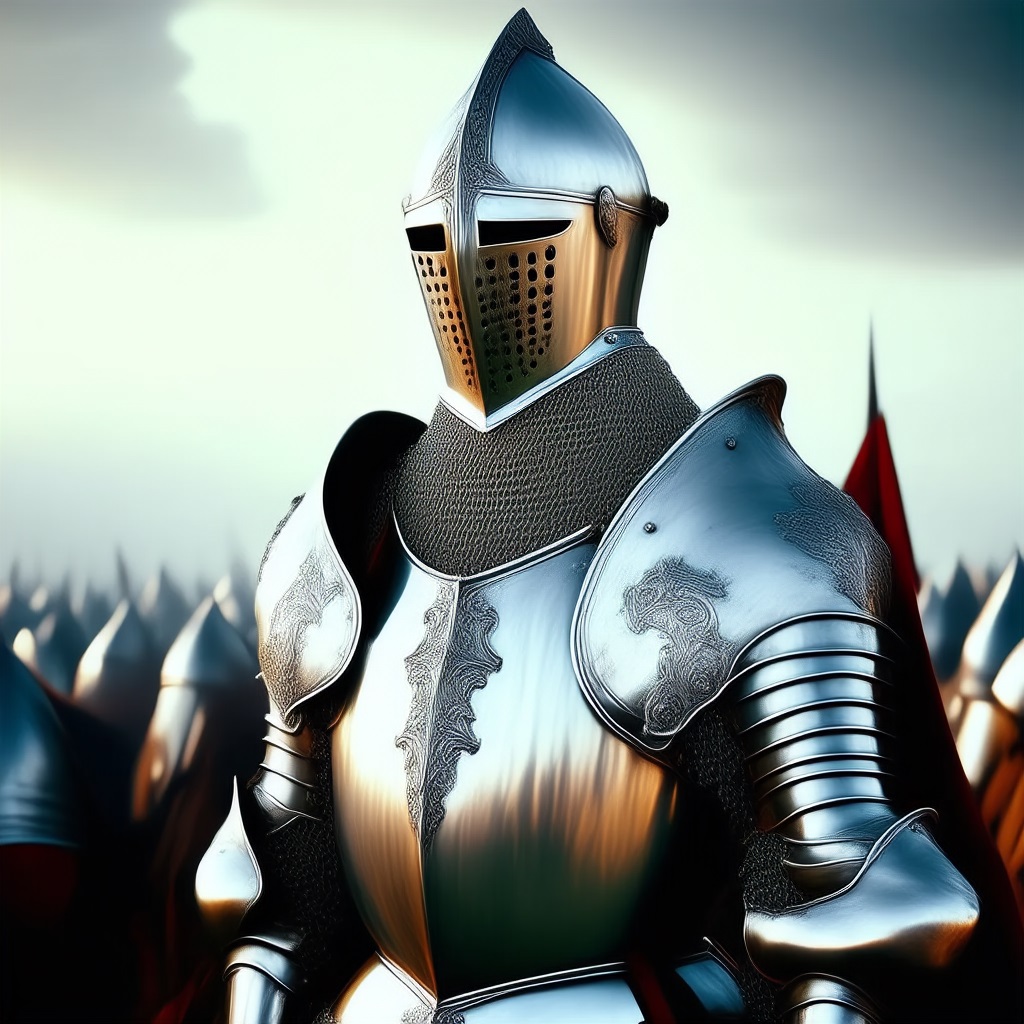} &
\cmpimg{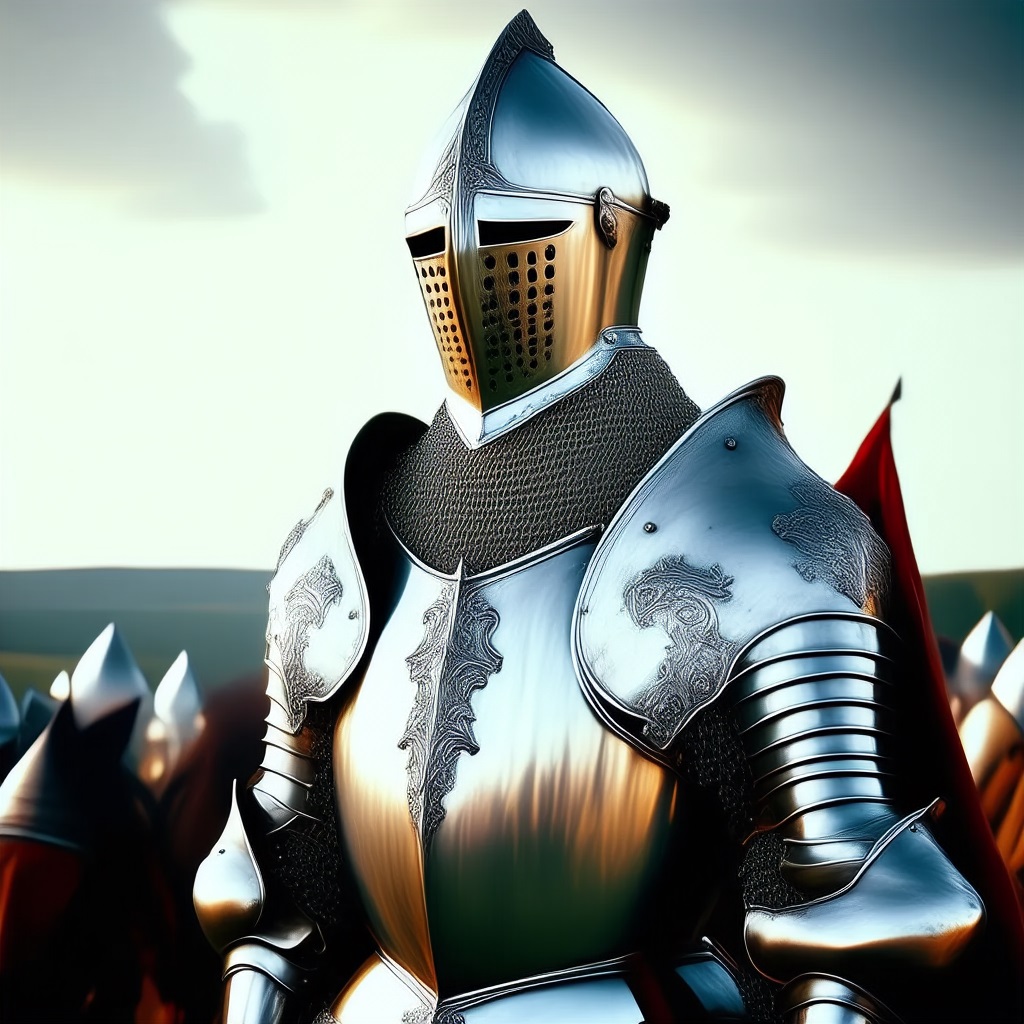} &
\cmpimg{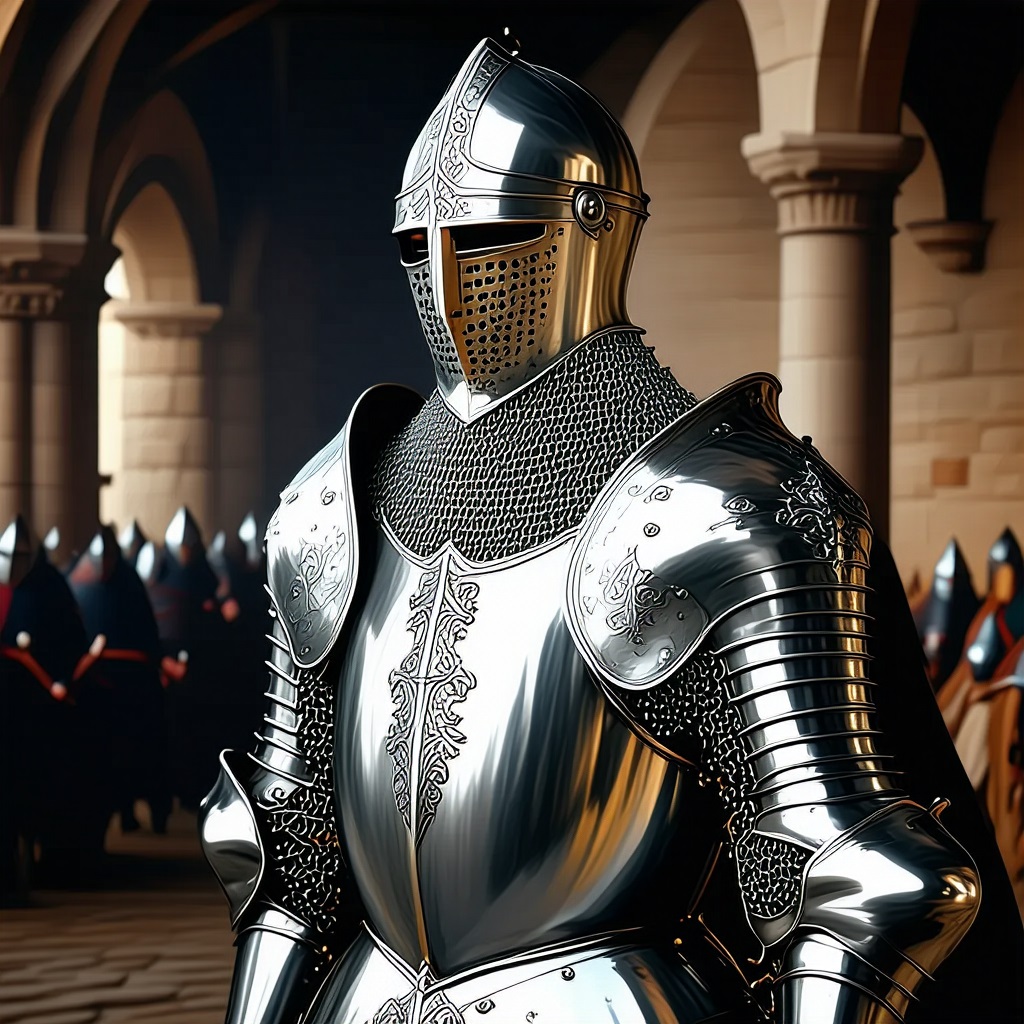} \\
\multicolumn{5}{c}{\parbox{0.96\linewidth}{\centering\footnotesize (a) cinematic portrait of a medieval knight wearing detailed silver armor, dramatic lighting, ultra realistic, 8k.}} \\[7pt]

\cmpimg{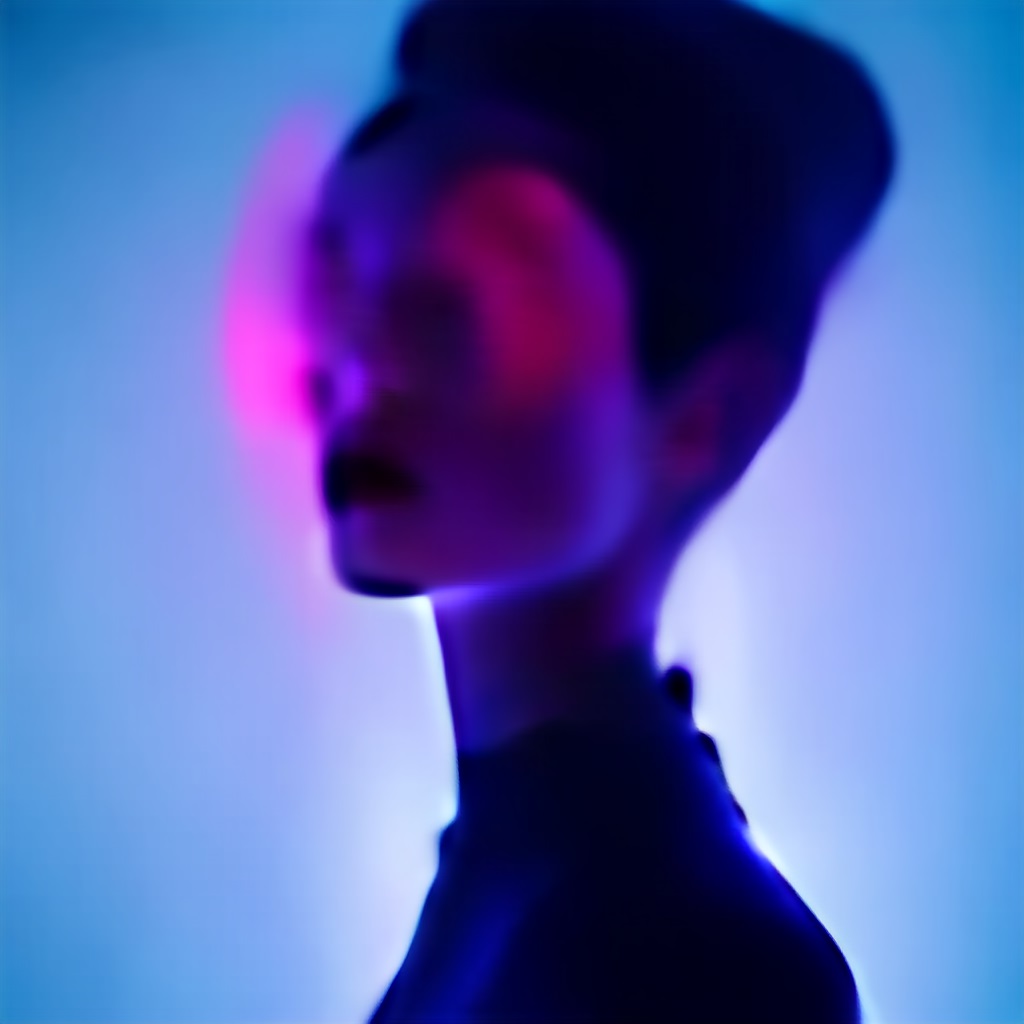} &
\cmpimg{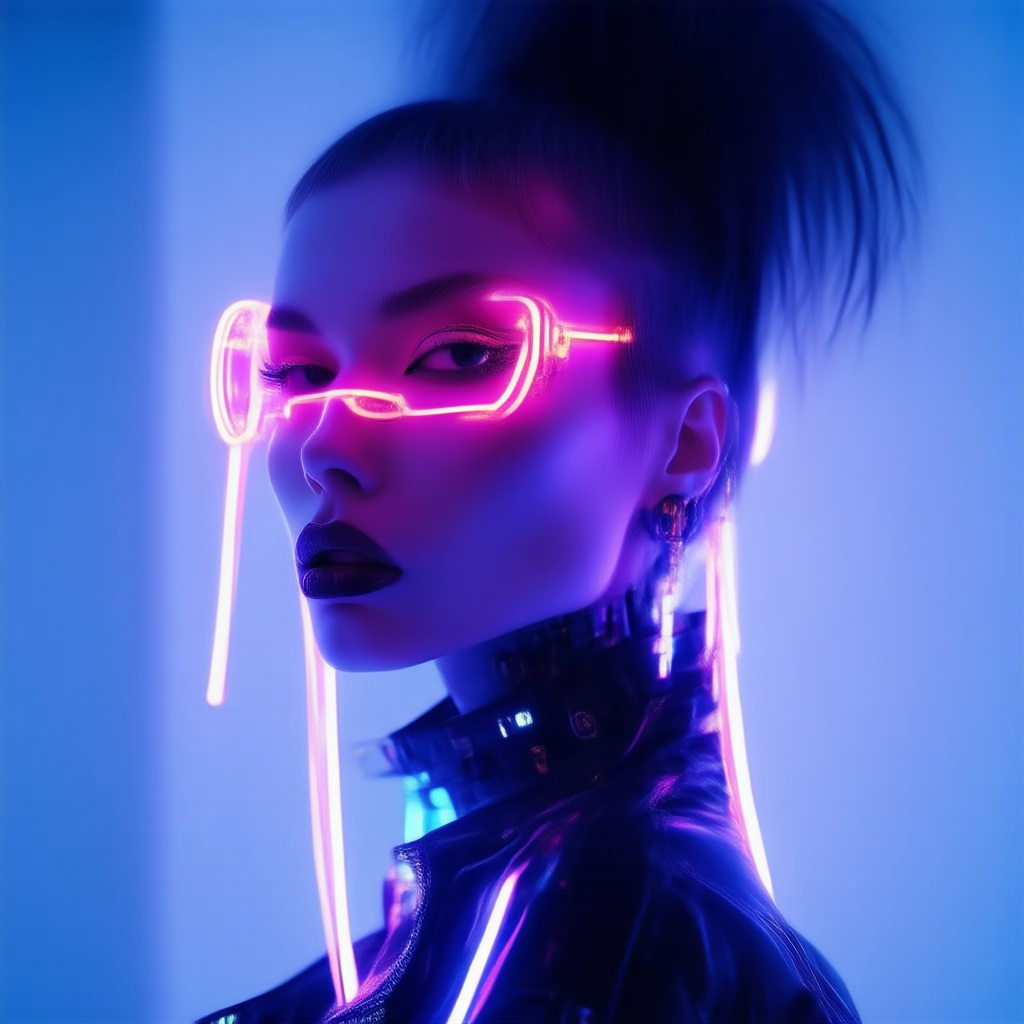} &
\cmpimg{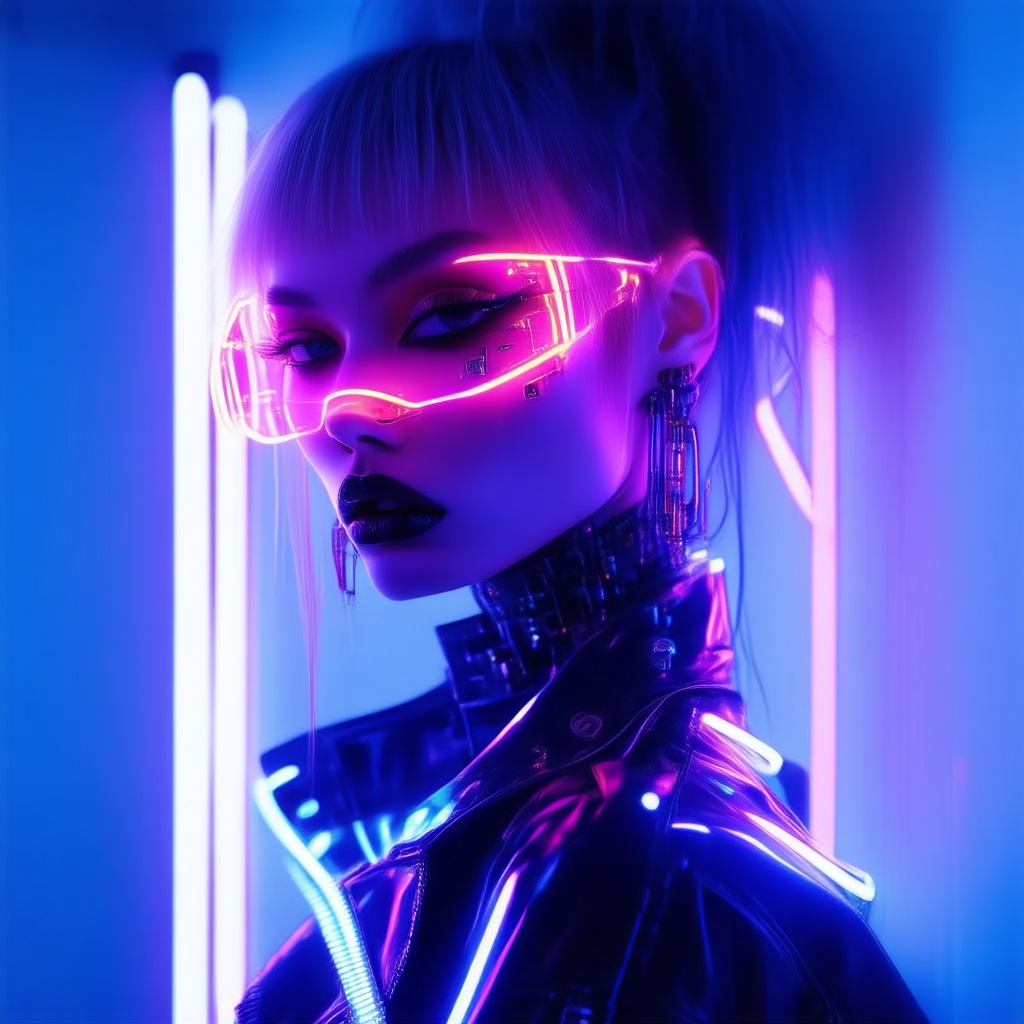} &
\cmpimg{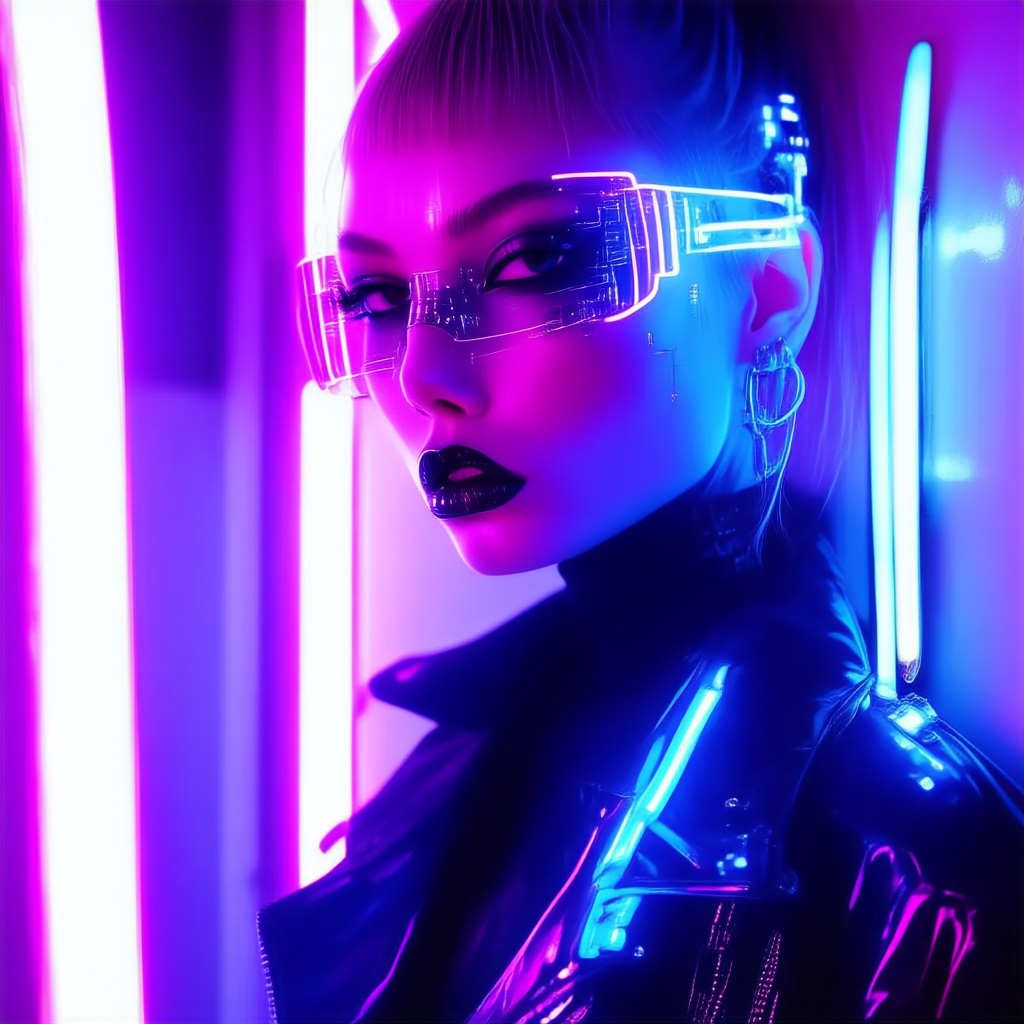} &
\cmpimg{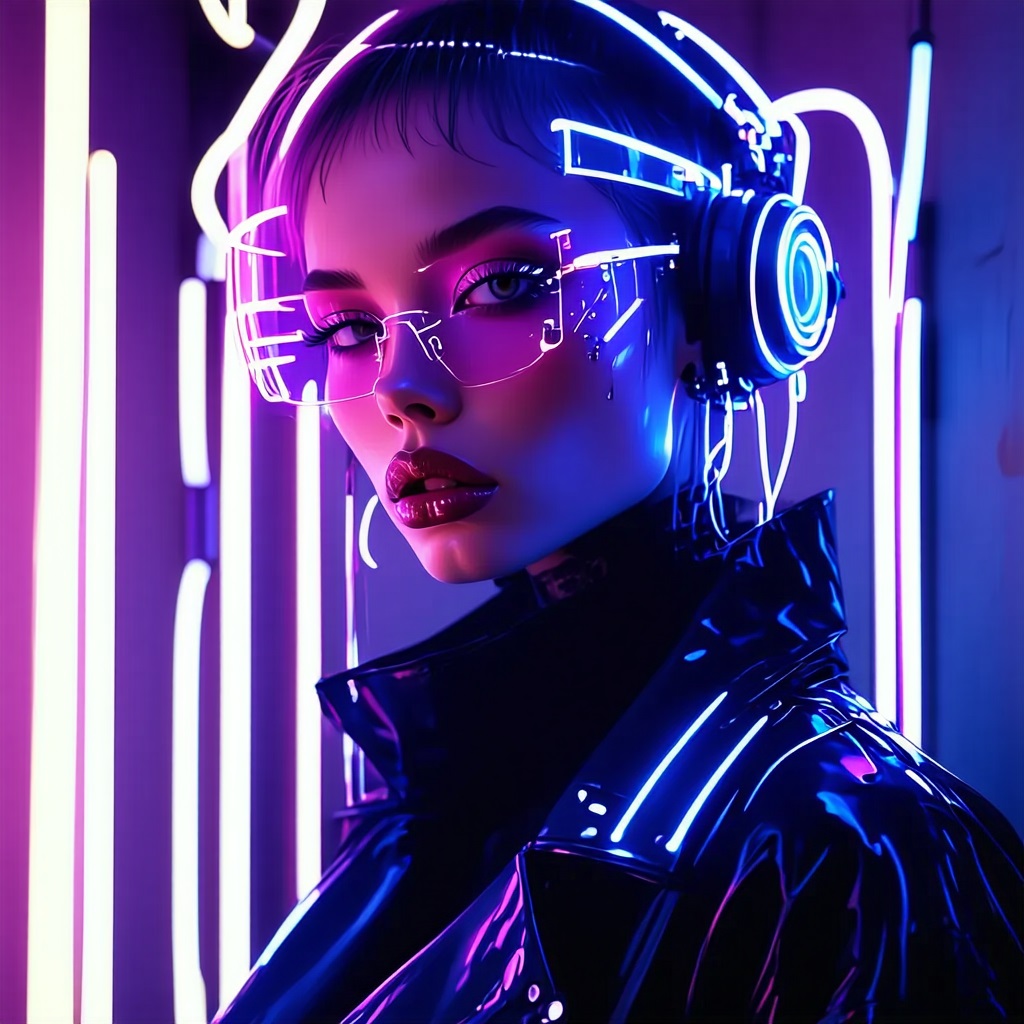} \\
\multicolumn{5}{c}{\parbox{0.96\linewidth}{\centering\footnotesize (b) fashion photography of a futuristic cyberpunk woman with neon lights, studio lighting, high detail.}} \\[7pt]

\cmpimg{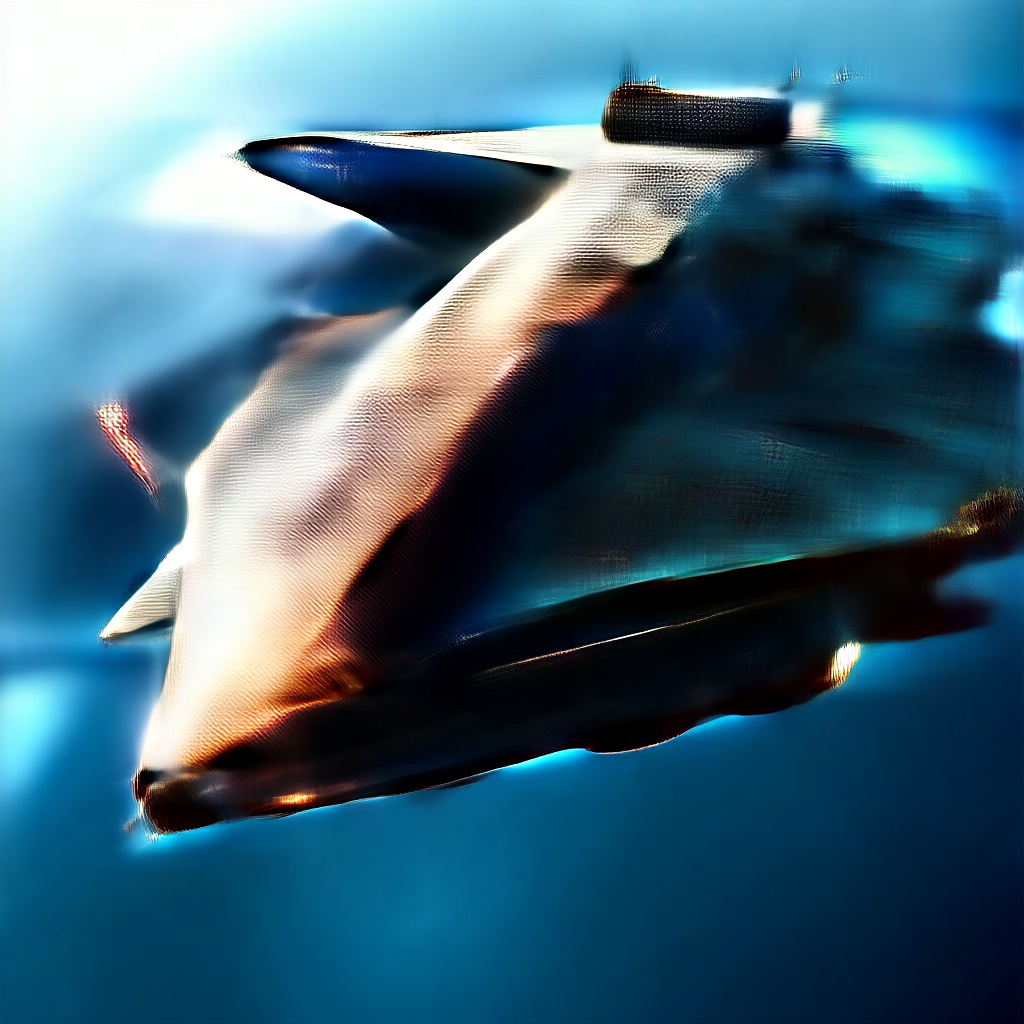} &
\cmpimg{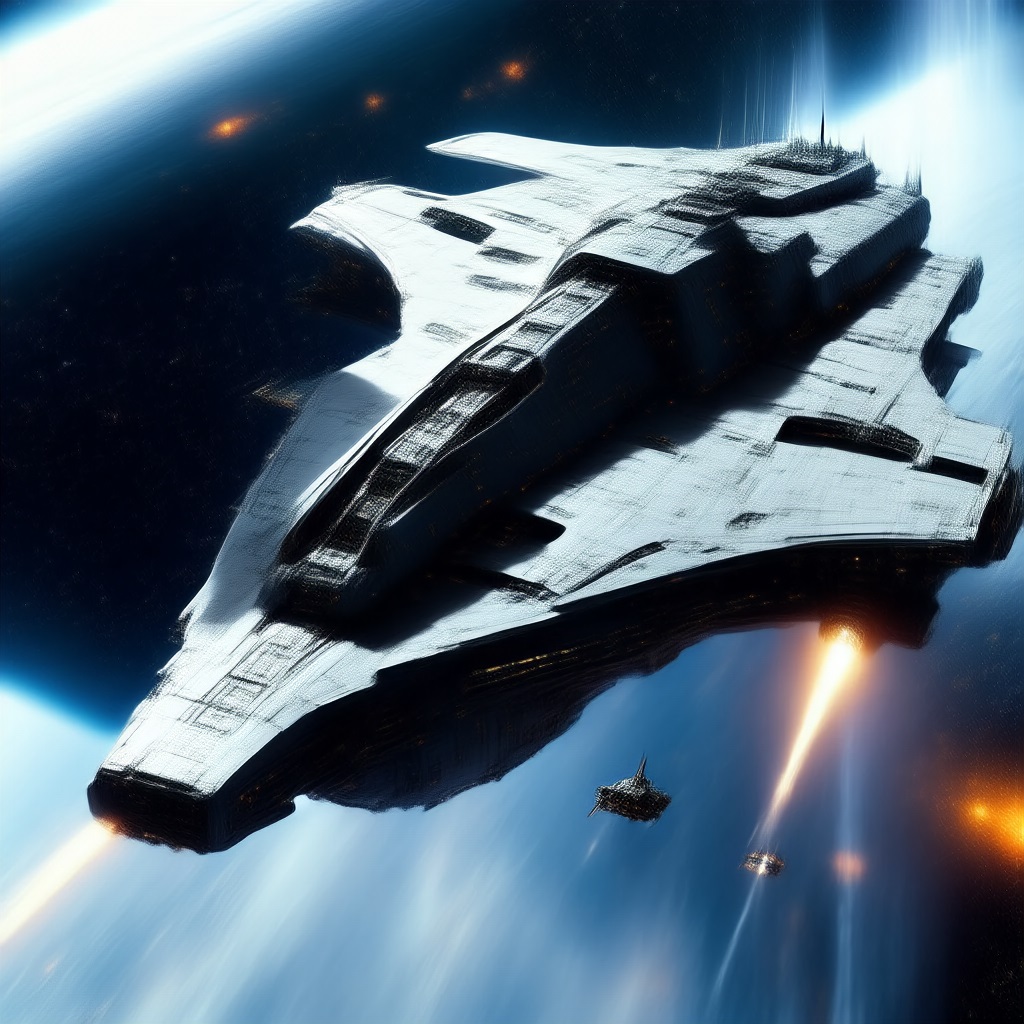} &
\cmpimg{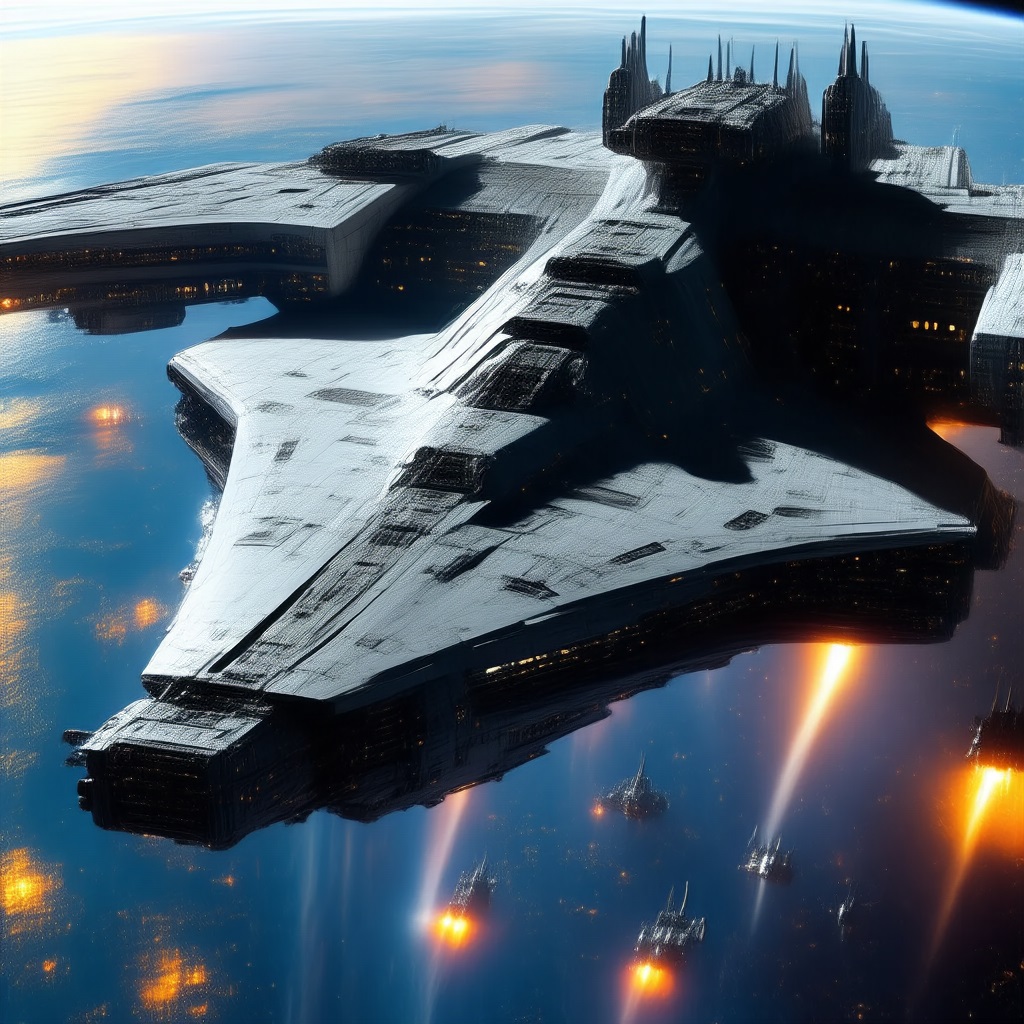} &
\cmpimg{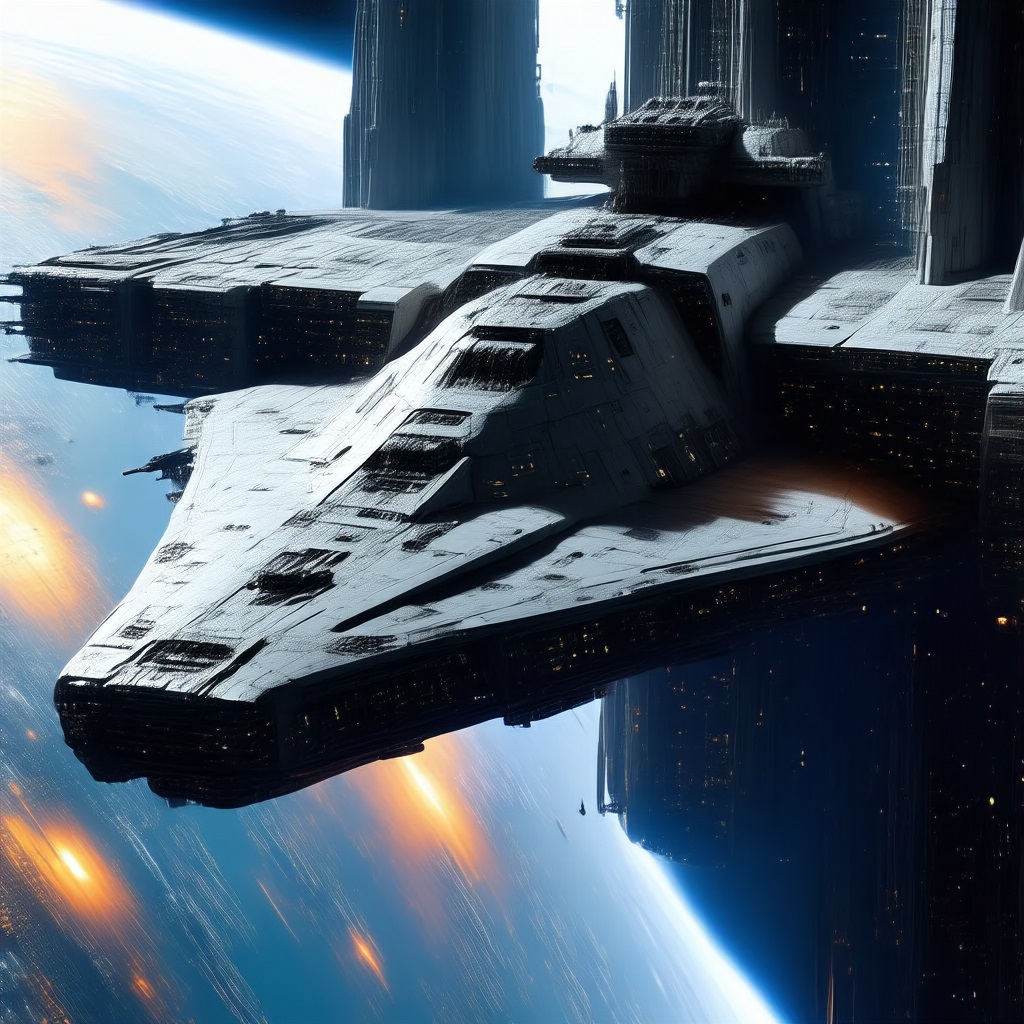} &
\cmpimg{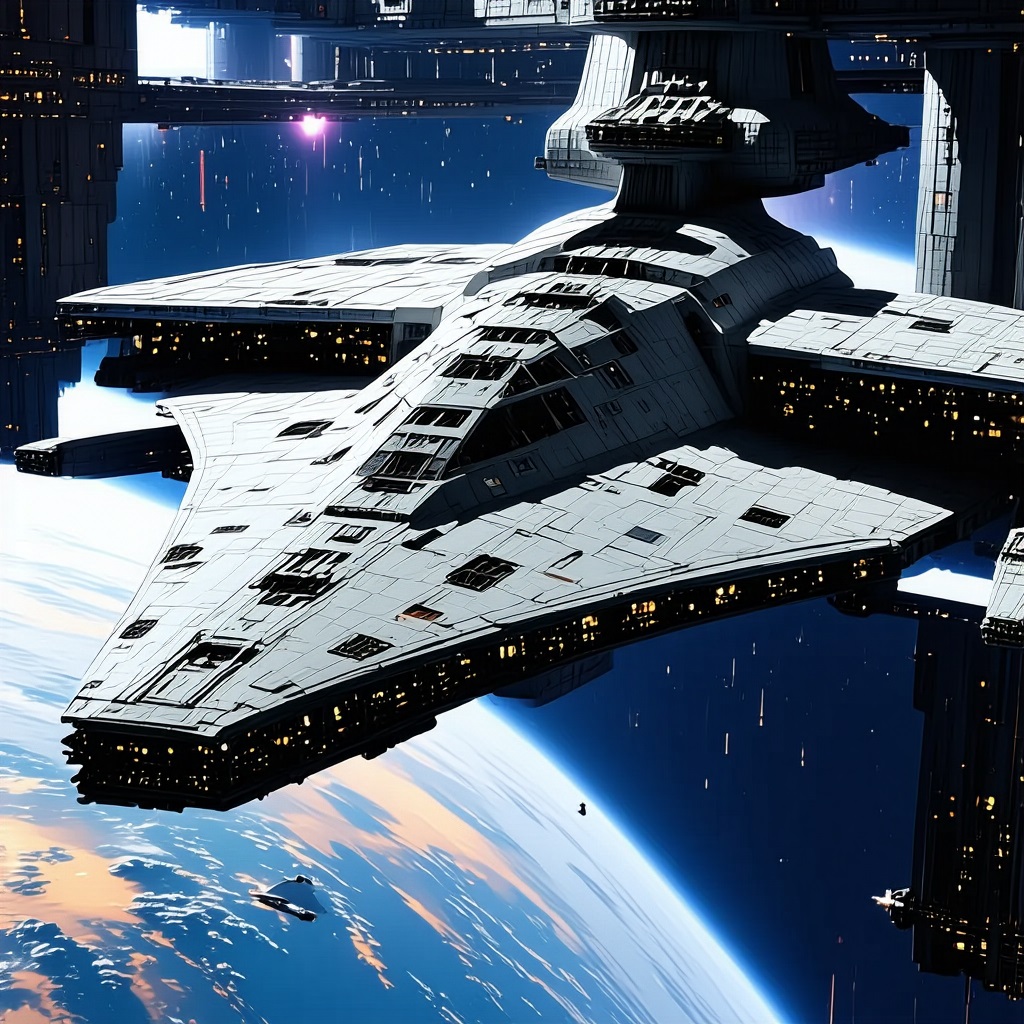} \\
\multicolumn{5}{c}{\parbox{0.96\linewidth}{\centering\footnotesize (c) a massive futuristic spaceship docking at a space station, cinematic science fiction scene.}} \\[7pt]

\cmpimg{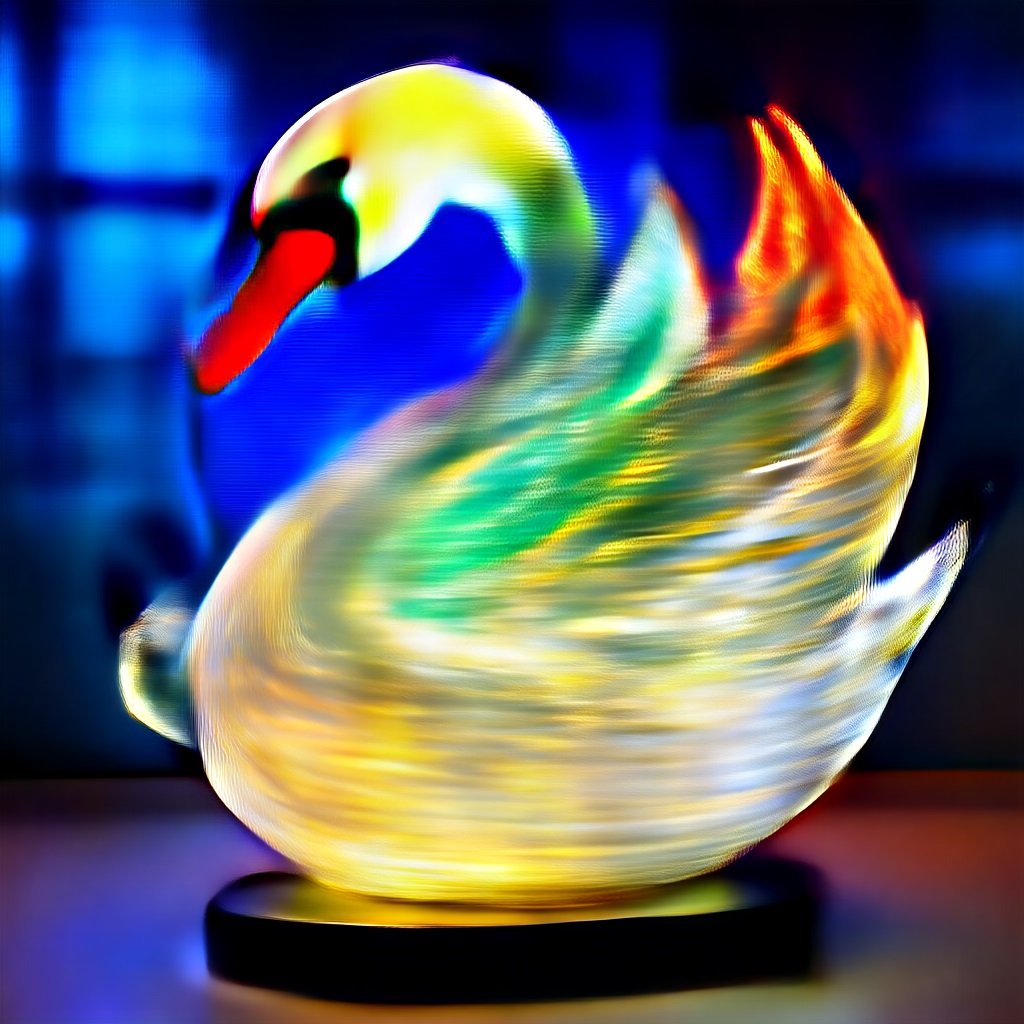} &
\cmpimg{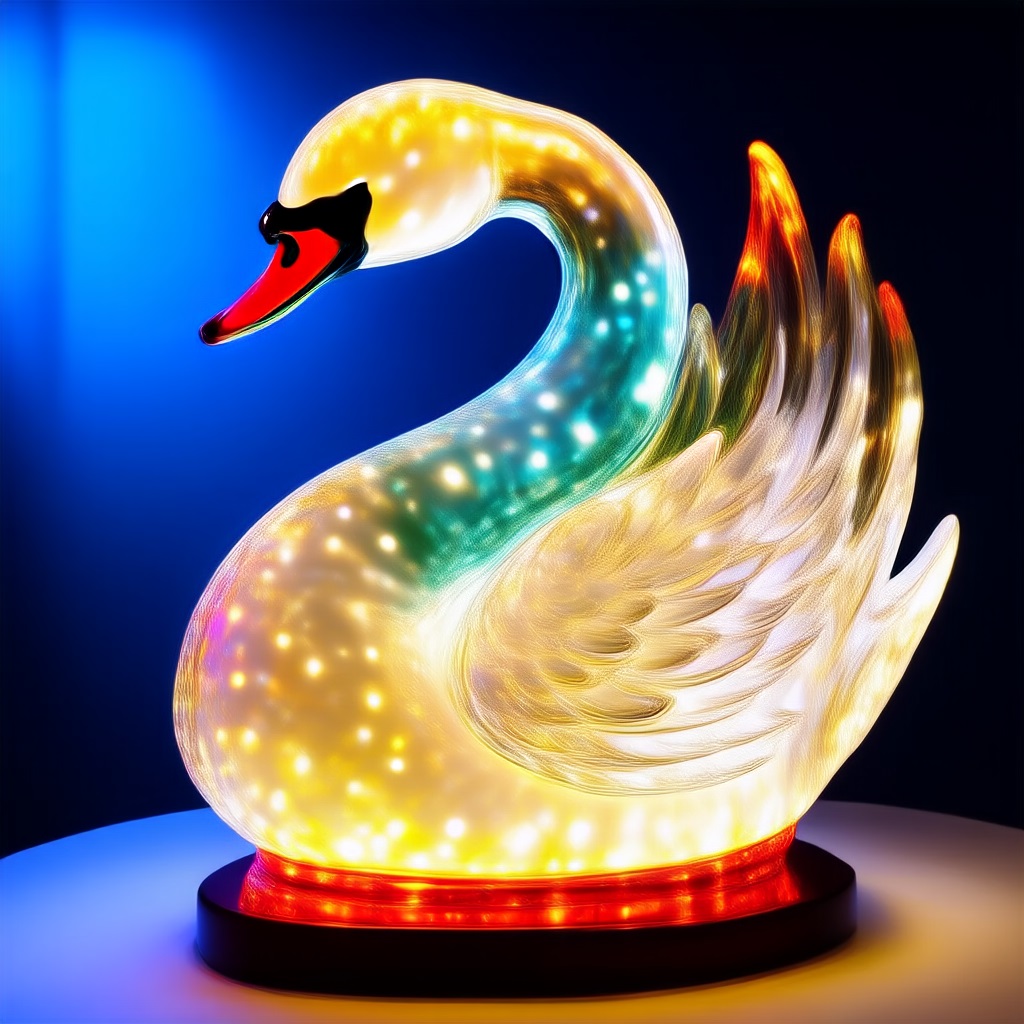} &
\cmpimg{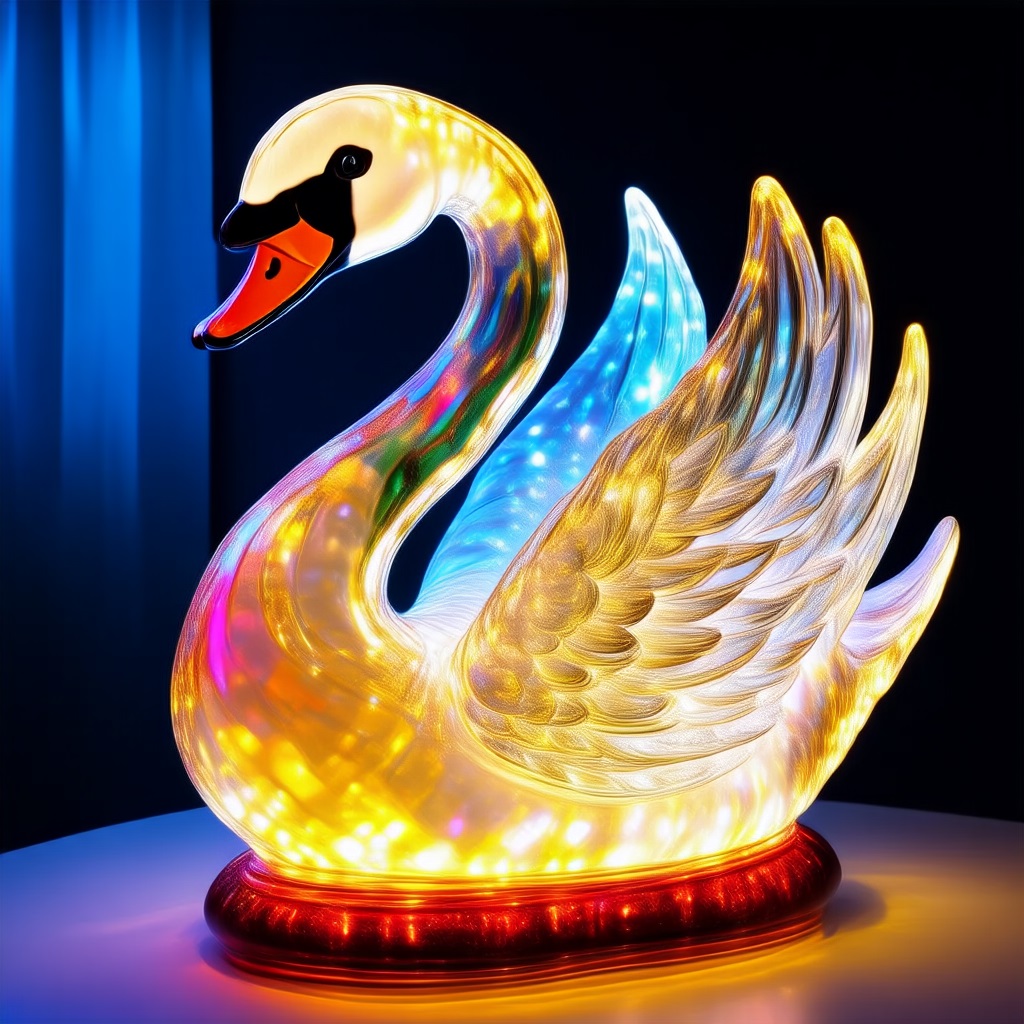} &
\cmpimg{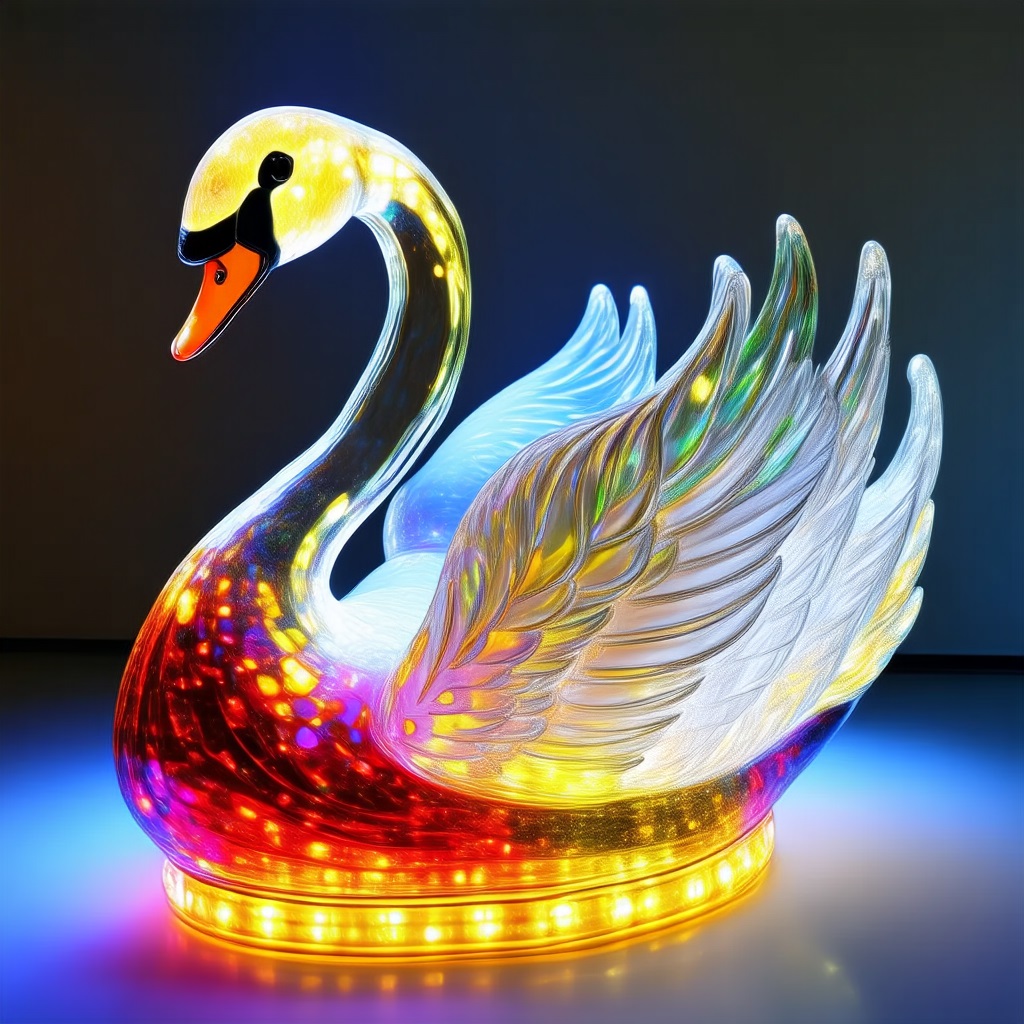} &
\cmpimg{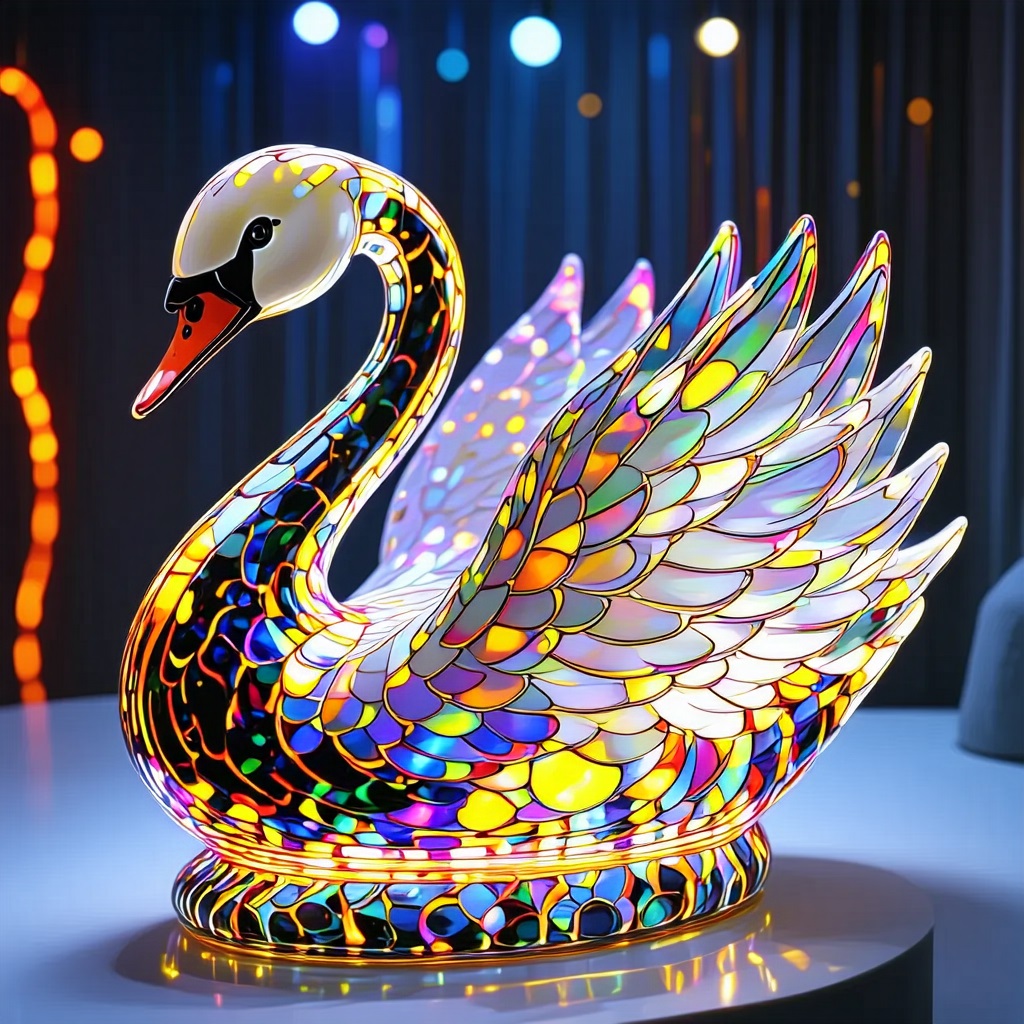} \\
\multicolumn{5}{c}{\parbox{0.96\linewidth}{\centering\footnotesize (d) a glass sculpture shaped like a swan, illuminated by colorful lights, ultra detailed.}} \\[7pt]

\cmpimg{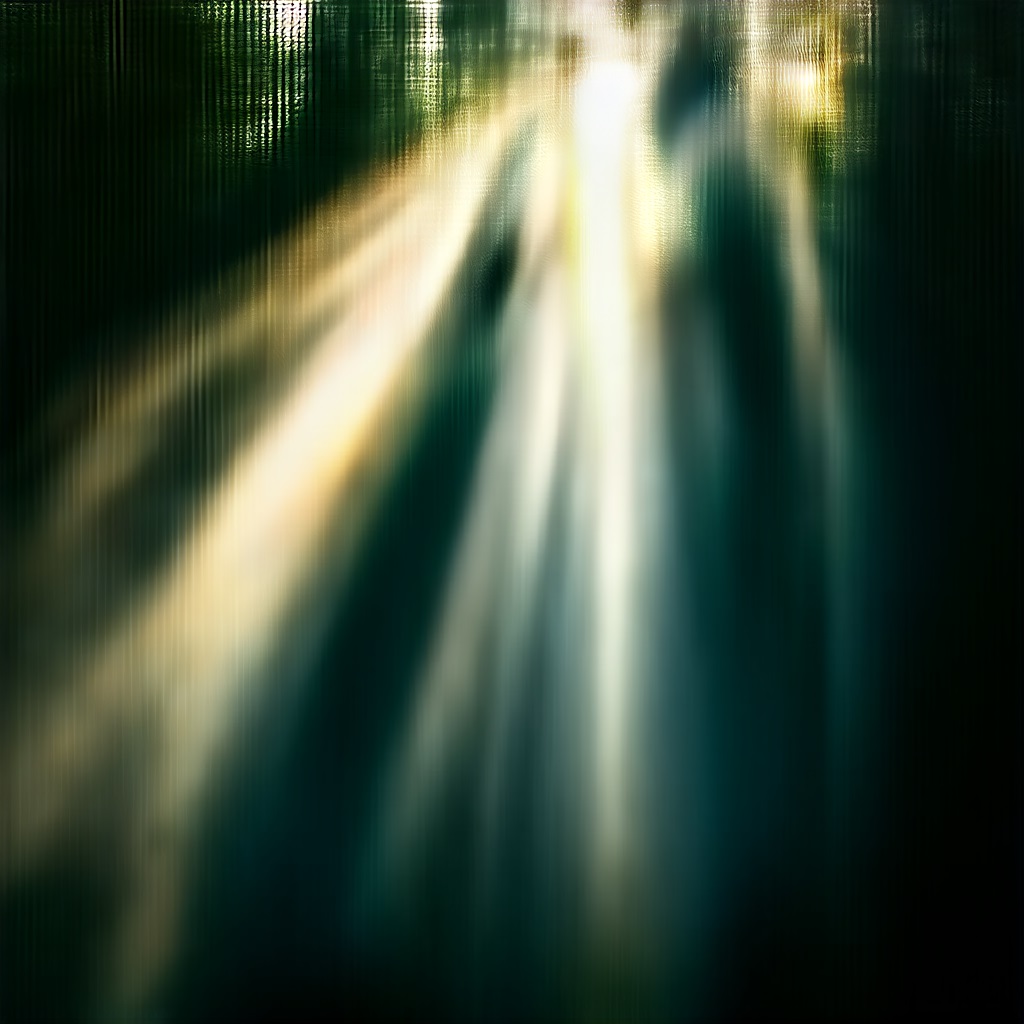} &
\cmpimg{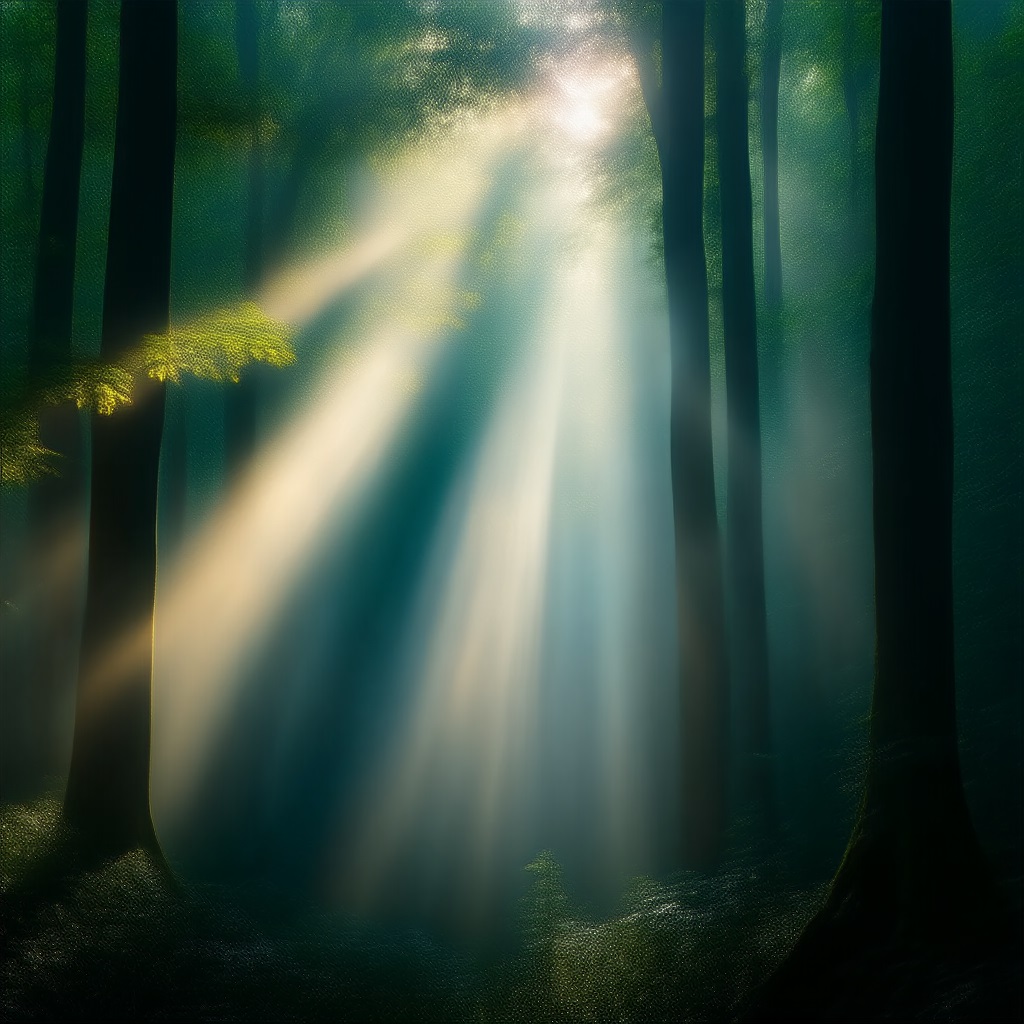} &
\cmpimg{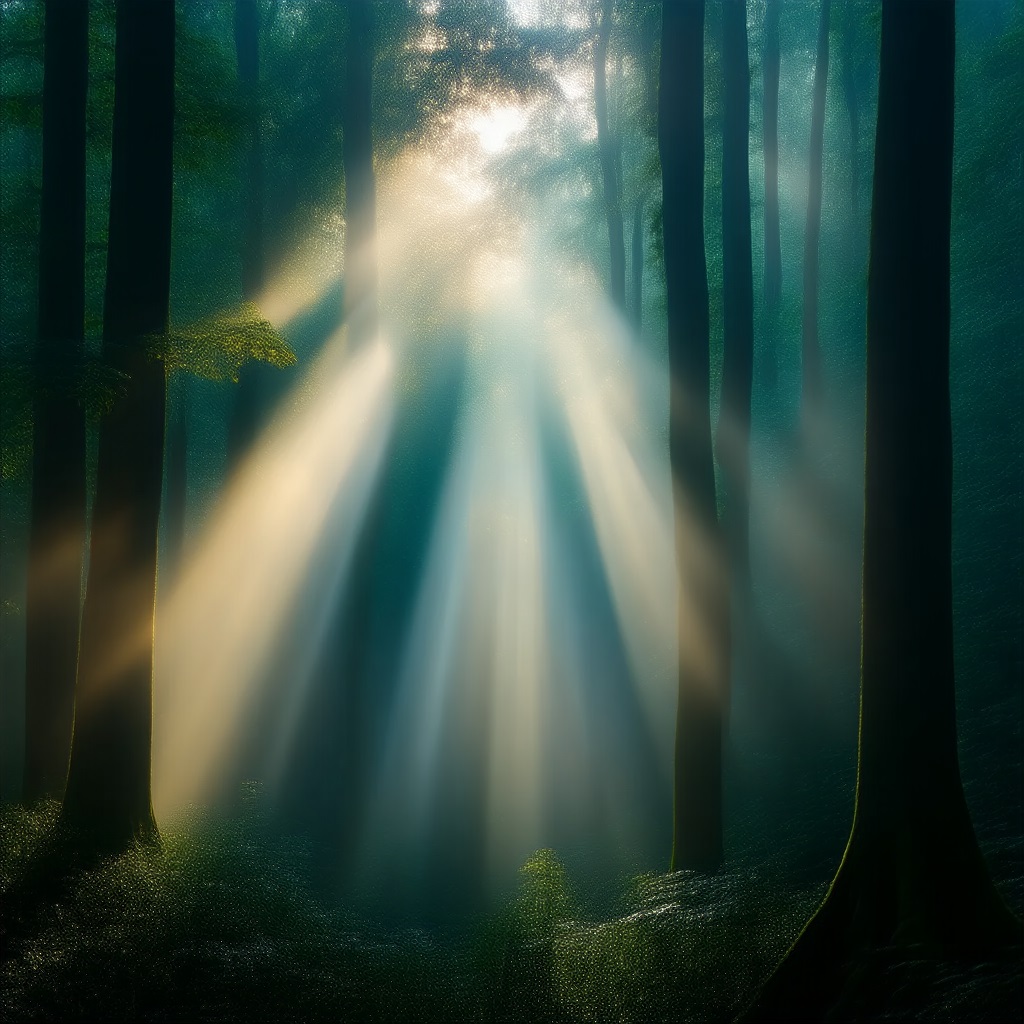} &
\cmpimg{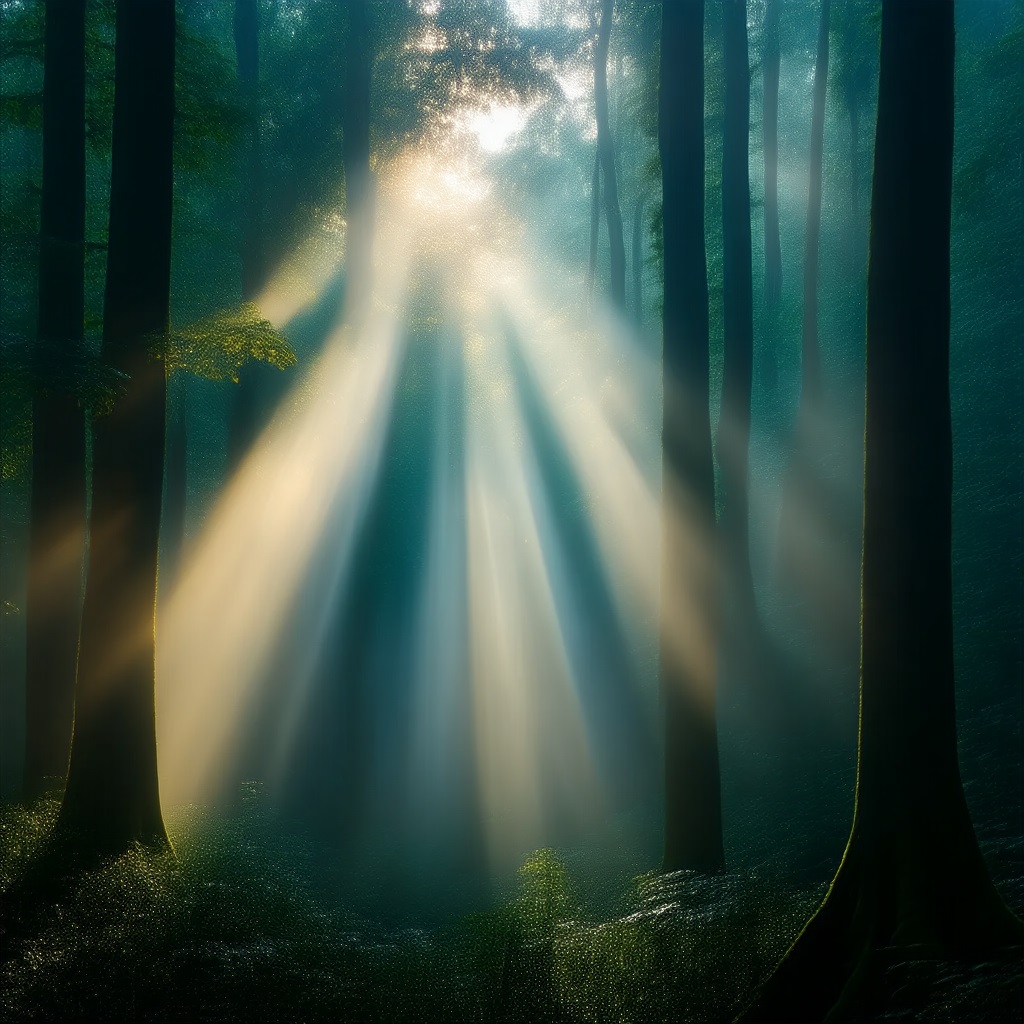} &
\cmpimg{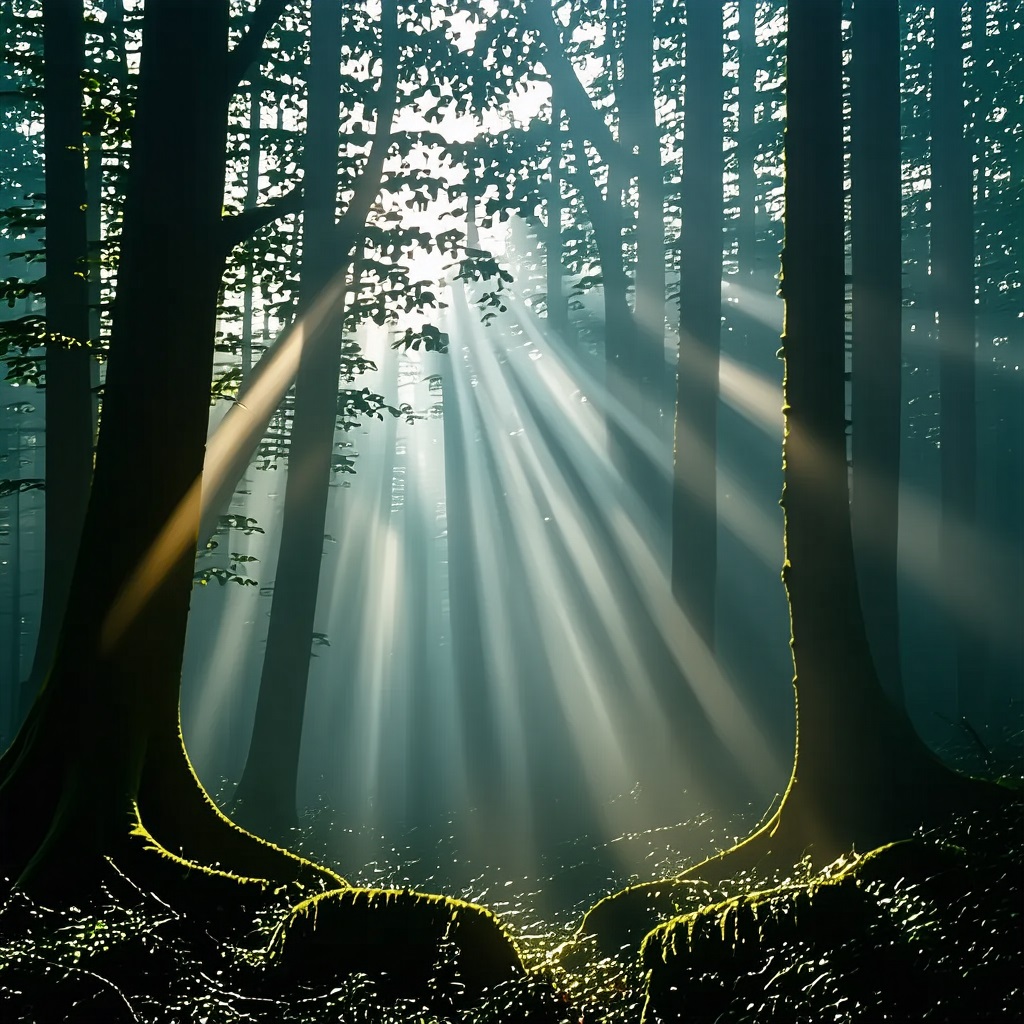} \\
\multicolumn{5}{c}{\parbox{0.96\linewidth}{\centering\footnotesize (e) A majestic misty forest at dawn, volumetric god rays piercing through dense canopy, ethereal atmosphere, cinematic photography.}} \\[4pt]

\end{tabular}

\caption{Qualitative visual comparison under different sampling step settings. Each row presents a representative prompt, and the columns show images generated by the original 28-step base model when sampled with 4, 8, 16, and 28 steps, followed by the result of our distilled 4-step model. When the undistilled base model is directly evaluated with fewer sampling steps, image quality tends to degrade, with visible artifacts such as structural instability, loss of fine-grained details, and reduced semantic fidelity. In contrast, the distilled 4-step model recovers much of the visual quality of the original 28-step model in these examples, showing that DMD2 distillation enables a $7\times$ reduction in sampling steps with limited perceptual degradation.}
\label{fig:28vs4}
\end{figure}

\subsubsection{Distillation Efficacy and Acceleration}
Fig.~\ref{fig:28vs4} illustrates the effectiveness of our DMD2 distillation strategy.
While the non-distilled model suffers from structural collapse at lower inference steps (4, 8, and 16 steps), our 4-step distilled model maintains visual fidelity nearly identical to the 28-step baseline.
This $7\times$ acceleration introduces virtually no perceptual degradation, yielding an optimal efficiency-fidelity trade-off for on-device deployment~\cite{yin2024improved_dmd,luo2023lcm}.

\begin{figure}[htbp]
\centering
\setlength{\tabcolsep}{1pt}
\renewcommand{\arraystretch}{1.05}

\newcommand{\poetryimg}[1]{%
\includegraphics[
  width=0.235\linewidth,
  height=0.20\textheight,
  keepaspectratio
]{#1}
}

\begin{tabular}{cccc}

\multicolumn{1}{c}{\textbf{\footnotesize Pop Art}} &
\multicolumn{1}{c}{\textbf{\footnotesize 3D Animated Film}} &
\multicolumn{1}{c}{\textbf{\footnotesize Comic}} &
\multicolumn{1}{c}{\textbf{\footnotesize Chinese Classical Painting}} \\[6pt]

\poetryimg{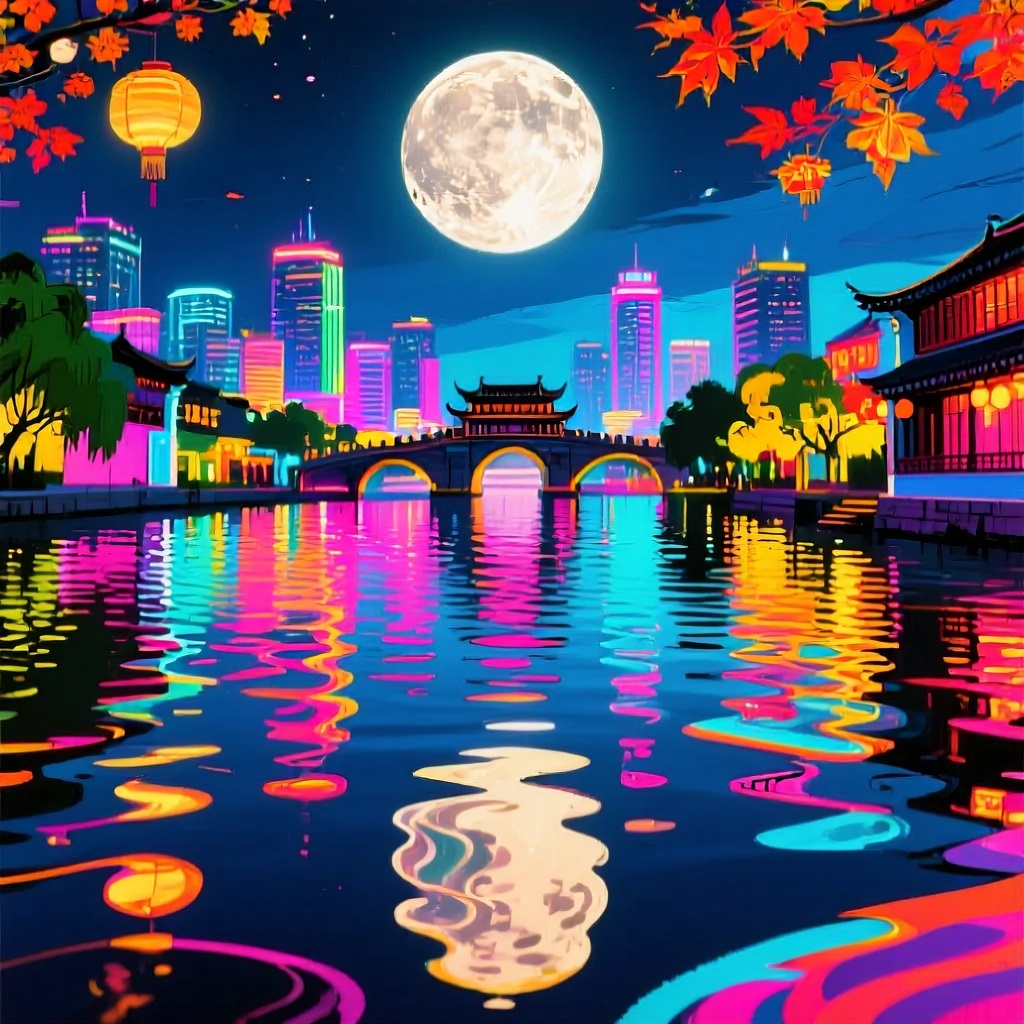} &
\poetryimg{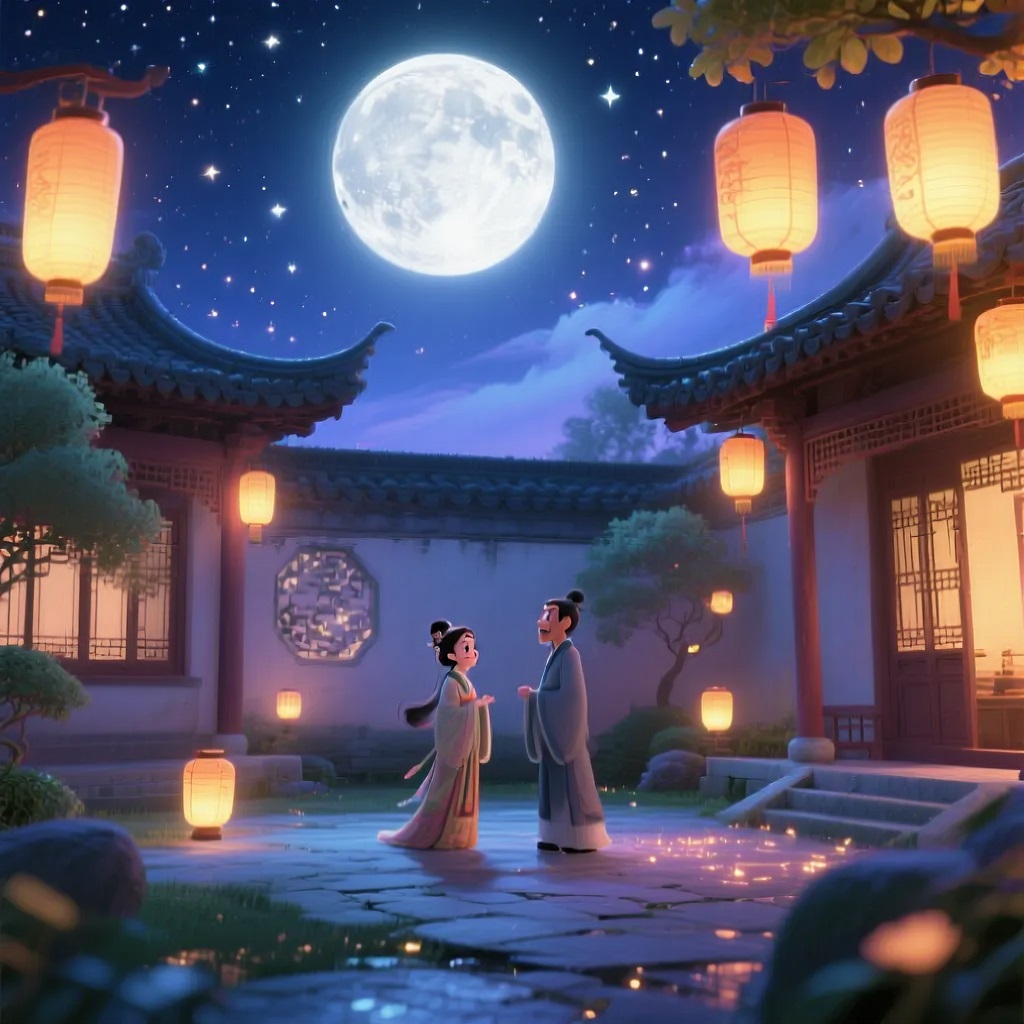} &
\poetryimg{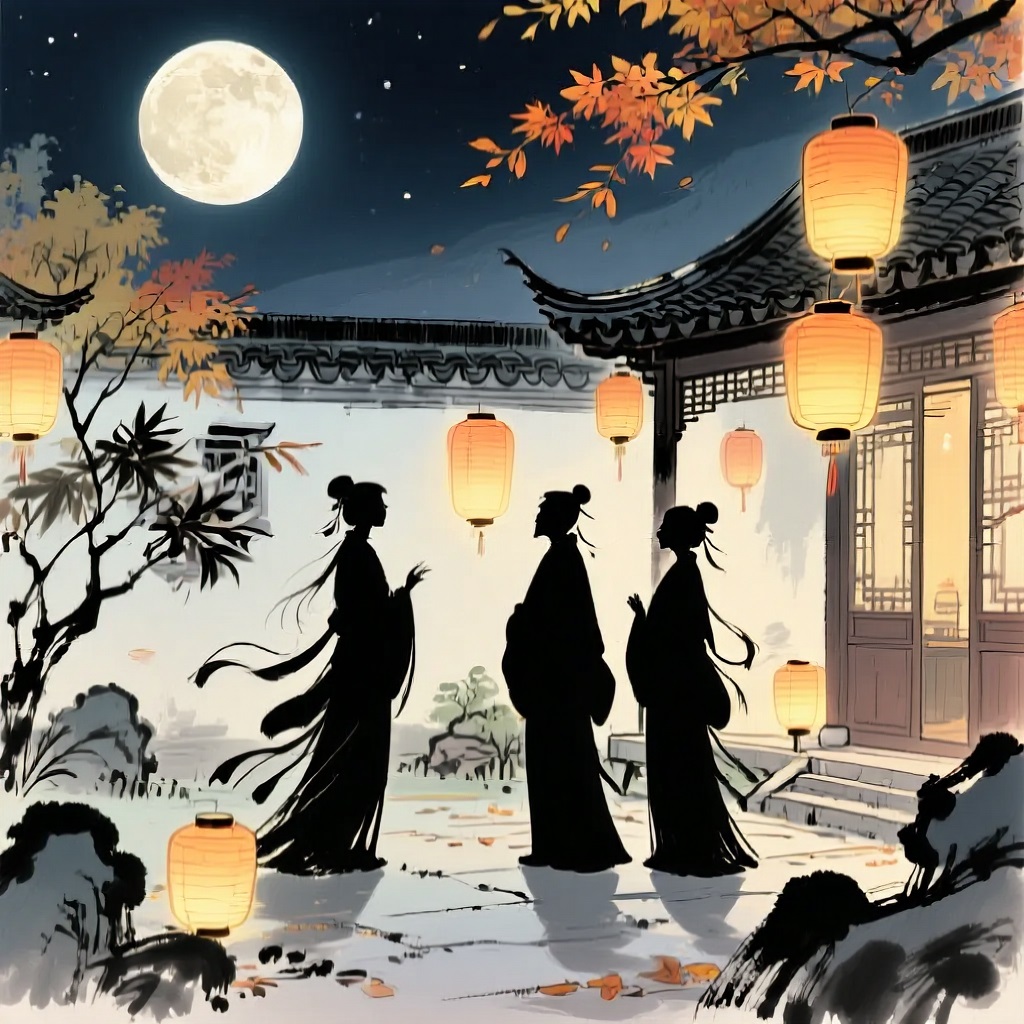} &
\poetryimg{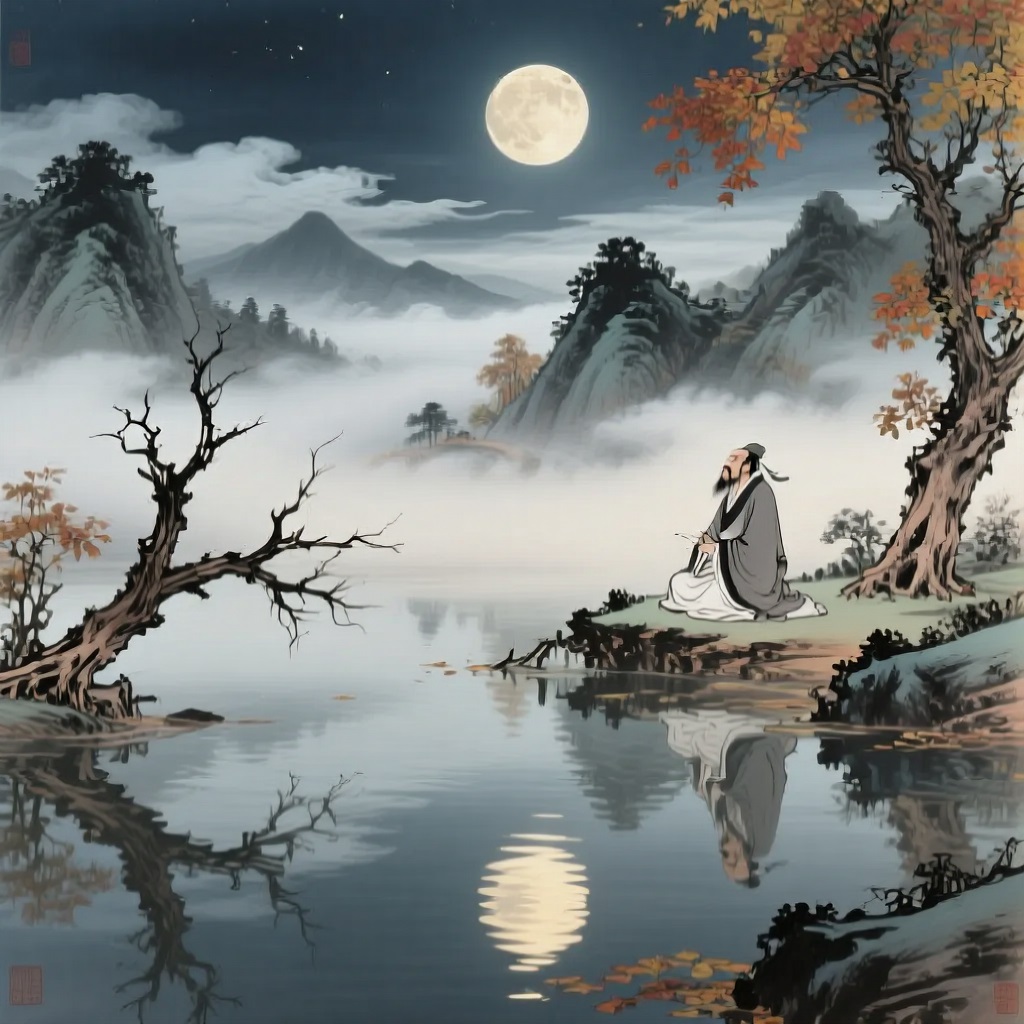} \\
\multicolumn{4}{c}{\footnotesize (a) \zh{月到中秋例属苏}} \\[10pt]

\poetryimg{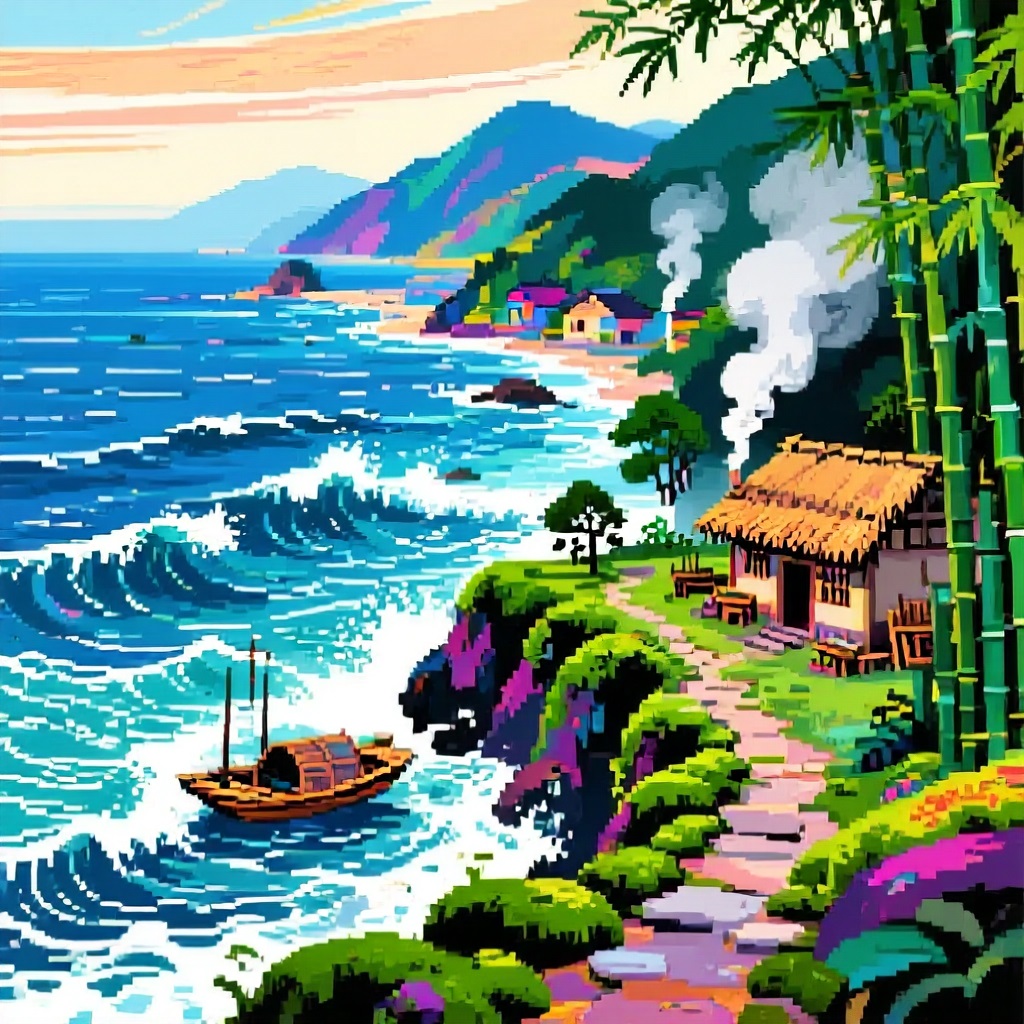} &
\poetryimg{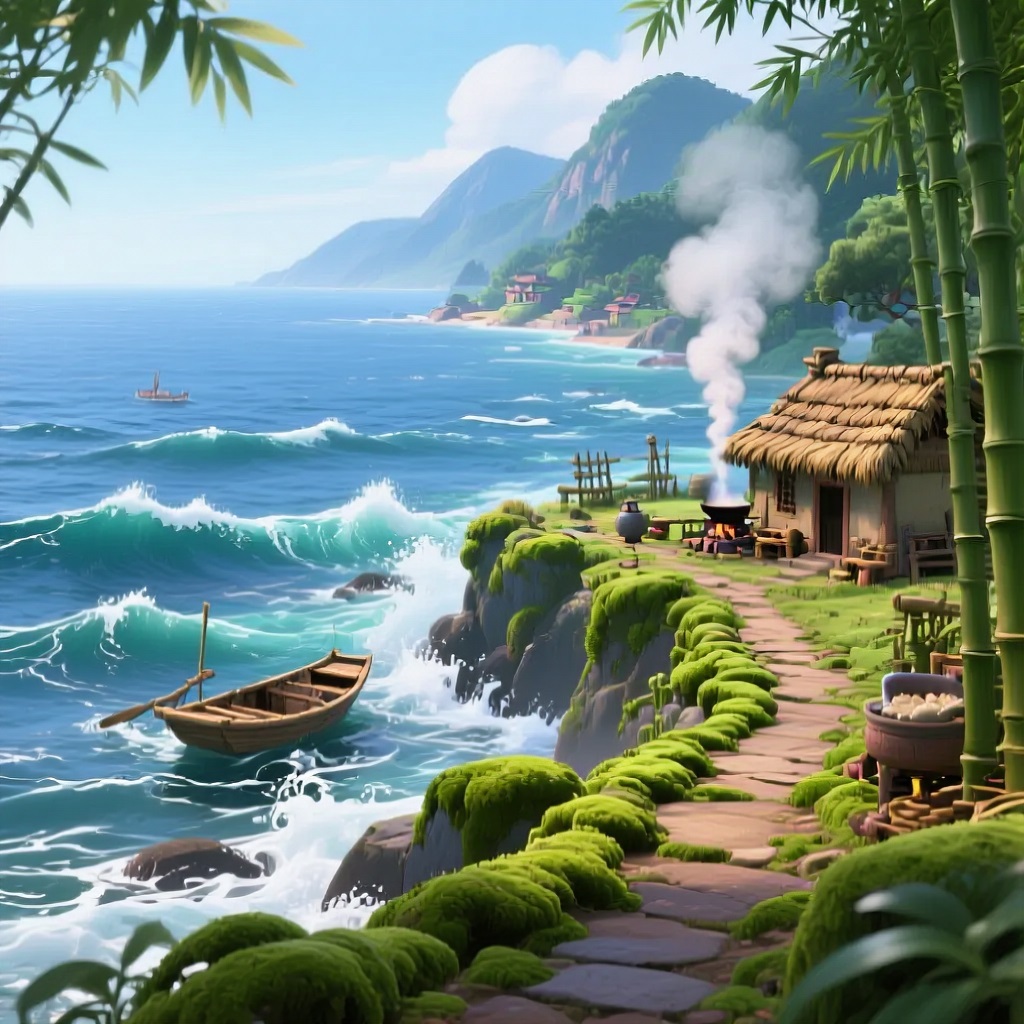} &
\poetryimg{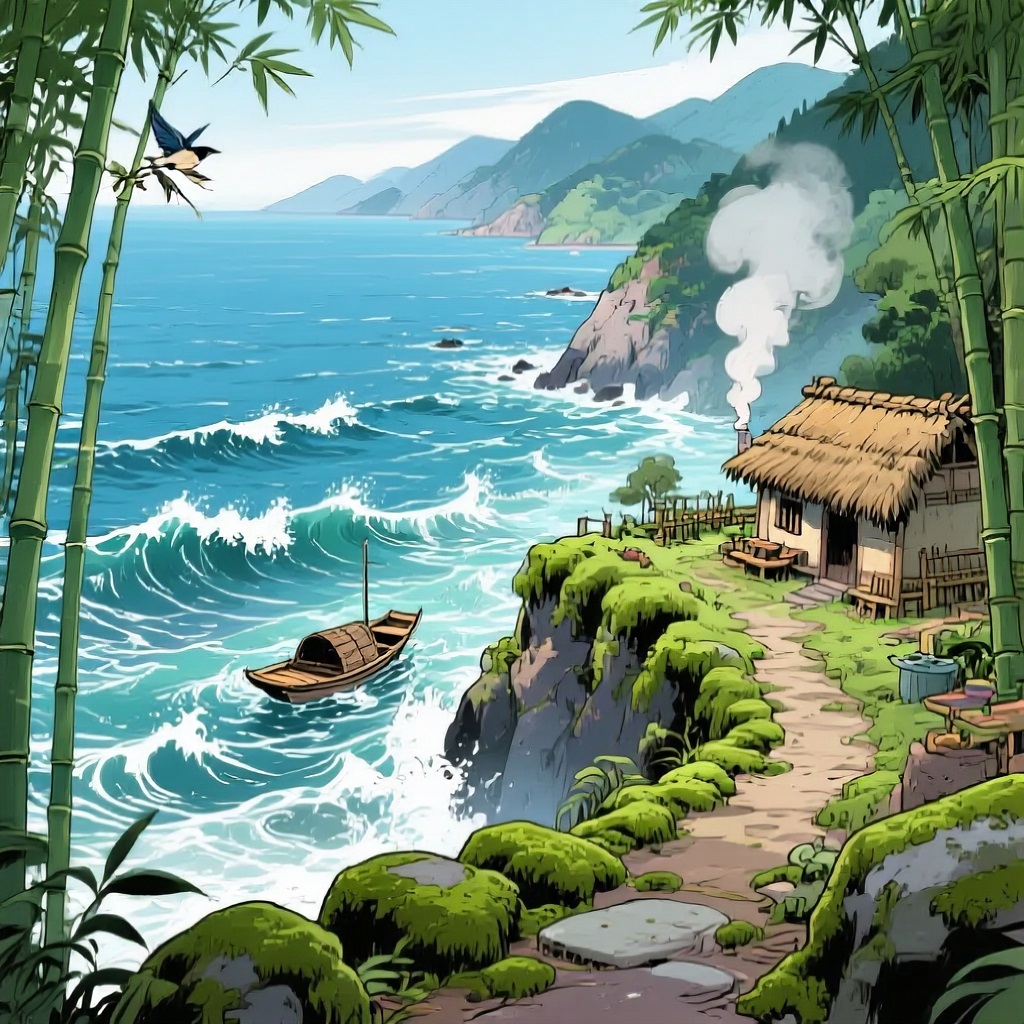} &
\poetryimg{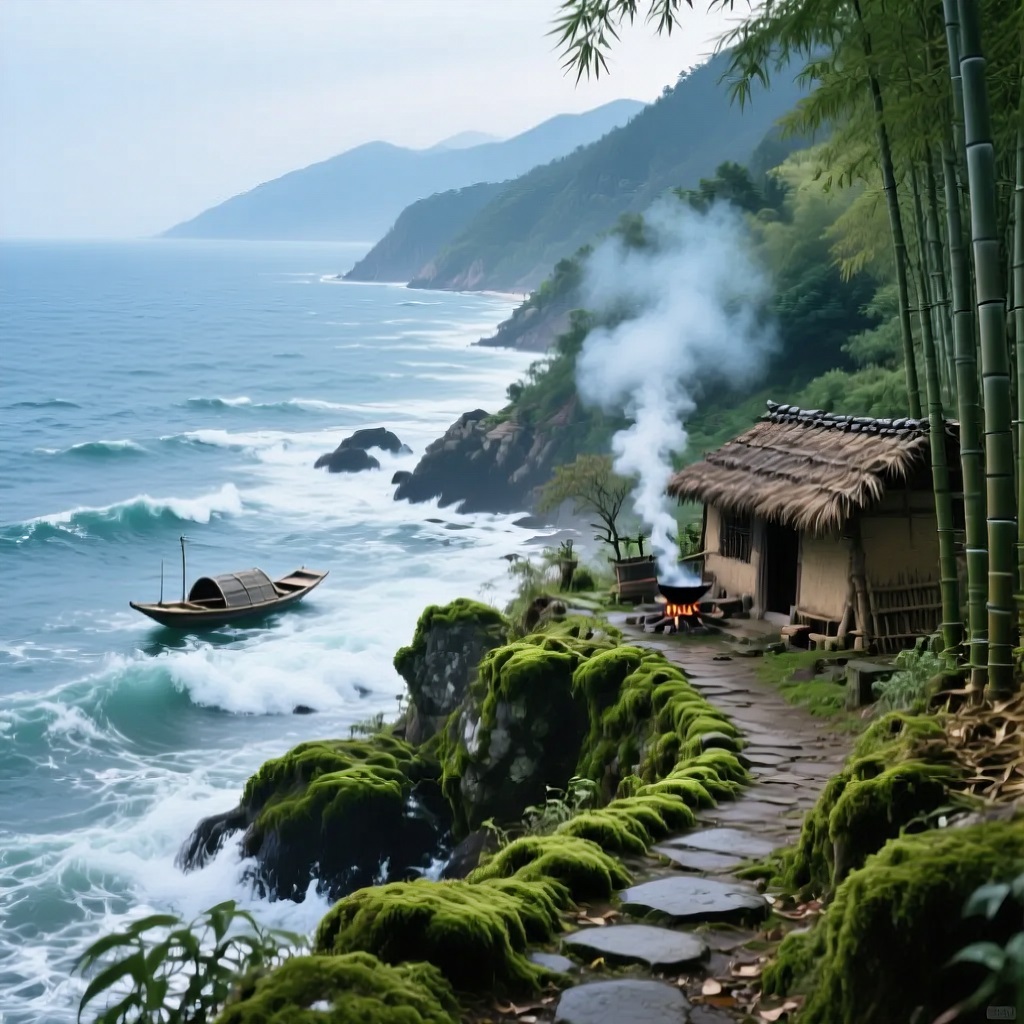} \\
\multicolumn{4}{c}{\footnotesize (b) \zh{曾经沧海难为水，从此桃源便是家}} \\[10pt]

\poetryimg{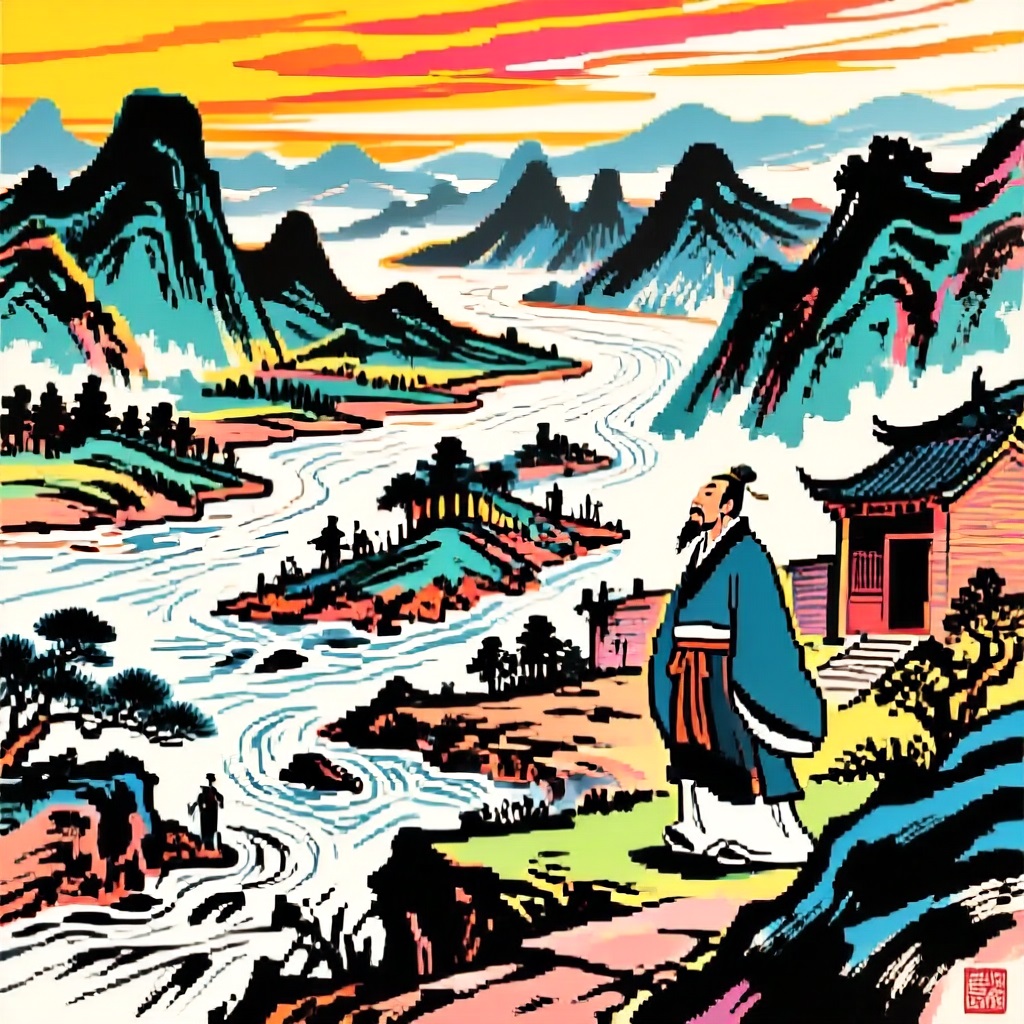} &
\poetryimg{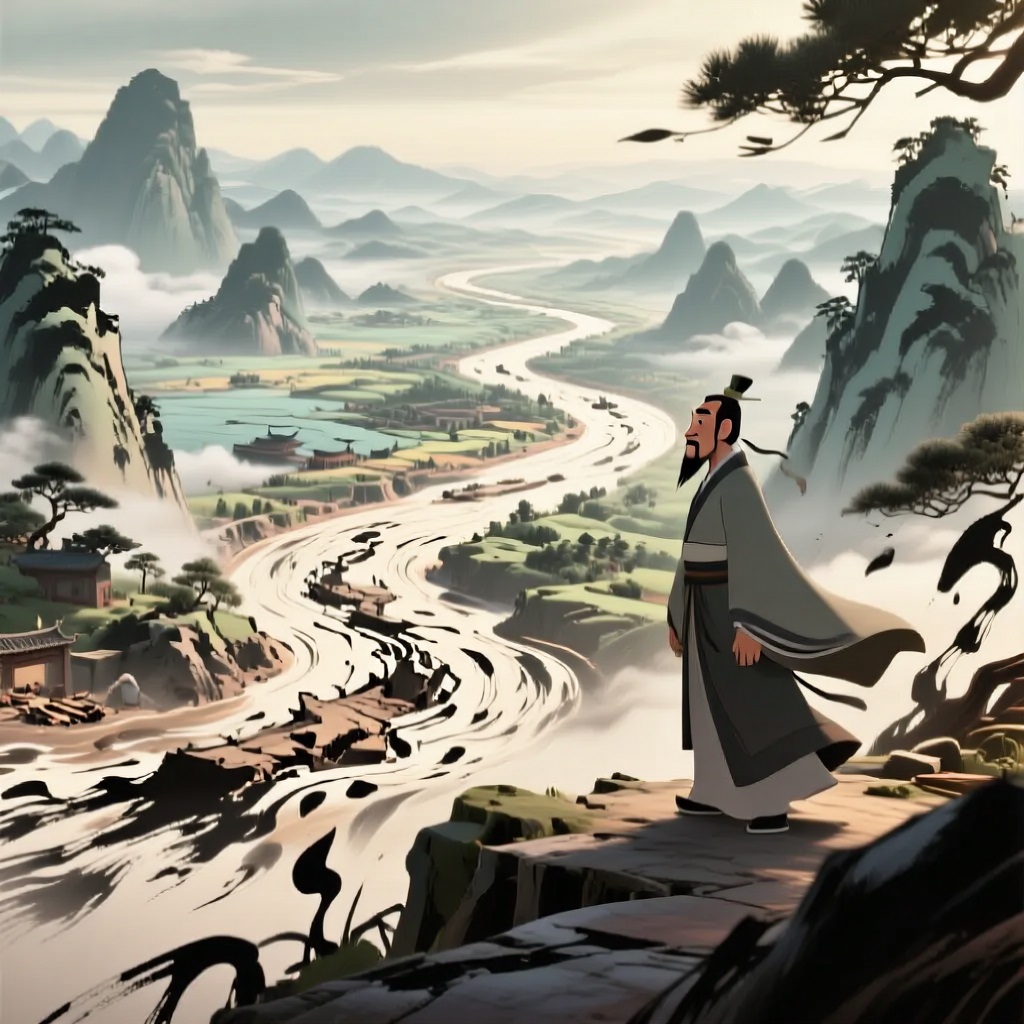} &
\poetryimg{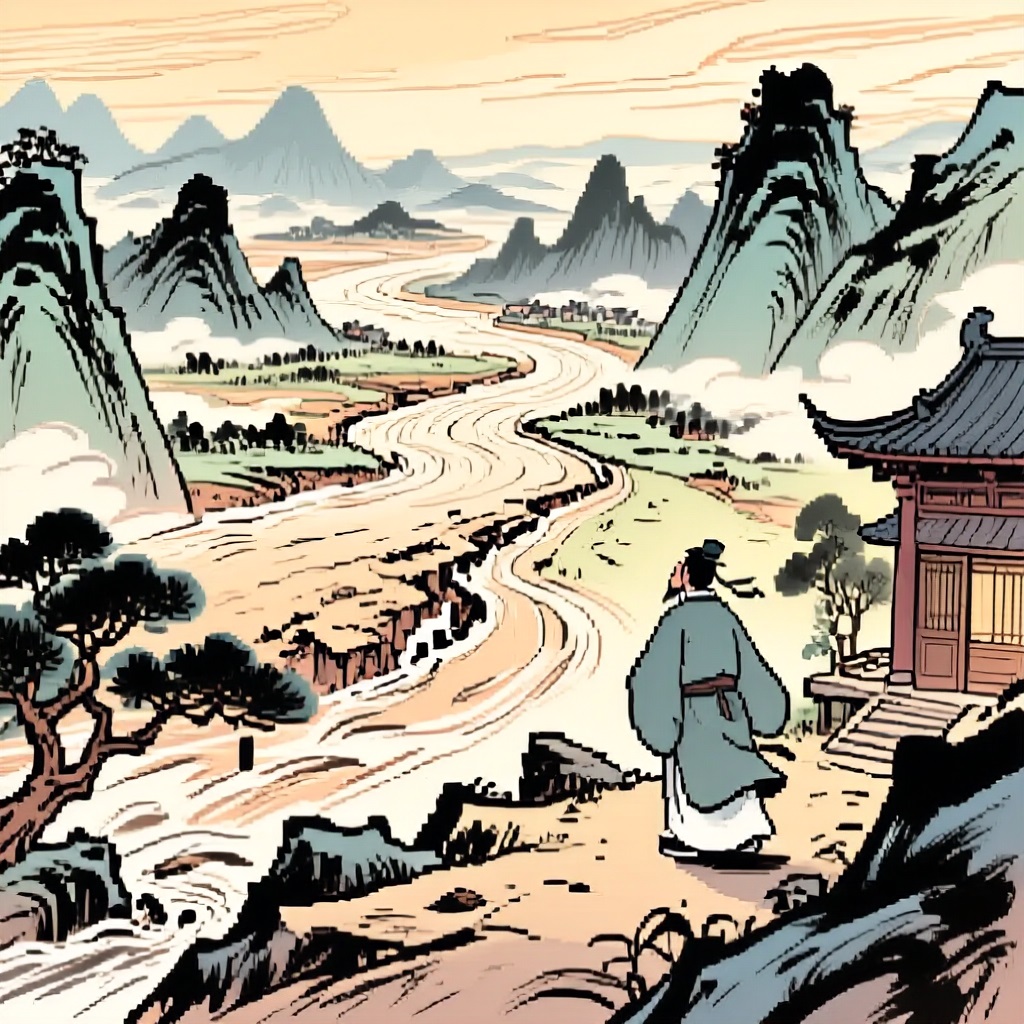} &
\poetryimg{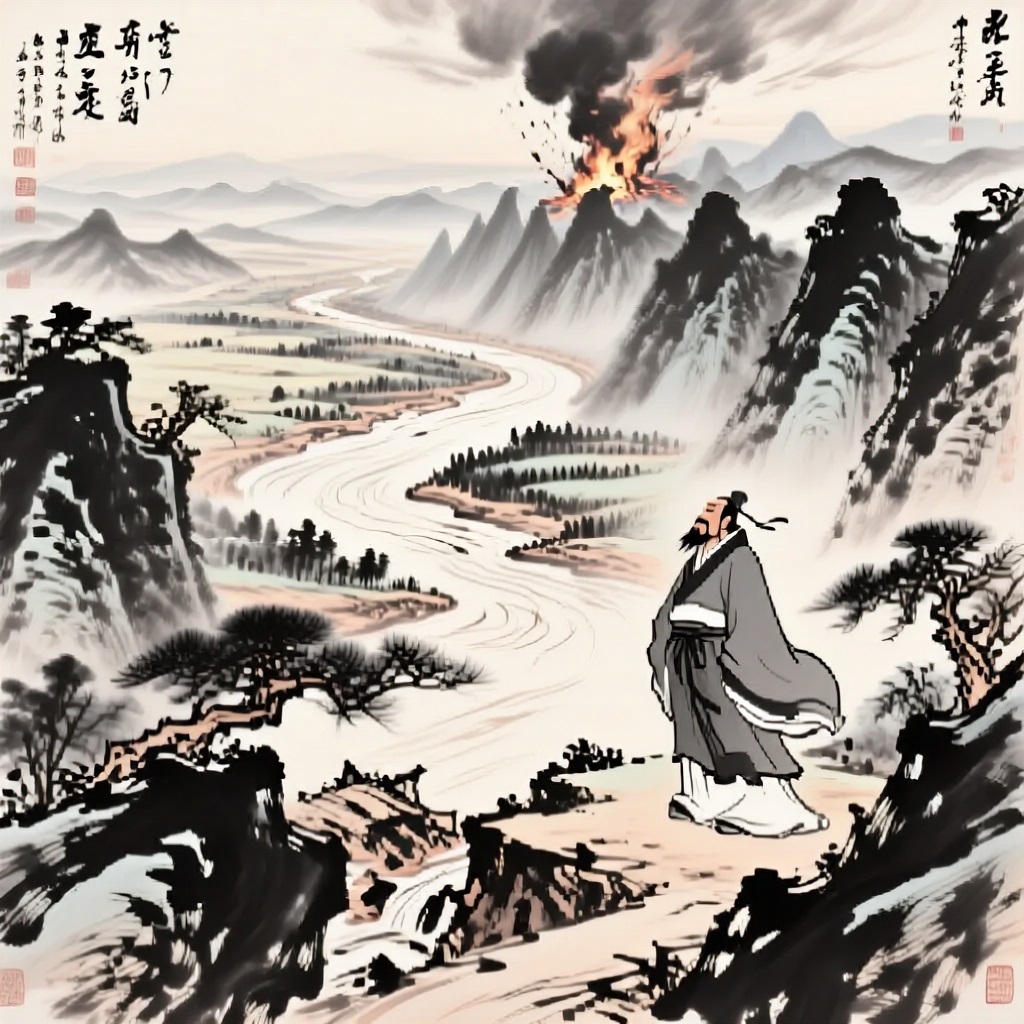} \\
\multicolumn{4}{c}{\footnotesize (c) \zh{几千年兴与亡，人间正道是沧桑}} \\[10pt]

\poetryimg{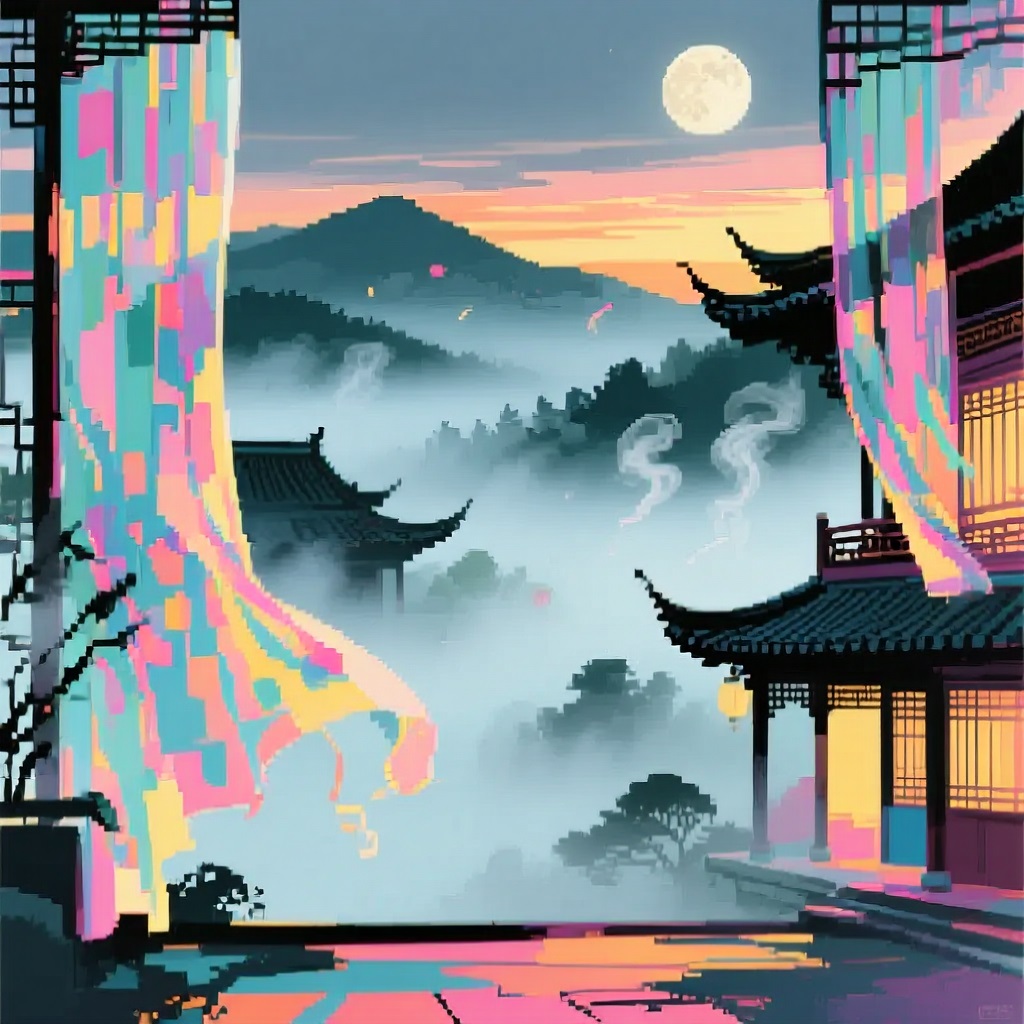} &
\poetryimg{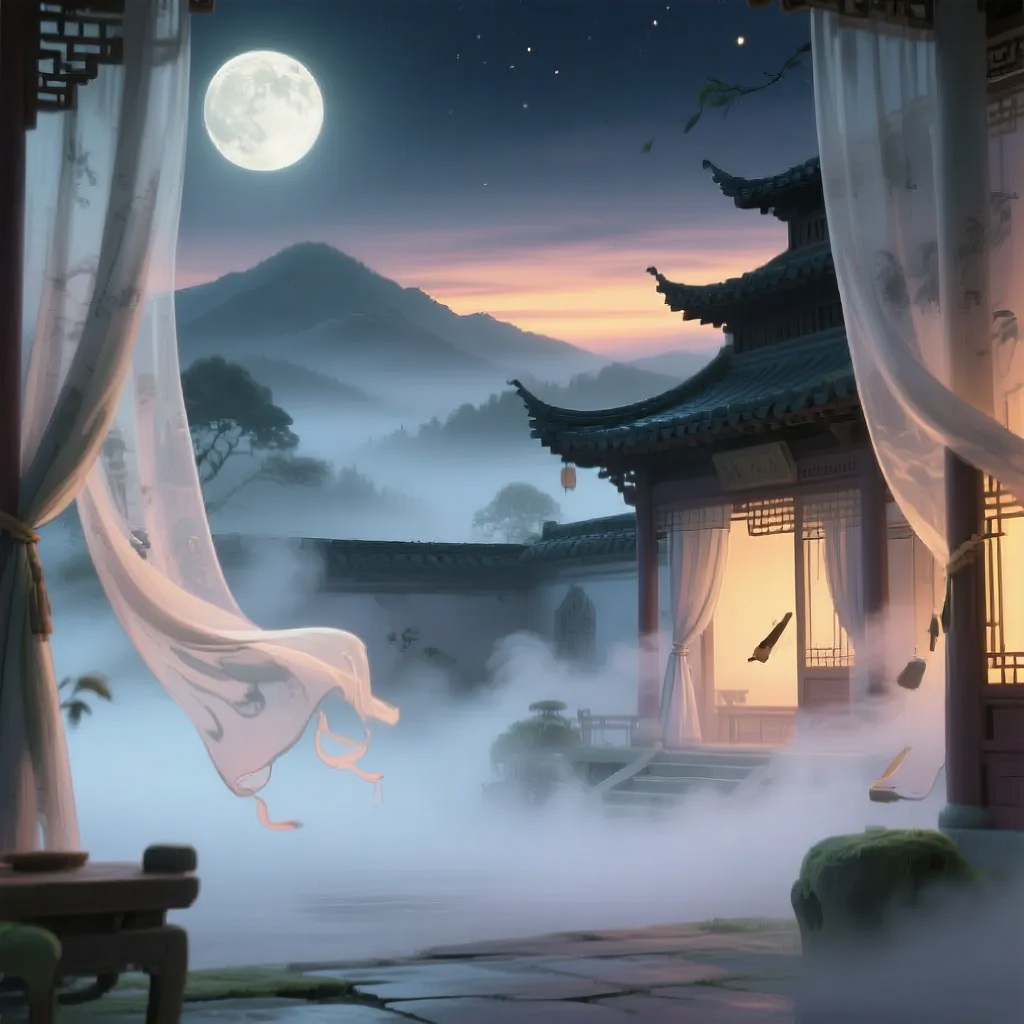} &
\poetryimg{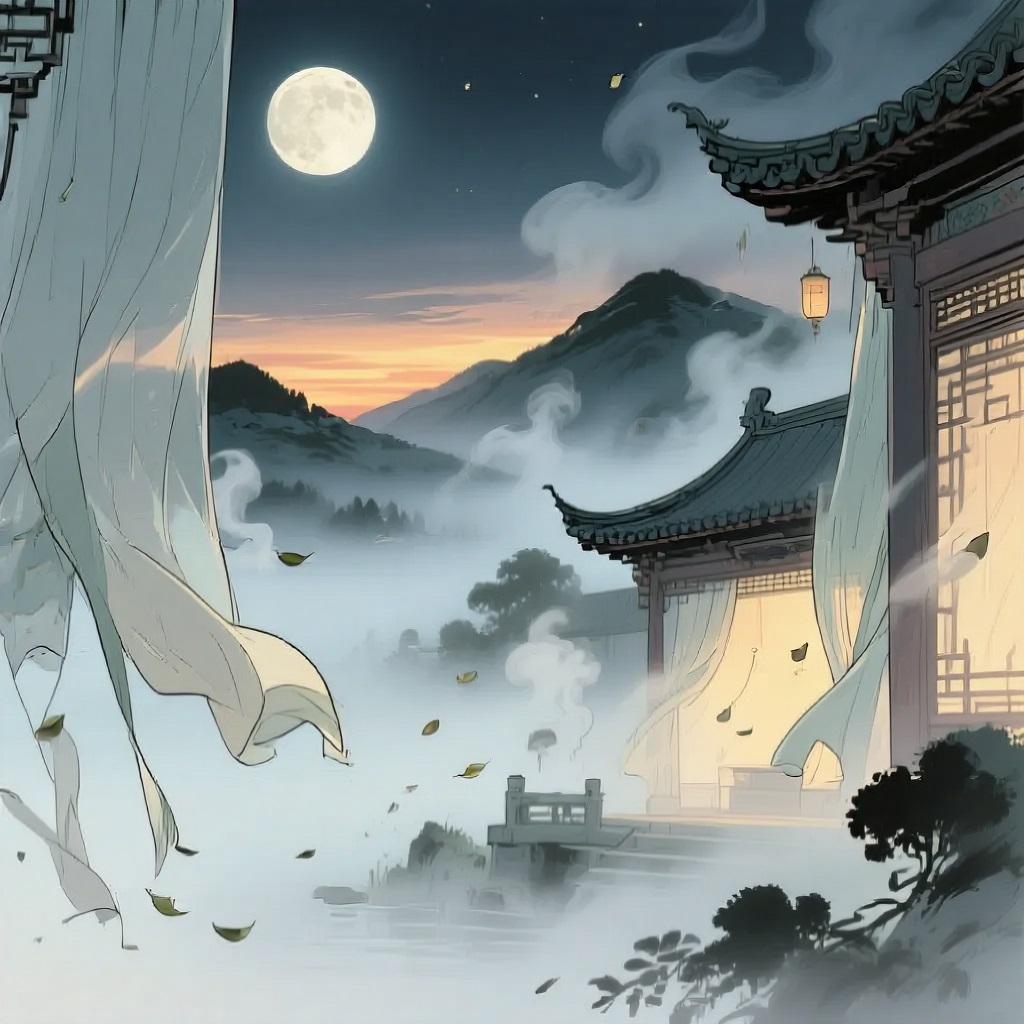} &
\poetryimg{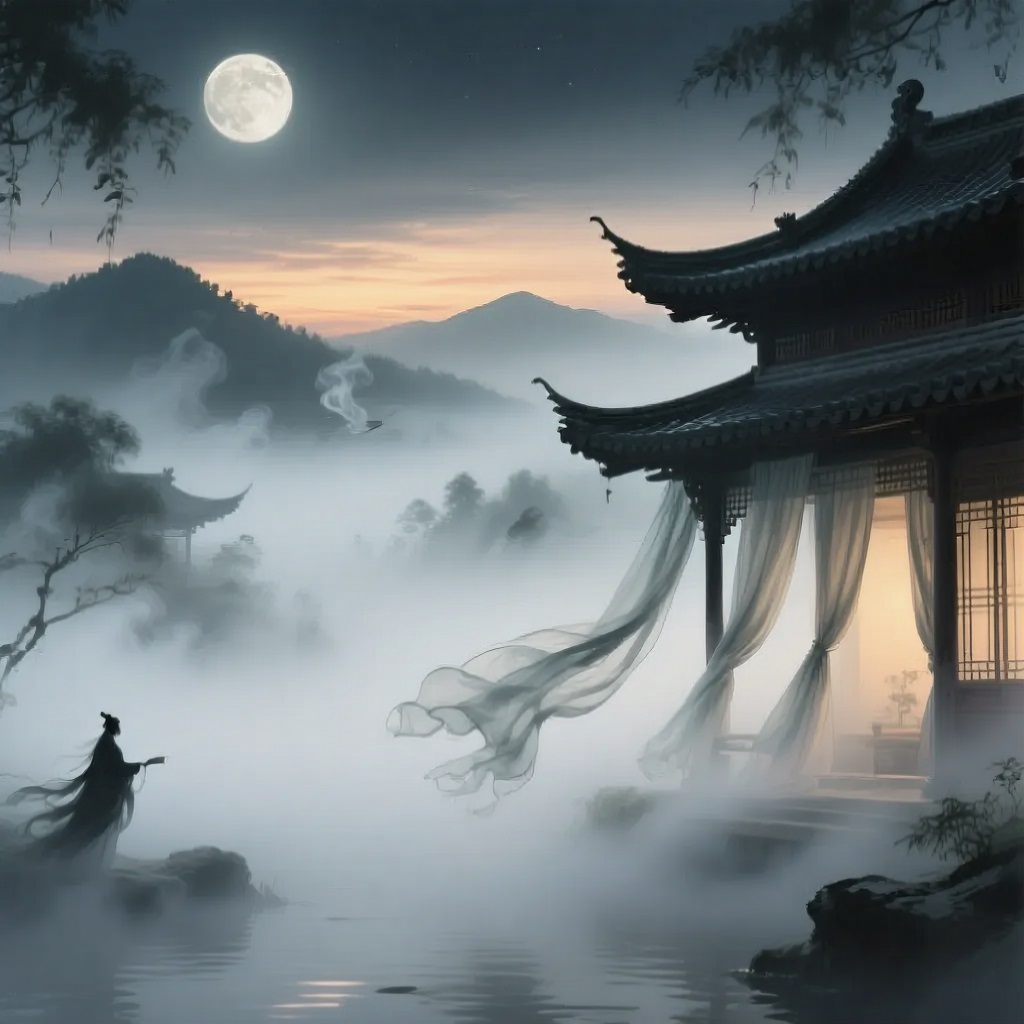} \\
\multicolumn{4}{c}{\footnotesize (d) \zh{梦凝白阑干，化为飞雾}} \\[6pt]

\end{tabular}

\caption{Visualization of images generated directly from classical Chinese poetry by our distilled 4-step text-to-image model under diverse artistic styles. Each row corresponds to a representative poetic verse, while each column presents the generated result in a
different style, including Pop Art, 3D Animated Film, Comic, and Chinese Classical Painting. The results show that the model is able to understand and render the semantic content, visual imagery, and aesthetic mood of classical Chinese poetry with high fidelity and
strong stylistic controllability, without requiring external English translation.}
\label{fig:chinese-poetry}
\end{figure}

\section{Application: Classical Chinese Poetry Image Generation}

To demonstrate the practical applicability of JuZhou 1.0 on a real-world culturally grounded task, we developed a mobile application named Mojie (\zh{墨界}). The application is built upon our poetry-oriented generative model, which is fine-tuned on a large-scale classical Chinese poetry dataset constructed through our own curation pipeline. After fine-tuning, the model is deployed as an Android edge application, enabling users to access poetry-to-image generation capabilities directly on mobile devices. The Android application is publicly available for download.\footnote{\url{https://www.pgyer.com/mojiemobilellm-android}}

\subsection{Application Overview}

Mojie is a practical mobile system that transforms classical Chinese poetry into high-quality images. By deploying the fine-tuned model directly on Android devices rather than relying on cloud inference, it demonstrates user-facing on-device generation in real-world applications.

The underlying model is fine-tuned on our synthetic poem-image corpus to better capture poetic semantics, scene composition, and traditional Chinese artistic imagery. Through this domain adaptation, Mojie generates visually coherent and semantically aligned images from poetic inputs. As an end-to-end prototype, it integrates data curation, model fine-tuning, edge deployment, and user interaction. Fig.~\ref{fig:mojie_overview} shows the overall interface and main functional modules of the Android application.

\subsection{Offline On-Device Poetry-to-Image Generation}

One of the key features of Mojie is its ability to generate high-quality poetry images entirely in offline and disconnected environments. Once the application and model files are installed on the device, users can perform local inference without requiring network connectivity. This design is particularly valuable for scenarios involving poor connectivity, strict privacy requirements, or a preference for low-latency local execution.

From the user perspective, the interaction is straightforward: a user may input a poem directly or select one from the built-in poetry library, and the application then performs on-device inference to produce a corresponding image. This functionality demonstrates that high-quality poetry-to-image generation can be achieved directly on Android smartphones, validating the feasibility of deploying specialized generative models in practical edge settings. Fig.~\ref{fig:generation_in_mojie} shows an example of offline poetry-to-image generation in Mojie.

\subsection{Built-in Classical Poetry Library}

To improve usability and enrich the application experience, Mojie includes a built-in poetry library containing tens of thousands of classical Chinese poems, covering a wide historical range from the pre-Qin period to the modern era. This extensive collection allows users to conveniently browse poems across dynasties and literary styles within a unified mobile interface.

Beyond simple reading functionality, the poetry library is tightly integrated with the generative model. Users can long-press a poem to directly trigger the image generation process, transforming the selected literary content into a visual result. This interaction design lowers the barrier for non-expert users and provides an intuitive bridge between classical literature and generative AI. Fig.~\ref{fig:built_in_poetry_library} illustrates the built-in poetry library spanning multiple historical periods.

\subsection{Community Sharing and Social Interaction}

In addition to local generation, Mojie incorporates a community-sharing mechanism that supports lightweight social interaction. After generating an image, users can publish the result to the in-app community, where other users can browse the generated content, express appreciation through likes, and participate in discussions through comments.

This design extends the application from a single-user generation tool into a creative and interactive platform. On the one hand, it increases user engagement and encourages repeated use. On the other hand, it creates a channel for sharing diverse visual interpretations of classical poetry, thereby enhancing both the cultural expressiveness and the social value of the system. Fig.~\ref{fig:community_mojie} presents the community page where users can publish and browse poetry-image results.

\subsection{Data Curation}
\label{sec:Data Curation}

The poetry-to-image training corpus is constructed using the dedicated pipeline described in Section~2. In summary, the pipeline transforms approximately 330,000 classical Chinese poems into 1,773,880 high-quality image-text pairs by combining style augmentation, image synthesis, quality filtering, and LLM-based recaptioning.

\subsection{Model Training and Evaluation}
\label{sec:modelTrainingAndEvaluation}

We now present the training and evaluation of the specialized poetry-to-image model deployed in Mojie. The following subsections detail the domain-specific fine-tuning procedure on the high-quality ancient Chinese poetry dataset, the subsequent 4-step distillation strategy for efficient on-device deployment, and both quantitative and qualitative evaluations against mainstream diffusion baselines.

\subsubsection{Training Details}
\label{sec:TrainingDetails}
To adapt our foundational model into a specialized text-to-image generator for ancient Chinese poetry, we fine-tune it on the high-quality ancient Chinese poetry dataset introduced earlier. The entire training process strictly follows the progressive resolution scaling and co-optimization procedures described in the Experiments section. This domain adaptation, inspired by recent advances in personalized and domain-specific diffusion~\cite{ruiz2023dreambooth,gal2022textual,hu2023animate}, aligns the model with the distinctive artistic imagery and visual aesthetics of classical Chinese poetry while preserving its general generative capacity. After foundational fine-tuning, we further apply a 4-step accelerated distillation strategy~\cite{luo2023lcm,luo2023lcm_loRA,hu2024lora} to enable efficient on-device inference. All adaptation experiments are conducted on the Sugon K100 AI cluster using BF16 mixed-precision training.

\subsubsection{Quantitative Results} 
\label{sec:QuantitativeResults} 
To quantitatively evaluate classical Chinese poetry-to-image generation, we assess two complementary aspects: text--image semantic alignment and distributional image fidelity. We compare \textbf{JuZhou~1.0} with representative diffusion baselines, including Stable Diffusion v2.1, SDXL, LCM, SANA-0.6B, and SDXL-Lightning.\footnote{SDXL-Lightning is evaluated with SDXL Base 1.0 as the backbone, and SANA-0.6B uses the 1024-pixel Diffusers checkpoint.}
Since most baselines are primarily optimized for English inputs and provide limited native support for classical Chinese, we evaluate them under both direct Chinese prompting and translated English prompting. In both settings, the original poems are first expanded into generation-oriented prompts using a large language model; for the English setting, the enhanced Chinese prompts are further translated into English before image generation. Chinese-prompt outputs are evaluated using CN-CLIP, while English-prompt outputs are evaluated using CLIP.

\begin{table}[t]
    \centering
    \caption{\textbf{Quantitative evaluation on the poetry image generation benchmark.}
    We report results under Chinese and English prompt settings. Chinese prompts are evaluated with CN-CLIP, while English prompts are evaluated with CLIP. FID is computed against an in-domain held-out reference set.}
    \label{tab:poetry_quantitative}
    \footnotesize
    \setlength{\tabcolsep}{8.0pt}
    \renewcommand{\arraystretch}{1.08}

    \begin{tabular}{lccccccc}
        \toprule
        \textbf{Model} 
        & \textbf{Params$\downarrow$}
        & \textbf{Mobile}
        & \textbf{Prompt}
        & \textbf{Evaluator}
        & \textbf{Steps} 
        & \textbf{CLIP Score$\uparrow$}
        & \textbf{FID$\downarrow$} \\
        \midrule

        \multirow{2}{*}{SDXL}
        & \multirow{2}{*}{2.6B}
        & \multirow{2}{*}{{\color{red}\ding{55}}}
        & CN & CN-CLIP & 50 & 18.74 & 135.65 \\
        & & & EN & CLIP & 50 & 29.96 & \textit{77.04} \\
        \addlinespace[2pt]

        \multirow{2}{*}{SDXL-Lightning}
        & \multirow{2}{*}{2.6B}
        & \multirow{2}{*}{{\color{red}\ding{55}}}
        & CN & CN-CLIP & 4  & 18.54 & 164.62 \\
        & & & EN & CLIP & 4  & 29.21 & 73.70 \\
        \addlinespace[2pt]
    
        \multirow{2}{*}{SDv2.1}
        & \multirow{2}{*}{1.3B}
        & \multirow{2}{*}{{\color{red}\ding{55}}}
        & CN & CN-CLIP & 50 & 17.27 & 109.42 \\
        & & & EN & CLIP & 50 & 28.51 & 77.76 \\
        \addlinespace[2pt]
    
        \multirow{2}{*}{LCM-4-Step}
        & \multirow{2}{*}{\textit{0.86B}}
        & \multirow{2}{*}{{\color{red}\ding{55}}}
        & CN & CN-CLIP & 4  & 19.63 & 159.90 \\
        & & & EN & CLIP & 4  & 28.34 & 98.83 \\
        \addlinespace[2pt]
        
        \multirow{2}{*}{LCM-28-Step}
        & \multirow{2}{*}{\textit{0.86B}}
        & \multirow{2}{*}{{\color{red}\ding{55}}}
        & CN & CN-CLIP & 28 & 18.86 & 159.61 \\
        & & & EN & CLIP & 28 & 27.78 & 93.73 \\
        \addlinespace[2pt]
    
        \multirow{2}{*}{SANA-0.6B}
        & \multirow{2}{*}{\underline{0.6B}}
        & \multirow{2}{*}{{\color{red}\ding{55}}}
        & CN & CN-CLIP & 20 & 22.02 & 84.32 \\
        & & & EN & CLIP & 20 & \underline{29.99} & 75.05 \\
        \midrule
    
        \rowcolor{DeepTeal!15}
        \textbf{JuZhou 1.0 (Ours)}
        & \textbf{0.387B}
        & {\color{green}\ding{51}}
        & CN & CN-CLIP & 28 & \textbf{30.32} & \textbf{26.11} \\
    
        \rowcolor{DeepTeal!15}
        \textbf{JuZhou 1.0 (distilled)}
        & \textbf{0.387B}
        & {\color{green}\ding{51}}
        & CN & CN-CLIP & 4 & 29.01 & \underline{52.54} \\
    
        \bottomrule
    \end{tabular}
    
    \renewcommand{\arraystretch}{1.0}
    \setlength{\tabcolsep}{6pt}
\end{table}

\textbf{Text--Image Alignment.} 
For the Chinese prompt setting, we use Chinese-CLIP (CN-CLIP) to measure the semantic consistency between generated images and their corresponding enhanced Chinese prompts. As shown in Tab.~\ref{tab:poetry_quantitative}, \textbf{JuZhou~1.0} achieves the highest CN-CLIP score of \textbf{30.32}, substantially outperforming all baselines evaluated with direct Chinese prompts. Its 4-step distilled variant obtains the second-highest score of \underline{29.01}, indicating that DMD2 distillation largely preserves Chinese-language semantic alignment while reducing the sampling process from 28 to 4 steps. In contrast, the Chinese-prompt scores of general-purpose baselines range from 17.27 to 22.02, suggesting that models primarily optimized for English inputs struggle to interpret classical Chinese poetry under direct Chinese prompting.

For the translated English prompt setting, SANA-0.6B achieves the highest CLIP score of \underline{29.99}, closely followed by SDXL with 29.96 and SDXL-Lightning with 29.21. These results show that English-oriented models can benefit from translating Chinese poetry prompts into English. However, since CN-CLIP and CLIP rely on different vision--language encoders and embedding spaces, their scores reflect performance within separate language settings and should not be directly compared on a shared numerical scale.

\textbf{Distributional Image Fidelity.}
We use Fr\'echet Inception Distance (FID)~\cite{fid2023rethinking} to evaluate the distributional similarity between generated images and a held-out reference set from our curated poetry-image benchmark. Because this reference set is constructed using the same poetry-oriented data pipeline, the reported FID should be interpreted primarily as an \emph{in-domain} distributional fidelity measure, rather than as direct evidence of generalization to arbitrary real-world image distributions. Moreover, FID captures distribution-level similarity and does not independently assess whether each generated image accurately reflects the content of its corresponding poem.

Under this in-domain setting, \textbf{JuZhou~1.0} achieves the lowest FID of \textbf{26.11}, while its 4-step distilled variant obtains the second-lowest FID of \underline{52.54} among all compared settings. Both variants substantially outperform the best baseline result, which is 73.70 achieved by SDXL-Lightning under the translated English setting. These results indicate that JuZhou~1.0 more closely matches the visual characteristics of the target poetry-image distribution, while the distilled model retains strong in-domain fidelity despite its substantially reduced sampling cost.

Overall, the quantitative results show that JuZhou~1.0 achieves both strong native Chinese semantic alignment and strong in-domain visual fidelity on the constructed classical poetry-to-image benchmark. Meanwhile, the distilled variant reduces the sampling process from \textbf{28 steps} to \textbf{4 steps} while retaining a CN-CLIP score of 29.01 and an FID of 52.54, providing a substantially more efficient alternative for deployment-oriented image synthesis.

\subsubsection{Qualitative Results}
\label{sec:QualitativeResults}
As illustrated in Fig.~\ref{fig:chinese-poetry}, JuZhou 1.0 demonstrates exceptional qualitative performance in rendering classical Chinese poetry. Unlike conventional baselines that rely on external translation modules---which inherently dilute cultural nuances---JuZhou 1.0 directly and accurately interprets profound literary metaphors and abstract artistic conceptions (e.g., the solitude of a lone boat or the grandeur of an expansive desert). In merely 4 inference steps, the model synthesizes high-fidelity and stylistically diverse artworks, meticulously capturing the intricate aesthetics of traditional Chinese art, from the subtle brushstrokes of ink wash paintings to authentic historical attire and architectural elements. This fusion of poetic abstraction and precise visual grounding highlights the value of native Chinese pretraining for culturally sensitive generative tasks.
\section{Conclusion}
We presented JuZhou~1.0, an ultra-lightweight text-to-image foundation model for fully local on-device image generation after one-time model installation. JuZhou~1.0 uses a 0.385B-parameter denoising U-Net and a 1.90M-parameter distilled decoder, forming a compact 0.387B image-generation backbone with 4-step distilled inference and direct Chinese prompting. The 28-step base model achieves a GenEval overall score of 0.70 under a unified evaluation protocol, outperforming SD3-Medium despite its substantially smaller model size. For deployment, the full poetry-to-image pipeline is validated on Android, while the core CLIP--U-Net--VAE branch is validated on iOS. On a Snapdragon\textsuperscript{\textregistered} 8 Elite Gen 5 device, the 4-step $1024 \times 1024$ U-Net denoising branch runs in approximately $1.6$~s. The model training and DMD2 step-compression pipeline was completed on domestic Sugon K100 hardware, providing a practical reference for Chinese-native generation, domestic-compute training, and edge-side visual generation.

\section*{Core Contributors}
Ce Chen, Congrui Wang, Yonglin Li, Zhenchen Wan, Mingyang Geng, Junhao Xiao, Zhengpeng Xing, Yaqing Hu, Yao Wu, Zhaoyang Qu

\section*{Contributors}
Long Lan, Xinwang Liu, Yingqi Peng, Shijia Li, Zufeng Zhang (Tsinghua University), Chen Ma (City University of Hong Kong), Jingjing Zhou (SUGON), Xingyu Wang (SUGON), Qilin Lu (SUGON), Bin Jiang (SUGON), Qilin Sun (The Chinese University of Hong Kong, Shenzhen)

\section*{Project Leaders}
Shanzhi Gu, Yaoguang Jin

\section*{Acknowledgments}
Tongliang Liu, Kede Ma, and Yifan Peng (The University of Hong Kong).

\section*{Special Acknowledgments}
JuZhou V1.0 was publicly unveiled at a launch event\footnote{\url{https://www.icswb.com/default.php?mod=live_text&live_id=914&temp=live_video}}  in Changsha, China, on May 21, 2025. The accompanying livestream attracted approximately 426,000 cumulative online views. As we prepare for the forthcoming release of JuZhou V2.0, this technical report provides a consolidated account of JuZhou V1.0, covering its design, training pipeline, deployment practice, and empirical evaluation.

We sincerely thank Sugon for its support in domestically developed AI computing infrastructure, including approximately 80 PFLOPS of K100 accelerator resources, which enabled the end-to-end training pipeline of this work. We also gratefully acknowledge support from the Hunan Provincial Key Research and Development Program under Grant No.~2025JK2146 and the Hunan Provincial College Student Entrepreneurship Investment Fund.

\bibliographystyle{ieeetr}
\bibliography{references}

\end{document}